\documentclass[11pt,a4paper]{article}
\usepackage[hyperref]{emnlp2020}
\usepackage{times,xcolor}
\usepackage{latexsym}
\usepackage{amsfonts}
\usepackage{amsmath}
\usepackage{subcaption,grffile,booktabs,graphicx}
\usepackage{amssymb}
\usepackage{multicol}
\usepackage{rotating}
\usepackage{scrextend}
\usepackage{enumitem}
\usepackage{comment}
\usepackage{dblfloatfix}
\usepackage{setspace}


\usepackage{microtype}

\aclfinalcopy %

\newenvironment{itemize*}%
  {\begin{itemize}[noitemsep,topsep=0pt]%
    \setlength{\itemsep}{0.9pt}%
    \setlength{\parskip}{0.9pt}%
    \setlength{\topsep}{0.9pt}}%
  {\end{itemize}}

\title{When BERT Plays the Lottery, All Tickets Are Winning}

\author{Sai Prasanna\thanks{~~Equal contribution} \\

\small Zoho Labs \\
\small  Zoho Corporation \\

\small   Chennai, India\\ 
\small   \texttt{saiprasanna.r@zohocorp.com} \\\And
  Anna Rogers\footnotemark[1] \\
\small Center for Social Data Science \\
\small Copenhagen University \\
\small  Copenhagen, Denmark\\
\small   \texttt{arogers@sodas.ku.dk} \\\And
  Anna Rumshisky \\
\small Dept. of Computer Science \\
\small Univ. of Massachusetts Lowell \\
\small Lowell, USA \\
\small   \texttt{arum@cs.uml.edu}\\}

\date{}

\begin{document}

\maketitle
\begin{abstract}
Large Transformer-based models were shown to be reducible to a smaller number of self-attention heads and layers. We consider this phenomenon from the perspective of the lottery ticket hypothesis, using both structured and magnitude pruning. For fine-tuned BERT, we show that (a) it is possible to find subnetworks achieving performance that is comparable with that of the full model, and (b) similarly-sized subnetworks sampled from the rest of the model perform worse. Strikingly, with structured pruning even the worst possible subnetworks remain highly trainable, indicating that most pre-trained BERT weights are potentially useful.
We also study the ``good" subnetworks to see if their success can be attributed to superior linguistic knowledge, but find them unstable, and not explained by meaningful self-attention patterns.
\end{abstract}

\section{Introduction}

Much of the recent progress in NLP is due to the transfer learning paradigm in which Transformer-based models first try to learn task-independent linguistic knowledge from large corpora, and then get fine-tuned on small datasets for specific tasks. However, these models are overparametrized: we now know that most Transformer heads and even layers can be pruned without significant loss in performance \cite{VoitaTalbotEtAl_2019_Analyzing_Multi-Head_Self-Attention_Specialized_Heads_Do_Heavy_Lifting_Rest_Can_Be_Pruneda,KovalevaRomanovEtAl_2019_Revealing_Dark_Secrets_of_BERT,MichelLevyEtAl_2019_Are_Sixteen_Heads_Really_Better_than_One}. 

One of the most famous Transformer-based models is BERT \cite{DevlinChangEtAl_2019_BERT_Pre-training_of_Deep_Bidirectional_Transformers_for_Language_Understanding}. It became a must-have baseline and inspired dozens of studies probing it for various kinds of linguistic information \cite{RogersKovalevaEtAl_2020_Primer_in_BERTology_What_we_know_about_how_BERT_works}. 

We conduct a systematic case study of fine-tuning BERT on GLUE tasks \cite{WangSinghEtAl_2018_GLUE_A_Multi-Task_Benchmark_and_Analysis_Platform_for_Natural_Language_Understanding} from the perspective of the lottery ticket hypothesis \cite{FrankleCarbin_2019_Lottery_Ticket_Hypothesis_Finding_Sparse_Trainable_Neural_Networks}. 
We experiment with and compare magnitude-based weight pruning and importance-based pruning of BERT self-attention heads \cite{MichelLevyEtAl_2019_Are_Sixteen_Heads_Really_Better_than_One}, which we extend to multi-layer perceptrons (MLPs) in BERT.

With both techniques, we find the ``good" subnetworks that achieve 90\% of full model performance, and perform considerably better than similarly-sized subnetworks sampled from other parts of the model. However, in many cases even the ``bad" subnetworks can be re-initialized to the pre-trained BERT weights and fine-tuned separately to achieve strong performance. %
We also find that the ``good" networks are unstable across random initializations at fine-tuning, and their self-attention heads do not necessarily encode meaningful linguistic patterns.

\begin{table*}[]
\centering
\footnotesize
\begin{tabular}{p{.9cm}p{8.8cm}lll}
\toprule
Task & Dataset & Train & Dev & Metric \\
\midrule
CoLA & Corpus of Linguistic Acceptability Judgements \cite{WarstadtSinghEtAl_2019_Neural_Network_Acceptability_Judgments} & 10K & 1K & Matthews \\
SST-2 & The Stanford Sentiment Treebank \cite{SocherPerelyginEtAl_2013_Recursive_deep_models_for_semantic_compositionality_over_sentiment_treebank} & 67K & 872 & accuracy \\
MRPC & Microsoft Research Paraphrase Corpus \cite{DolanBrockett_2005_Automatically_constructing_a_corpus_of_sentential_paraphrases} & 4k & n/a & accuracy\\ %
STS-B & Semantic Textual Similarity Benchmark \cite{CerDiabEtAl_2017_SemEval-2017_Task_1_Semantic_Textual_Similarity_Multilingual_and_Crosslingual_Focused_Evaluation} & 7K & 1.5K & Pearson \\
QQP & Quora Question Pairs\footnote{\url{https://www.quora.com/q/quoradata/First-Quora-Dataset-Release-Question-Pairs}} \cite{WangSinghEtAl_2018_GLUE_A_Multi-Task_Benchmark_and_Analysis_Platform_for_Natural_Language_Understanding} & 400K & n/a & accuracy \\
MNLI & The Multi-Genre NLI Corpus (matched) \cite{WilliamsNangiaEtAl_2017_A_Broad-Coverage_Challenge_Corpus_for_Sentence_Understanding_through_Inference} & 393K & 20K & accuracy \\
QNLI & Question NLI \cite{RajpurkarZhangEtAl_2016_SquAD_100000+_Questions_for_Machine_Comprehension_of_Text,WangSinghEtAl_2018_GLUE_A_Multi-Task_Benchmark_and_Analysis_Platform_for_Natural_Language_Understanding} & 108K & 11K & accuracy \\
RTE & Recognizing Textual Entailment \cite{DaganGlickmanEtAl_2005_PASCAL_Recognising_Textual_Entailment_Challenge,HaimDaganEtAl_2006_Second_Pascal_Recognising_Textual_Entailment_Challenge,GiampiccoloMagniniEtAl_2007_Third_PASCAL_Recognizing_Textual_Entailment_Challenge,BentivogliClarkEtAl_2009_Fifth_PASCAL_Recognizing_Textual_Entailment_Challenge} & 2.7K & n/a & accuracy \\
WNLI & Winograd NLI \cite{LevesqueDavisEtAl_2012_Winograd_Schema_Challenge} & 706 & n/a & accuracy \\
\bottomrule
\end{tabular}
\caption{GLUE tasks \cite{WangSinghEtAl_2018_GLUE_A_Multi-Task_Benchmark_and_Analysis_Platform_for_Natural_Language_Understanding}, dataset sizes and the metrics reported in this study}
\label{tab:glue-tasks}
\end{table*}

\section{Related Work}

Multiple studies of BERT concluded that it is considerably overparametrized. In particular, it is possible to ablate elements of its architecture without loss in performance or even with slight gains \cite{KovalevaRomanovEtAl_2019_Revealing_Dark_Secrets_of_BERT,MichelLevyEtAl_2019_Are_Sixteen_Heads_Really_Better_than_One,VoitaTalbotEtAl_2019_Analyzing_Multi-Head_Self-Attention_Specialized_Heads_Do_Heavy_Lifting_Rest_Can_Be_Pruneda}. This explains the success of multiple BERT compression studies  \cite{model:distilBERT,jiao2019tinybert,mccarley2019pruning,model:albert}. 

While NLP focused on building larger Transformers, the computer vision community was exploring the Lottery Ticket Hypothesis \cite[LTH: ][]{FrankleCarbin_2019_Lottery_Ticket_Hypothesis_Finding_Sparse_Trainable_Neural_Networks,LeeAjanthanEtAl_2018_SNIP_Single-shot_network_pruning_based_on_connection_sensitivity,ZhouLanEtAl_2019_Deconstructing_Lottery_Tickets_Zeros_Signs_and_Supermask}. It is formulated as follows: \textit{``dense, randomly-initialized, feed-forward networks contain subnetworks (winning tickets) that -- when trained in isolation -- reach test accuracy comparable to the original network in a similar number of iterations"} \cite{FrankleCarbin_2019_Lottery_Ticket_Hypothesis_Finding_Sparse_Trainable_Neural_Networks}. The ``winning tickets"  generalize across vision datasets \cite{MorcosYuEtAl_2019_One_ticket_to_win_them_all_generalizing_lottery_ticket_initializations_across_datasets_and_optimizers}, and exist both in LSTM and Transformer models for NLP \cite{YuEdunovEtAl_2020_Playing_lottery_with_rewards_and_multiple_languages_lottery_tickets_in_RL_and_NLP}.

However, so far LTH work focused on the ``winning" random initializations. In case of BERT, there is a large pre-trained language model, used in conjunction with a randomly initialized task-specific classifier; this paper and concurrent work by \citet{ChenFrankleEtAl_2020_Lottery_Ticket_Hypothesis_for_Pre-trained_BERT_Networks} are the first to explore LTH in this context. The two papers provide complementary results for magnitude pruning, but we also study structured pruning, posing the question of whether ``good" subnetworks can be used as an tool to understand how BERT works. Another contemporaneous study by \citet{Gordon2020CompressingBS} also explores magnitude pruning, showing that BERT pruned before fine-tuning still reaches performance similar to the full model.

Ideally, the pre-trained weights would provide transferable linguistic knowledge, fine-tuned only to learn a given task. But we do not know what knowledge actually gets used for inference, except that BERT is as prone as other models to rely on dataset biases \cite{McCoyPavlickEtAl_2019_Right_for_Wrong_Reasons_Diagnosing_Syntactic_Heuristics_in_Natural_Language_Inference,RogersKovalevaEtAl_2020_Getting_Closer_to_AI_Complete_Question_Answering_Set_of_Prerequisite_Real_Tasks,JinJinEtAl_2020_Is_BERT_Really_Robust_Strong_Baseline_for_Natural_Language_Attack_on_Text_Classification_and_Entailment,NivenKao_2019_Probing_Neural_Network_Comprehension_of_Natural_Language_Arguments,ZellersHoltzmanEtAl_2019_HellaSwag_Can_Machine_Really_Finish_Your_Sentence}. 
At the same time, there is vast literature on probing BERT architecture blocks for different linguistic properties \cite{RogersKovalevaEtAl_2020_Primer_in_BERTology_What_we_know_about_how_BERT_works}. If there are ``good" subnetworks, then studying their properties might explain how BERT works.

\section{Methodology}
\label{sec:methodology}

All experiments in this study are done on the ``BERT-base lowercase" model from the Transformers library \cite{WolfDebutEtAl_2020_HuggingFaces_Transformers_State-of-the-art_Natural_Language_Processing}. It is fine-tuned\footnote{All experiments were performed with 8 RTX 2080 TI GPUs, 128 Gb of RAM, 2x CPU Intel(R) Xeon(R) CPU E5-2630 v4 @ 2.20GHz. Code repository: \url{https://github.com/sai-prasanna/bert-experiments}.} %
on 9 GLUE tasks, and evaluated with the metrics shown in \autoref{tab:glue-tasks}. All evaluation is done on the dev sets, as the test sets are not publicly distributed. %
For each experiment we test 5 random seeds.

\subsection{BERT Architecture}

BERT is fundamentally a stack of Transformer encoder layers  \cite{VaswaniShazeerEtAl_2017_Attention_is_all_you_need}. All layers have identical structure: a multi-head self-attention (MHAtt) block followed by an MLP, with residual connections around each. 

MHAtt consists of $N_h$ independently parametrized heads. An attention head $h$ in layer $l$ is parametrized by $ W_k^{(h,l)}, W_q^{(h,l)}, W_v^{(h,l)} \in \mathbb{R}^{d_h \times d} $, $ W_o^{(h,l)} \in \mathbb{R}^{d \times d_h} $. $d_h$ is typically set to $d/N_h$. 
Given \textit{n} $d$-dimensional input vectors $\text{x} = {x_1, x_2, .. x_n} \in \mathbb{R}^d$, MHAtt is the sum of the output of each individual head applied to input $x$:

\small
\begin{equation*}
\label{eq:mhatt}
\text{MHAtt}^{(l)}(\text{x}) = 
\sum\limits_{h=1}^{N_h} \text{Att}_{W_k^{(h,l)},W_q^{(h,l)},W_v^{(h,l)},W_o^{(h,l)}}^{(l)}(\text{x})
\end{equation*}
\normalsize

The MLP %
in layer $l$ consists of two feed-forward layers. It is applied separately to \textit{n} \textit{d}-dimensional vectors $\text{z} \in \mathbb{R}^d$ coming from the attention sub-layer. Dropout \cite{Srivastava2014DropoutAS} is used for regularization. Then inputs of the MLP are added to its outputs through a residual connection.

\subsection{Magnitude Pruning}
\label{sec:magnitude}

For magnitude pruning, we fine-tune BERT on each task and iteratively prune 10\% of the lowest magnitude weights across the entire model (excluding the embeddings, since this work focuses on BERT's body weights). We check the \textit{dev} set score in each iteration and keep pruning for as long as the performance remains above $90\%$ of the full fine-tuned model's performance. Our methodology and results are complementary to those by \citet{ChenFrankleEtAl_2020_Lottery_Ticket_Hypothesis_for_Pre-trained_BERT_Networks}, who perform \textit{iterative} magnitude pruning while fine-tuning the model to find the mask.

\begin{figure*}[t]
        \begin{subfigure}[t]{0.48\textwidth}
        \includegraphics[trim=-20 -10 -30 30,clip,width=\linewidth]{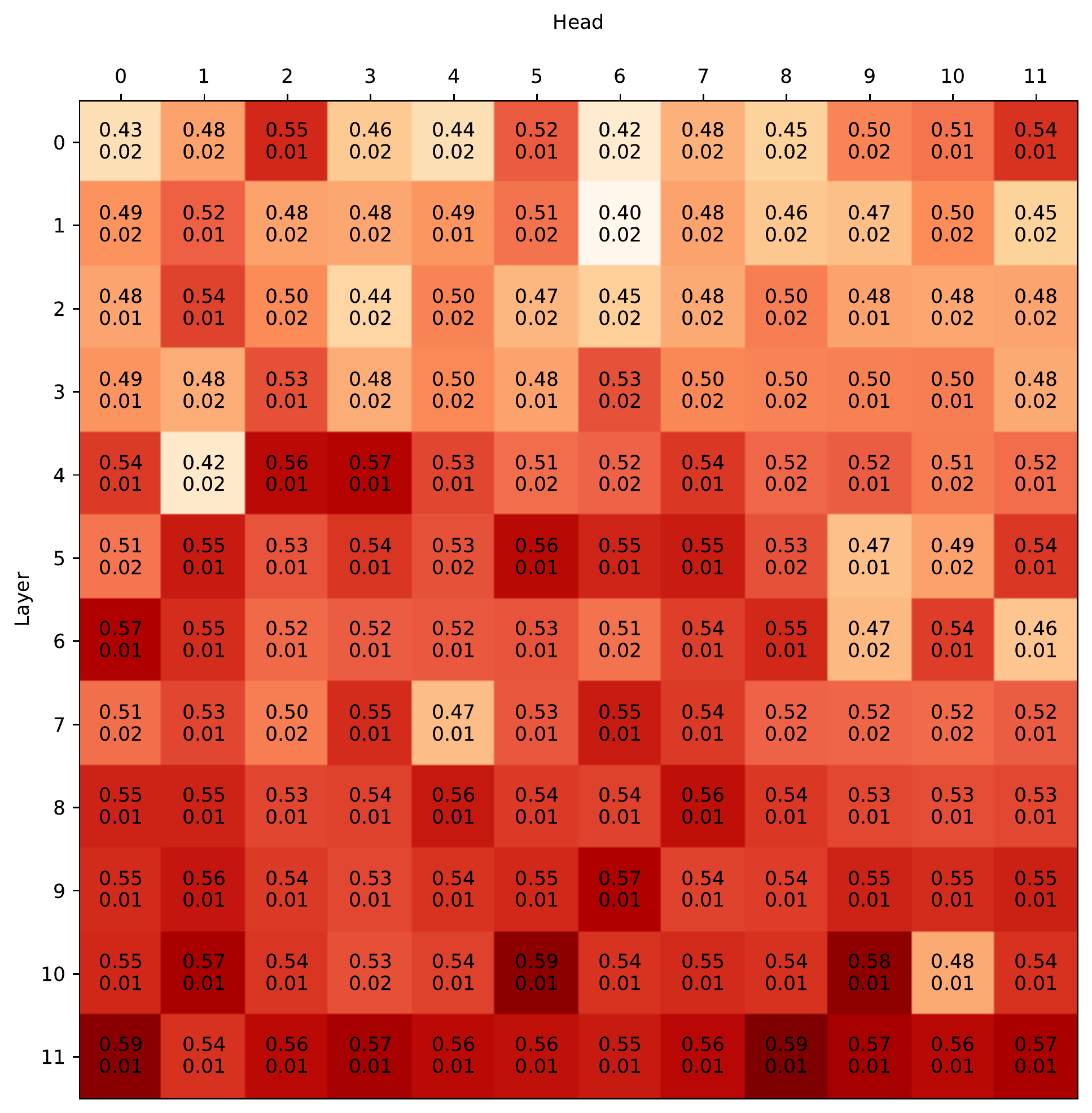}
        \includegraphics[trim=-20 -30 -20 30,clip,width=\linewidth]{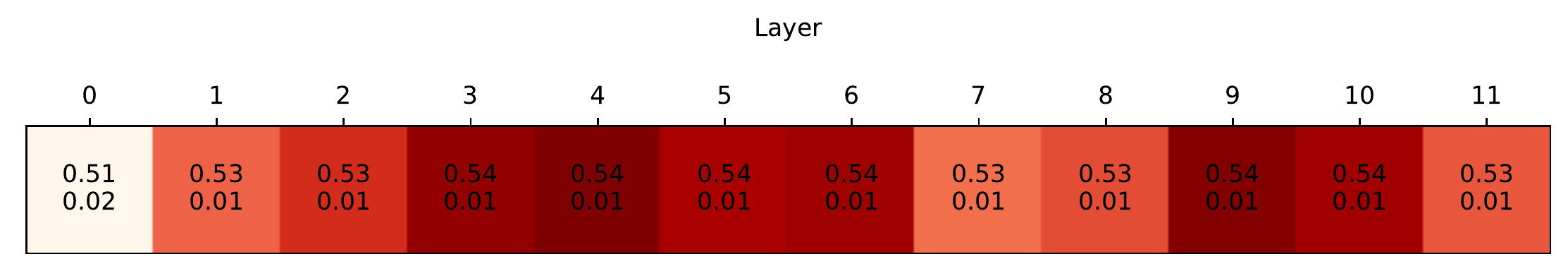}        
        \caption{M-pruning: each cell gives the percentage of surviving weights, and std across 5 random seeds.}        
        \label{fig:heatmap-magnitude}
        \end{subfigure}
\hfill
        \begin{subfigure}[t]{0.48\textwidth}
        \includegraphics[trim=-20 -10 -30 30,clip,width=\linewidth]{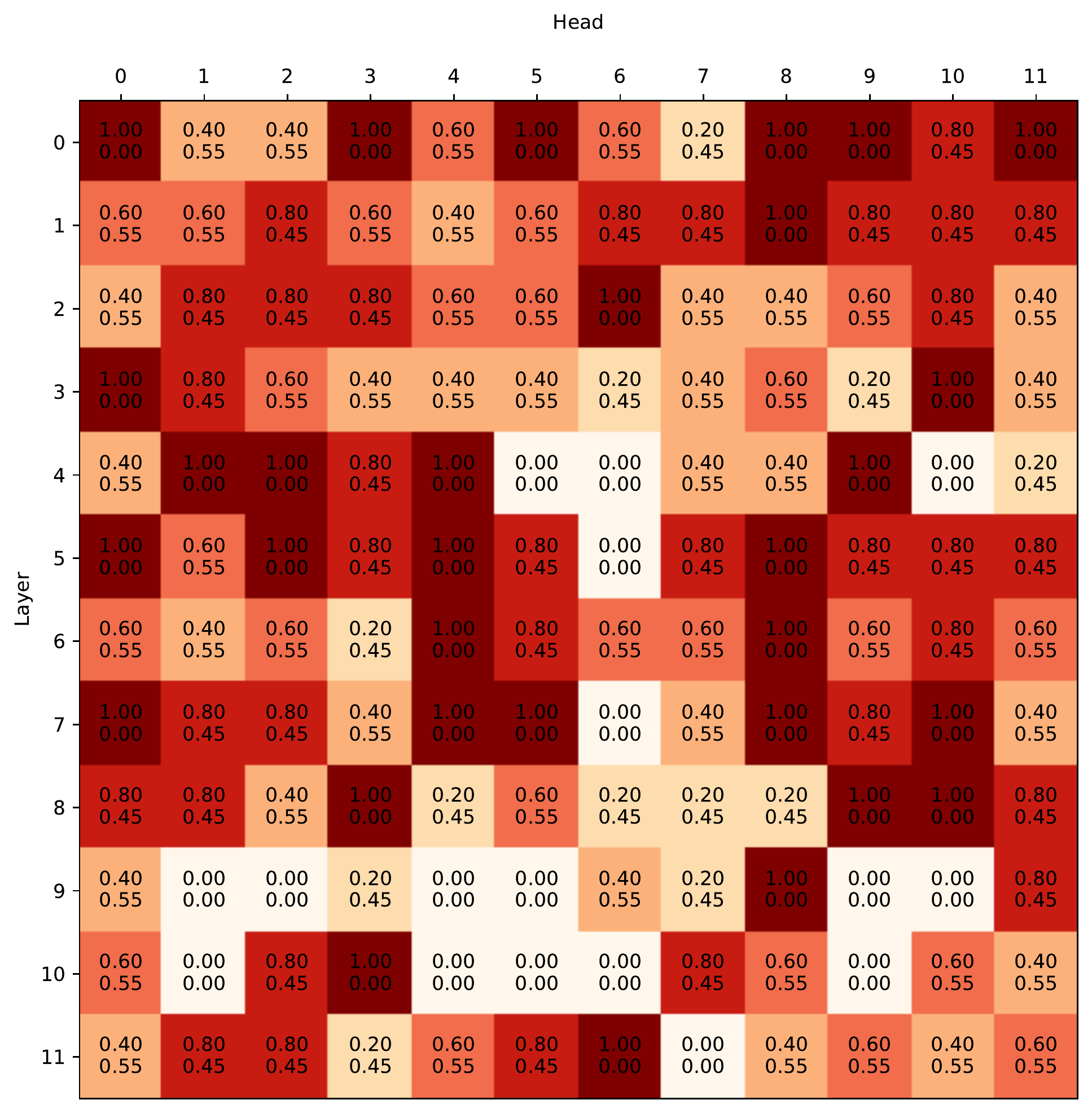}
        \includegraphics[trim=-20 -30 -20 30,clip,width=\linewidth]{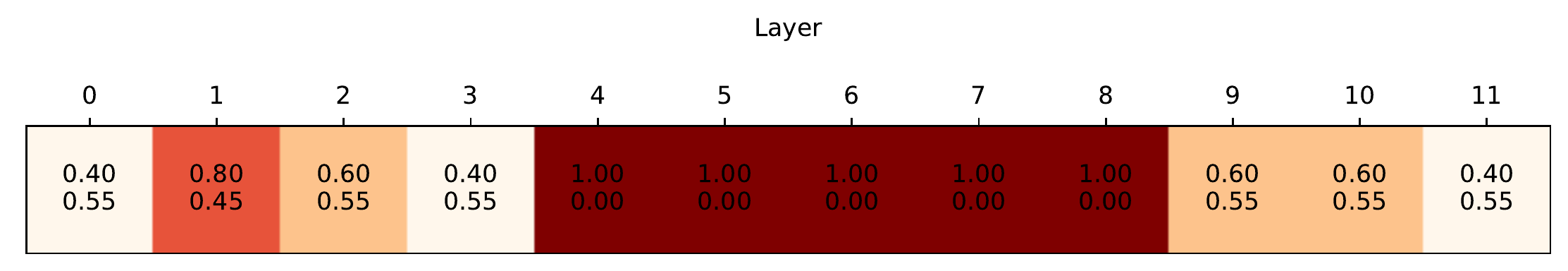}
        \caption{S-pruning: each cell gives the average number of random seeds in which a given head/MLP survived and std.}
        \label{fig:heatmap-importance}        
        \end{subfigure}

        \caption{The ``good" subnetworks for QNLI: self-attention heads (top, 12 x 12 heatmaps) and MLPs (bottom, 1x12 heatmaps), pruned together. Earlier layers start at 0.}

        \label{fig:good-subnetworks}
\end{figure*}

\subsection{Structured Pruning}
\label{subsec:Importance score}

We study structured pruning of BERT architecture blocks, masking them under the constraint that at least 90\% of full model performance is retained. Combinatorial search to find such masks is impractical, and 
\citet{MichelLevyEtAl_2019_Are_Sixteen_Heads_Really_Better_than_One} estimate the importance of attention heads as the expected sensitivity to the mask variable $\xi^{(h,l)}$: 

\small
\begin{equation*}
\label{eq:head_importance}
I_h^{(h,l)} = E_{x \sim X}\left|\dfrac{\partial\mathcal{L}(x)}{\partial \xi^{(h,l) }}\right|
\end{equation*}
\normalsize

\noindent
where $x$ is a sample from the data distribution $X$ and $\mathcal{L}(x)$ is the loss of the network outputs on that sample. We extend this approach to MLPs, with the mask variable $\nu^{(l)}$:

\small
\begin{equation*}
\label{eq:mlp_importance}
I_{mlp}^{(l)} = E_{x \sim X}\left|\dfrac{\partial\mathcal{L}(x)}{\partial \nu^{(l) }}\right|
\end{equation*}
\normalsize

If $I_h^{(h,l)}$ and $I_{mlp}^{(l)}$ are high, they have a large effect on the model output. Absolute values are calculated to avoid highly positive contributions nullifying highly negative contributions.

In practice, calculating $I_h^{(h,l)}$ and $I_{mlp}^{(l)}$ would involve computing backward pass on the loss over samples of the evaluation data\footnote{The GLUE dev sets are used as oracles to obtain the best heads and MLPs for the particular model and task.}. We follow \citeauthor{MichelLevyEtAl_2019_Are_Sixteen_Heads_Really_Better_than_One} in applying the recommendation of \citet{DBLP:conf/iclr/MolchanovTKAK17} to normalize the importance scores of the attention heads layer-wise (with $\ell2$ norm) before pruning. %
To mask the heads, we use a binary mask variable $\xi^{(h,l)}$. If $\xi^{(h,l)} = 0$, the head $h$ in layer $l$ is masked:

\small
\begin{equation*}
\label{eq:head_mask}
\text{MHAtt}^{(l)}(\text{x}) = %
\sum\limits_{h=1}^{N_h} \textcolor{red}{\xi^{(h,l)}} \text{Att}_{W_k^{(h,l)},W_q^{(h,l)},W_v^{(h,l)},W_o^{(h,l)}}^{(l)}(\text{x})
\end{equation*}
\normalsize

Masking MLPs in layer $l$ is performed similarly with a masking variable $\nu^{(l)}$:

\small
\begin{equation*}
\label{eq:mlp_mask}
\text{MLP}_{\text{out}}^{(l)}(\text{z}) = \textcolor{red}{\nu^{(l)}} {MLP}^{(l)}(\text{z}) + \text{z}
\end{equation*}
\normalsize

We compute head and MLP importance scores in a single backward pass, pruning 10\% heads and one MLP with the smallest scores until the performance on the \textit{dev} set is within 90\%. Then we continue pruning heads alone, and then MLPs alone. The process continues iteratively for as long as the pruned model retains over $90\%$ performance of the full fine-tuned model. 

We refer to magnitude and structured pruning as \textit{m-pruning} and \textit{s-pruning}, respectively.

\section{BERT Plays the Lottery}

\subsection{The ``Good" Subnetworks}
\label{sec:good-subnets}

\autoref{fig:good-subnetworks} shows the heatmaps for the ``good" subnetworks for QNLI, i.e. the ones that retain 90\% of full model performance after pruning.

For s-pruning,
we show the number of random initializations in which a given head/MLP survived the pruning. For m-pruning, 
we compute the percentage of 
surviving weights in BERT heads and MLPs in all GLUE tasks (excluding embeddings). We run each experiment with 5 random initializations of the task-specific layer (the same ones), and report averages and standard deviations. %
See \autoref{appendix:good-subnets-per-task} for other GLUE tasks.

\autoref{fig:heatmap-magnitude} shows that in m-pruning, all architecture blocks lose about half the weights (42-57\% weights), but the earlier layers get pruned more. %
With s-pruning (\autoref{fig:heatmap-importance}), the most important heads 
tend to be in the earlier and middle layers, while the important MLPs are more in the middle. Note that  \citet{LiuGardnerEtAl_2019_Linguistic_Knowledge_and_Transferability_of_Contextual_Representations} also find that the middle Transformer layers are the most transferable. %

In \autoref{fig:heatmap-importance}, the heads and MLPs were pruned together. 
The overall pattern 
is similar 
when they are pruned separately.
While fewer heads (or MLPs) remain when they are pruned separately (49\% vs 22\% for heads, 75\% vs 50\% for MLPs), pruning them together is more efficient overall (i.e., produces smaller subnetworks).
Full data is available in \autoref{appendix:pruning-modes}. 
This experiment hints at considerable interaction between BERT's self-attention heads and MLPs: with fewer MLPs available, the model is forced to rely more on the heads, raising their importance. This interaction was not explored in the previous studies %
\cite{MichelLevyEtAl_2019_Are_Sixteen_Heads_Really_Better_than_One,VoitaTalbotEtAl_2019_Analyzing_Multi-Head_Self-Attention_Specialized_Heads_Do_Heavy_Lifting_Rest_Can_Be_Pruneda,KovalevaRomanovEtAl_2019_Revealing_Dark_Secrets_of_BERT}, and deserves more attention in future work. 

\subsection{Testing LTH for BERT Fine-tuning: ~~~~~ The Good, the Bad and the Random}
\label{sec:lottery}

\begin{figure*}

        \begin{subfigure}[b]{\textwidth}
       
               \centering \includegraphics[trim=10 9 10 0,clip, scale=0.48]{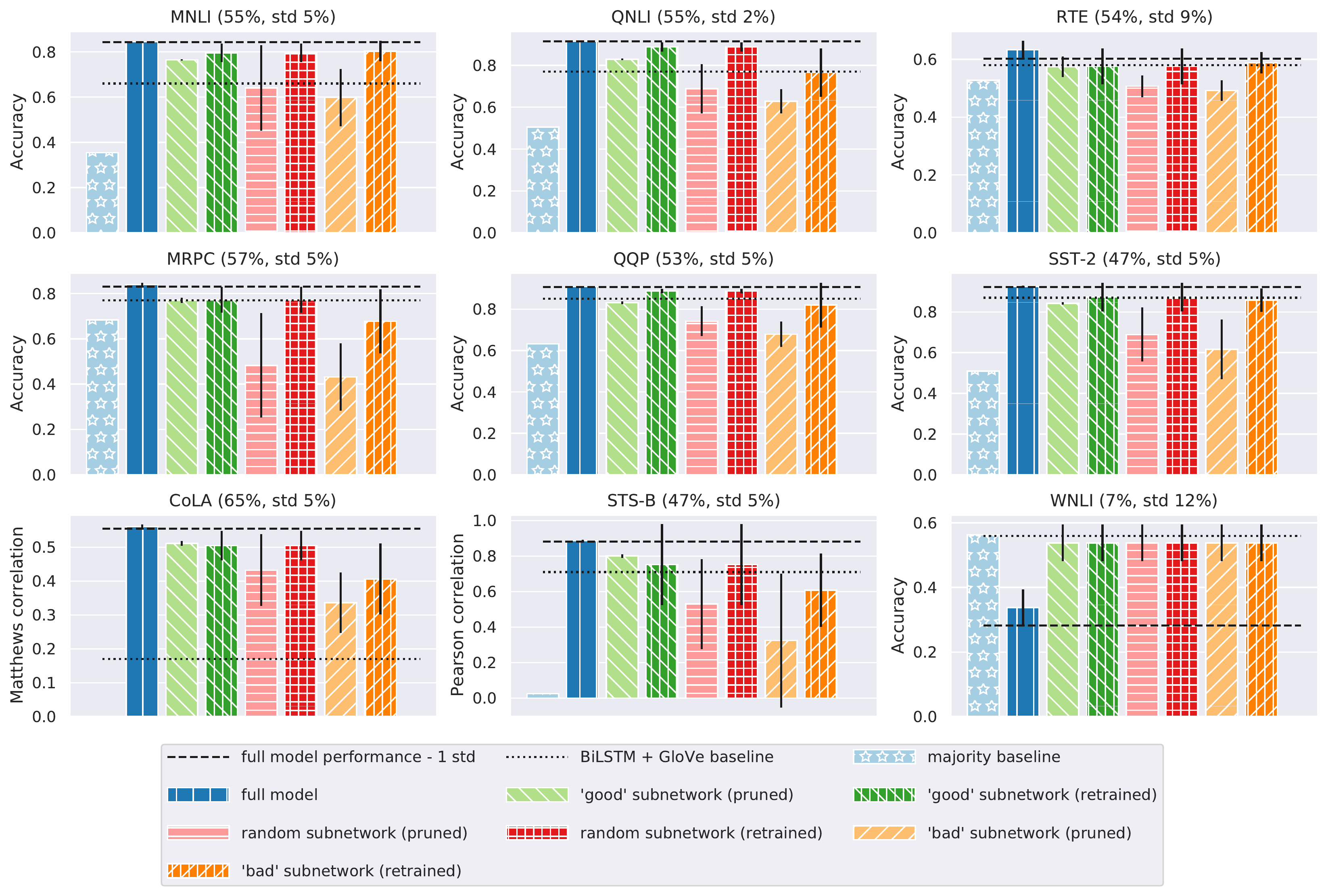}

\caption{S-pruning}
\end{subfigure}        

\begin{subfigure}[b]{\textwidth}
            
 \centering \includegraphics[trim=10 9 10 0,clip,scale=0.48]{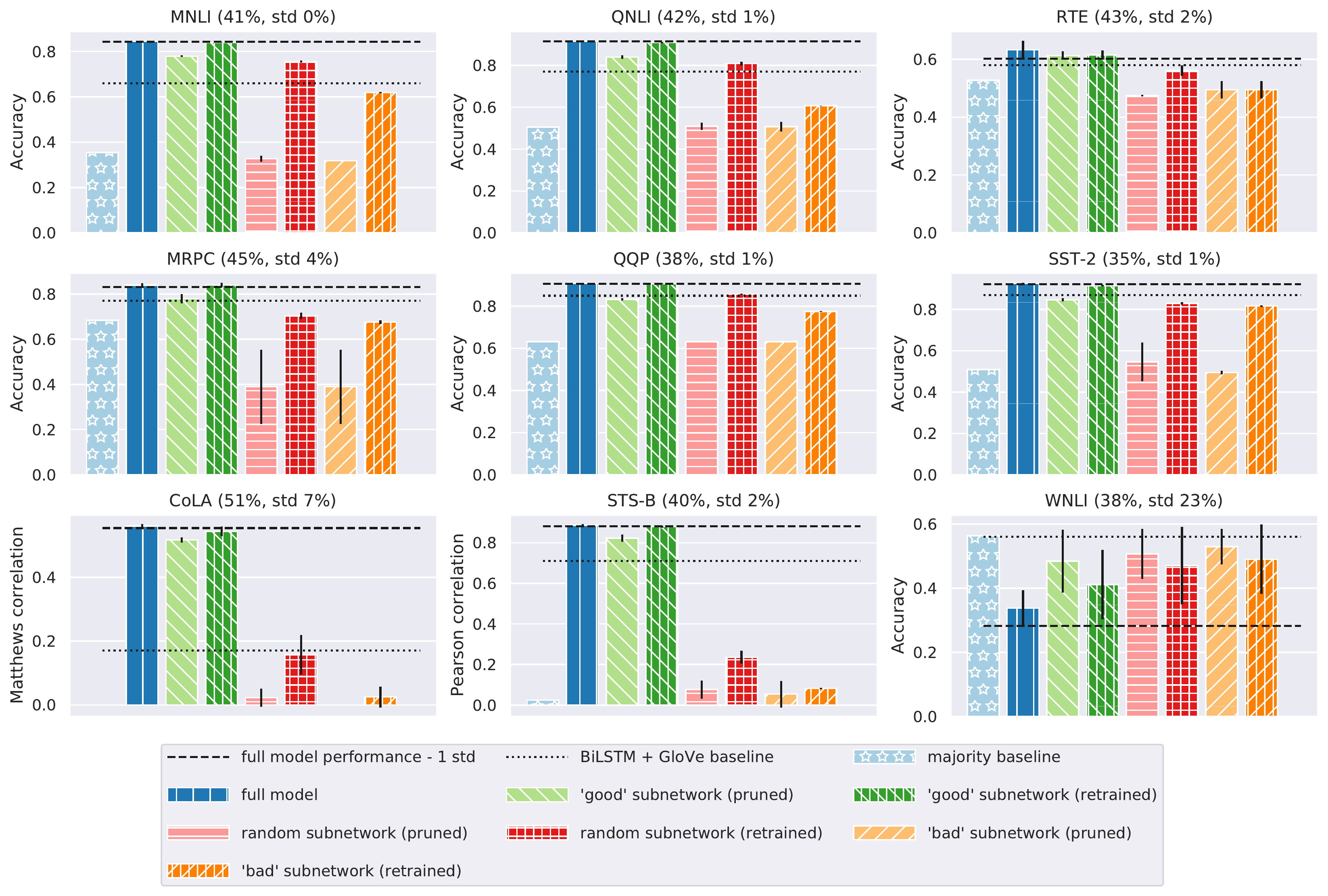}
        \label{fig:magnitude-barplot}
        \caption{M-pruning}
        \end{subfigure}
        
        \caption{The ``good" and ``bad" subnetworks in BERT fine-tuning: performance on GLUE tasks. `Pruned' subnetworks are only pruned, and `retrained' subnetworks are restored to pretrained weights and fine-tuned. Subfigure titles indicate the task and percentage of surviving weights. STD values and error bars indicate standard deviation of surviving weights and performance respectively, across 5 fine-tuning runs. See \autoref{appendix:evaluation-metrics-glue} for numerical results, and \autoref{sec:bad} for GLUE baseline discussion.}
\vspace{0.1cm}        
        
        \label{fig:lottery}

\end{figure*}

\begin{table*}[]
\footnotesize
\centering
\begin{tabular}{@{}p{4.1cm}p{.6cm}p{.8cm}p{.6cm}p{.6cm}p{.88cm}p{.6cm}p{.6cm}p{.6cm}p{.6cm}p{1.2cm}@{}}
\toprule
Model                   & CoLA & SST-2 & MRPC & QQP  & STS-B & MNLI & QNLI & RTE  & WNLI & Average\footref{glue-footnote}\\ \midrule
Majority class baseline & 0.00    & 0.51  & 0.68 & 0.63 & 0.02  & 0.35 & 0.51 & 0.53 & \textbf{0.56} & 0.42\\ 
CBOW                    & \textbf{0.46} & 0.79  & 0.75 & 0.75 & 0.70   & 0.57 & 0.62 & 0.71 & \textbf{0.56} & 0.61 \\
BILSTM + GloVe                  & 0.17 & 0.87  & \textbf{0.77} & 0.85 & \textbf{0.71}  & 0.66 & \textbf{0.77} & \textbf{0.58} & \textbf{0.56} & 0.66\\
BILSTM + ELMO           & 0.44 & \textbf{0.91}  & 0.70  & \textbf{0.88} & 0.70   & \textbf{0.68} & 0.71 & 0.53 & 0.56 & \textbf{0.68}\\
`Bad' subnetwork (s-pruning)                     & \underline{0.40}  & \underline{0.85} & \underline{0.67} & \underline{0.81} & \underline{0.60}  & \underline{0.80}  & \underline{0.76} & \underline{0.58} & \underline{0.53} & \underline{0.67}\\
`Bad' subnetwork (m-pruning)                            & 0.24 & 0.81 & \underline{0.67} & 0.77 & 0.08 & 0.61 & 0.6  & 0.49 & 0.49 & 0.51\\
Random init + random s-pruning & 0.00    & 0.78 & \underline{0.67} & 0.78 & 0.14 & 0.63 & 0.59 & 0.53 & 0.50 & 0.52\\
\bottomrule
\end{tabular}
\caption{``Bad" BERT subnetworks (the best one is underlined) vs basic baselines (the best one is bolded). The randomly initialized BERT is randomly pruned by importance scores to match the size of s-pruned `bad' subnetwork.}
\label{tab:bad-subnetworks}
\end{table*}

LTH predicts that the ``good" subnetworks trained from scratch should match the full network performance. We experiment with the following settings:

\begin{itemize*}
    \item \textit{``good" subnetworks}: the elements selected from the full model by either technique; %
    \item \textit{random subnetworks}: the same size as ``good" subnetworks, but with elements randomly sampled from the full model;
    \item \textit{``bad" subnetworks}: the elements sampled from those that did \textit{not} survive the pruning, plus a random sample of the remaining elements so as to match the size of the ``good" subnetworks.
\end{itemize*}

For both pruning methods, we evaluate the subnetworks (a) after pruning, (b) after retraining the same subnetwork. The model is re-initialized to pre-trained weights (except embeddings), and the task-specific layer is initialized with the same random seeds that were used to find the given mask.

As mentioned earlier, the evaluation is performed on the GLUE\footnote{The results for WNLI are unreliable: this dataset has similar sentences with opposite labels in train and dev data, and in s-pruning the whole model gets pruned away. See \autoref{appendix:good-subnets-per-task} for discussion of that.} \textit{dev} sets, 
which have also been used 
to identify the the ``good" subnetworks originally. 
These subnetworks were chosen to work well on this specific data, 
and the corresponding ``bad" subnetworks were defined only in relation to the ``good" ones.
We therefore do not expect these subnetworks to generalize to other data, and believe that they would best illustrate what exactly BERT ``learns" in fine-tuning.

Performance of each subnetwork type is shown in \autoref{fig:lottery}.%
The main LTH prediction is validated: the ``good" subnetworks can be successfully re-trained alone. Our m-pruning results are consistent with contemporaneous work by \citet{Gordon2020CompressingBS} and \citet{ChenFrankleEtAl_2020_Lottery_Ticket_Hypothesis_for_Pre-trained_BERT_Networks}. 

We observe the following differences between the two pruning techniques:

\begin{itemize*}

\item For 7 out of 9 tasks m-pruning yields considerably higher compression (10-15\% more weights pruned) than s-pruning.

\item Although m-pruned subnetworks are smaller, they mostly reach\footnote{For convenience, \autoref{fig:lottery} shows the performance of the full model minus one standard deviation -- the success criterion for the subnetwork also used by \citet{ChenFrankleEtAl_2020_Lottery_Ticket_Hypothesis_for_Pre-trained_BERT_Networks}.} the full network performance. For s-pruning, the ``good" subnetworks are mostly slightly behind the full network performance. %

 \item Randomly sampled subnetworks could be expected to perform better than the ``bad", but worse than the ``good" ones. That is the case for m-pruning, but for s-pruning they mostly perform on par with the ``good" subnetworks, suggesting the subset of ``good" heads/MLPs in the random sample suffices to reach the full ``good" subnetwork performance.

\end{itemize*}

Note that our pruned subnetworks are relatively large with both pruning methods (mostly over 50\% of the full model). For s-pruning, we also look at ``super-survivors": much smaller subnetworks consisting only of the heads and MLPs that consistently survived across all seeds for a given task. For most tasks, these subnetworks contained only about 10-26\% of the full model weights, but lost only about 10 performance points on average. See \autoref{appendix:super-survivors} for the details for this experiment.

\subsection{How Bad are the ``Bad" Subnetworks?}
\label{sec:bad}

Our study -- as well as work by \citet{ChenFrankleEtAl_2020_Lottery_Ticket_Hypothesis_for_Pre-trained_BERT_Networks} and \citet{Gordon2020CompressingBS} -- provides conclusive evidence for the existence of ``winning tickets", but it is intriguing that for most GLUE tasks random masks in s-pruning perform nearly as well as the ``good" masks - i.e. they could also be said to be ``winning". In this section we look specifically at the ``bad" subnetworks: since in our setup, we use the \textit{dev} set both to find the masks and to test the model, these parts of the model are the \textit{least} useful for that specific data sample, and their trainability could yield important insights for model analysis. 

\autoref{tab:bad-subnetworks} shows the results for the ``bad" subnetworks pruned with both methods and re-fine-tuned, together with \textit{dev} set results of three GLUE baselines by \citet{WangSinghEtAl_2018_GLUE_A_Multi-Task_Benchmark_and_Analysis_Platform_for_Natural_Language_Understanding}. The m-pruned `bad' subnetwork is at least 5 points behind the s-pruned one on 6/9 tasks, and is particularly bad on the correlation tasks (CoLA and STS-B). With respect to GLUE baselines, the s-pruned ``bad" subnetwork is comparable to BiLSTM+ELMO and BiLSTM+GloVe. Note that there is a lot of variation between tasks: the `bad' s-pruned subnetwork is competitive with BiLSTM+GloVe in 5/9 tasks, but it loses by a large margin in 2 more tasks, and wins in 2 more (see also \autoref{fig:lottery}). 

The last line of \autoref{tab:bad-subnetworks} presents a variation of experiment with fine-tuning randomly initialized BERT by \citet{KovalevaRomanovEtAl_2019_Revealing_Dark_Secrets_of_BERT}: we randomly initialize BERT and also apply a randomly s-pruned mask so as to keep it the same size as the s-pruned ``bad" subnetwork. Clearly, even this model is in principle trainable (and still beats the majority class baseline), but on average\footnote{GLUE leaderboard uses macro average of metrics to rank the participating systems. We only consider the metrics in \autoref{tab:glue-tasks} to obtain this average.\label{glue-footnote}} it is over 15 points behind the ``bad" mask over the pre-trained weights. This shows that even the worst possible selection of pre-trained BERT components for a given task still contains a lot of useful information. In other words, some lottery tickets are ``winning" and yield the biggest gain, but all subnetworks have a non-trivial amount of useful information.

Note that even the random s-pruning of a randomly initialized BERT is slightly better than the m-pruned ``bad" subnetwork. It is not clear what plays a bigger role: the initialization or the architecture. \citet{ChenFrankleEtAl_2020_Lottery_Ticket_Hypothesis_for_Pre-trained_BERT_Networks} report that pre-trained weights do not perform as well if shuffled, but they do perform better than randomly initialized weights. %
To test whether the ``bad'' s-pruned subnetworks might match the ``good'' ones with more training, we trained them for 6 epochs, but on most tasks the performance went down (see \autoref{appendix:train-badsubnets-longer}).

Finally, BERT is known to sometimes have degenerate runs (i.e. with final performance much lower than expected) on smaller datasets \cite{DevlinChangEtAl_2019_BERT_Pre-training_of_Deep_Bidirectional_Transformers_for_Language_Understanding}. Given the masks found with 5 random initializations, we find that standard deviation of GLUE metrics for both ``bad" and ``random" s-pruned subnetworks %
is over 10 points not only for the smaller datasets (MRPC, CoLA, STS-B), but also for MNLI and SST-2 (although on the larger datasets the standard deviation goes down after re-fine-tuning). This illustrates the fundamental cause of degenerate runs: the poor match between the model and final layer initialization. Since our ``good" subnetworks are specifically selected to be the best possible match to the specific random seed, the performance is the most reliable. As for m-pruning, standard deviation remains low even for the ``bad" and ``random" subnetworks in most tasks except MRPC. See \autoref{appendix:evaluation-metrics-glue} for full results.

\section{Interpreting BERT's Subnetworks}

In \autoref{sec:lottery} we showed that the subnetworks found by m- and s-pruning behave similarly in fine-tuning. However, s-pruning has an advantage in that the functions of BERT architecture blocks have been extensively studied (see detailed overview by \citet{RogersKovalevaEtAl_2020_Primer_in_BERTology_What_we_know_about_how_BERT_works}). If the better performance of the ``good" subnetworks comes from linguistic knowledge, they could tell a lot about the reasoning BERT actually performs at inference time.

\subsection{Stability of the ``Good" Subnetworks}
\label{sec:stability}

Random initializations in the task-specific classifier interact with the pre-trained weights, affecting the performance of fine-tuned BERT \cite{McCoy2019BERTsOA,DodgeIlharcoEtAl_2020_Fine-Tuning_Pretrained_Language_Models_Weight_Initializations_Data_Orders_and_Early_Stopping}. 
However, if better performance comes from linguistic knowledge, we would expect the ``good" subnetworks to better encode this knowledge, and to be relatively stable across fine-tuning runs for the same task. %

We found the opposite. For all tasks, Fleiss' kappa on head survival across 5 random seeds was in the range of 0.15-0.32, and %
Cochran Q test did not show that the binary mask of head survival obtained with five random seeds for each tasks were significantly similar at $\alpha=0.05$ (although masks obtained with some \textit{pairs} of seeds were). This means that the ``good" subnetworks are unstable, and depend on the random initialization more than utility of a certain portion of pre-trained weights for a particular task. 

\begin{figure}[!h]
    \centering
    \includegraphics[trim=10 0 10 0,clip,width=.7\linewidth]{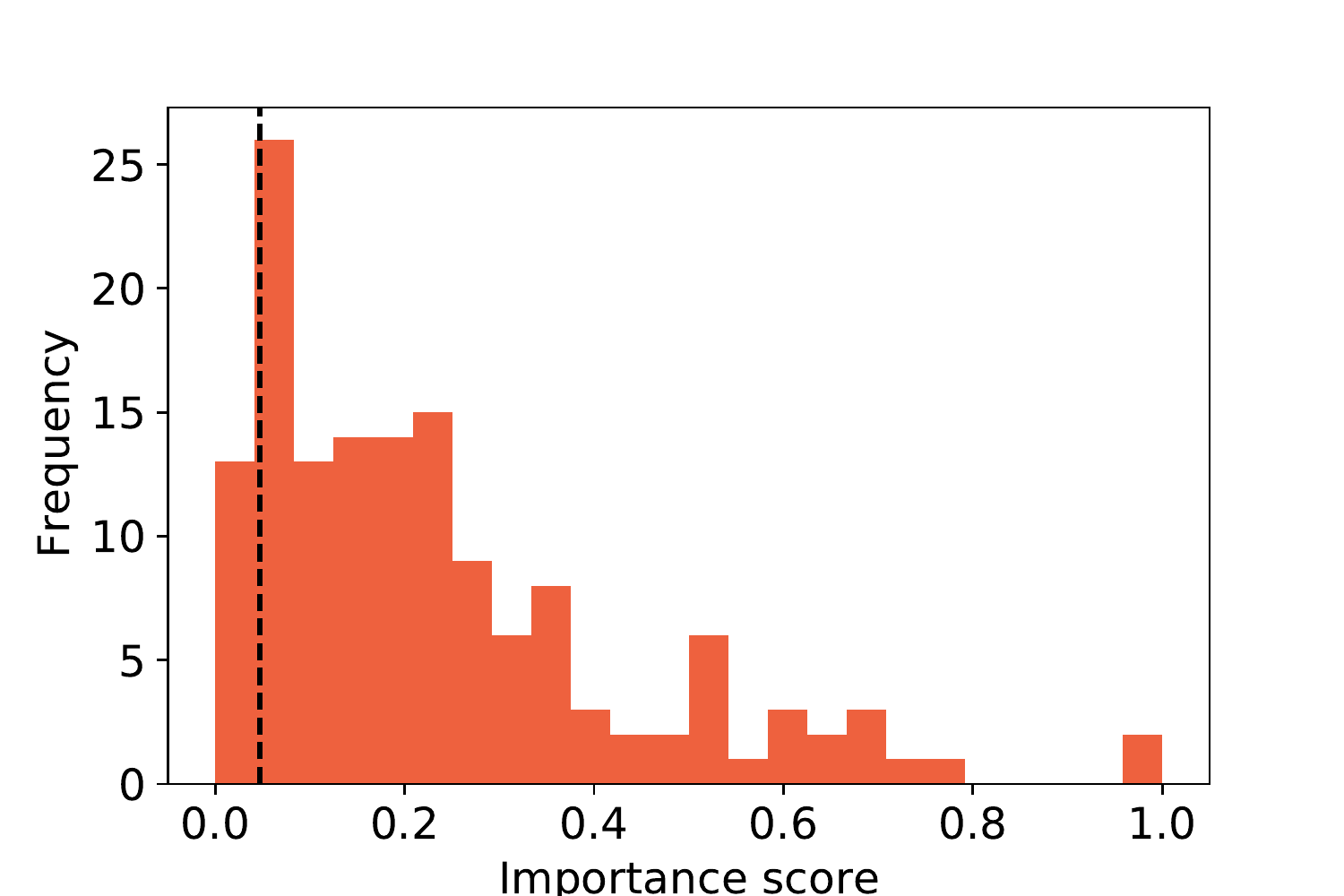}
    \caption{Head importance scores distribution (this example shows CoLA, pruning iteration 1)}
    \label{fig:importance-distribution}
\end{figure}

The distribution of importance scores, shown in \autoref{fig:importance-distribution}, explains why that is the case. At any given pruning iteration, most heads and MLPs have a low importance score, and could all be pruned with about equal success.

\begin{figure*}[!ht]
  \begin{subfigure}[b]{\textwidth}
    \begin{center}
    \includegraphics[clip,scale=0.6]{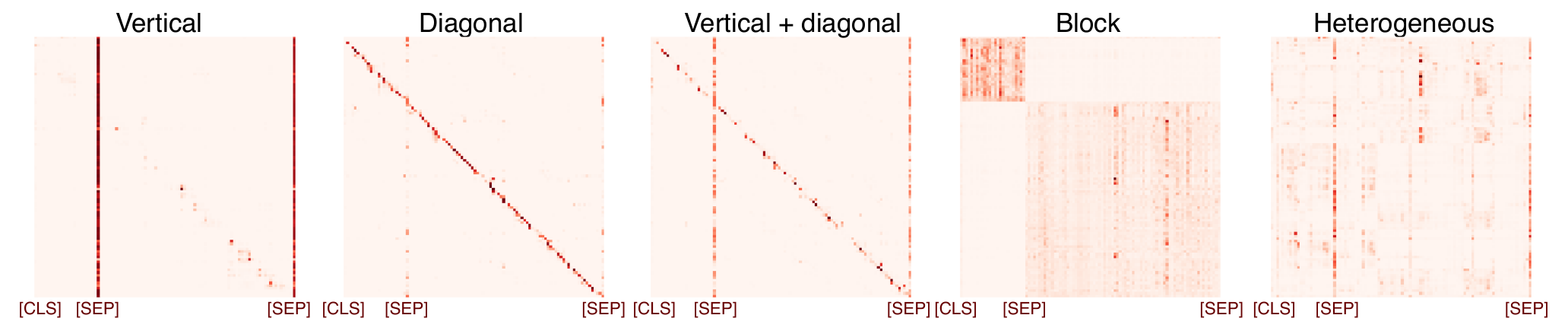}
    \end{center}
    \caption{Reference: typical BERT self-attention patterns by \citet{KovalevaRomanovEtAl_2019_Revealing_Dark_Secrets_of_BERT}.}
    \label{fig:attention-types}
  \end{subfigure}

    \vspace{0.5cm}
    \begin{subfigure}[b]{0.49\textwidth}
        \includegraphics[trim=5 5 5 5,clip,width=\linewidth]{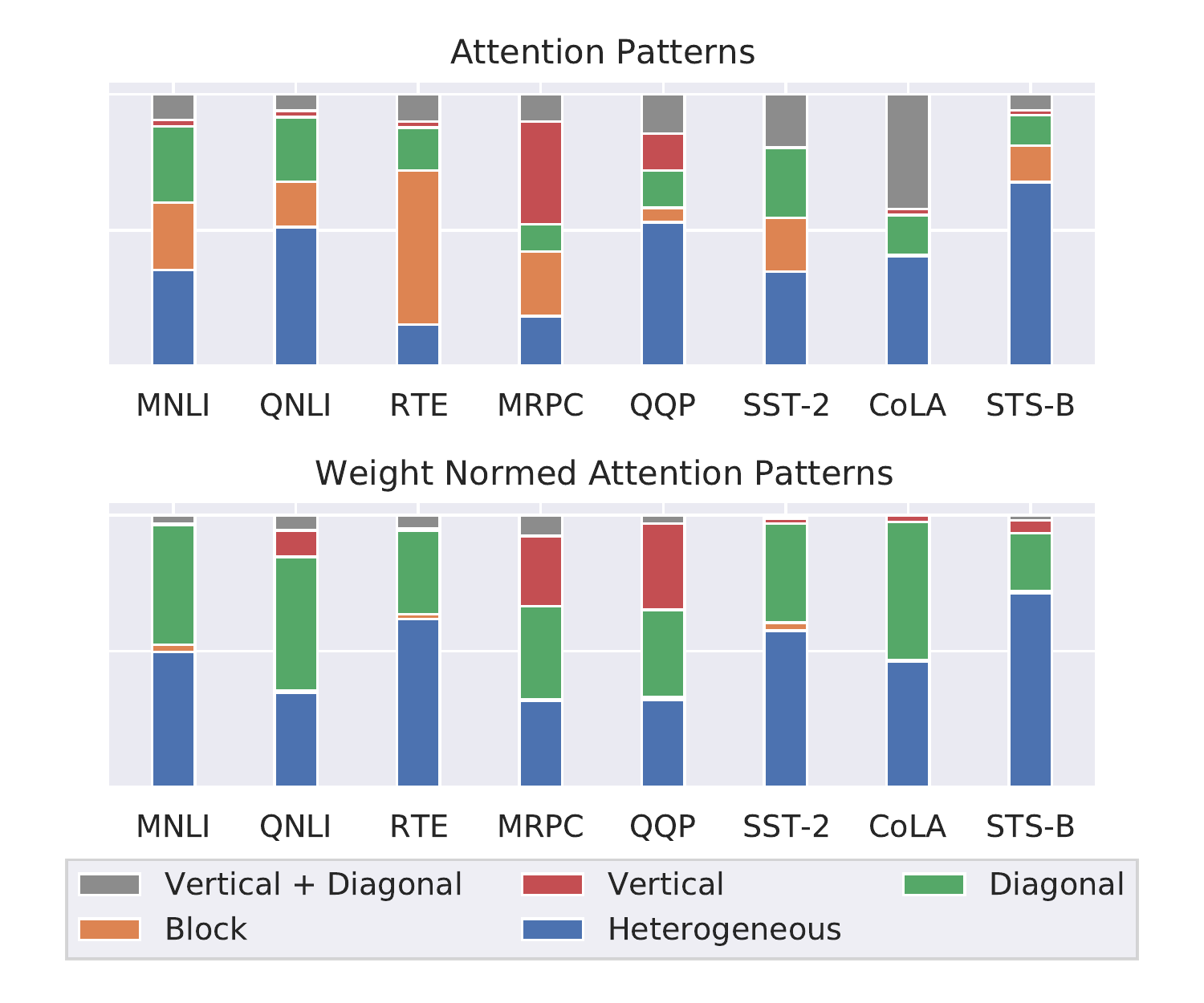}
        \subcaption{Super-survivor heads, fine-tuned}
        \label{fig:attention-types-super-survivors}
    \end{subfigure}
    \hfill
    \begin{subfigure}[b]{0.49\textwidth}
        \includegraphics[trim=5 5 5 5,clip,width=\linewidth]{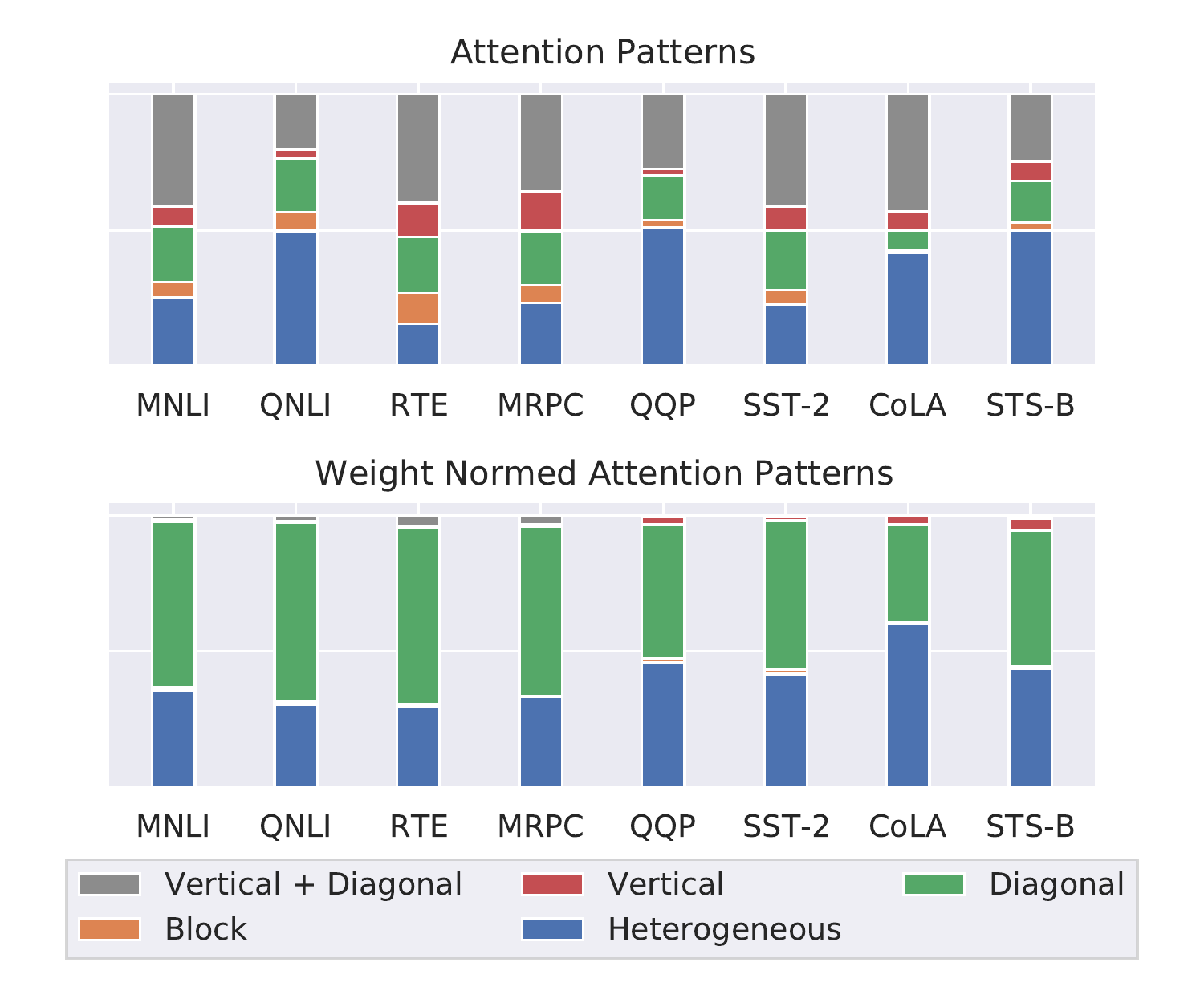}
        \subcaption{All heads, fine-tuned}
        \label{fig:attention-types-all}        
    \end{subfigure}
    \label{fig:attn-patterns}
    \caption{Attention pattern type distribution}
\end{figure*}

\subsection{How Linguistic are the ``Good" Subnetworks?}
\label{sec:linguistic}

A popular method of studying functions of BERT architecture blocks is to use probing classifiers for specific linguistic functions. However, ``the
fact that a linguistic pattern is not observed by our probing classifier does not guarantee that it is not there, and the observation of a pattern does not tell us how it is used" \cite{tenney2019bert}.

In this study we use a cruder, but more reliable alternative: the types of self-attention patterns, which \citet{KovalevaRomanovEtAl_2019_Revealing_Dark_Secrets_of_BERT} classified as diagonal (attention to previous/next word), block (uniform attention over a sentence), vertical (attention to punctuation and special tokens), vertical+diagonal, and heterogeneous (everything else) (see \autoref{fig:attention-types}). The fraction of heterogeneous attention can be used as an \textit{upper bound estimate} on non-trivial linguistic information. In other words, these patterns do not guarantee that a given head has some interpretable function -- only that it could have it.

This analysis is performed by image classification on generated attention maps from individual heads (100 for each GLUE task), for which we use a small CNN classifier with six layers. The classifier was trained on the dataset of 400 annotated attention maps by \citet{KovalevaRomanovEtAl_2019_Revealing_Dark_Secrets_of_BERT}.

Note that attention heads can be seen as a weighted sum of linearly transformed input vectors. \citet{KobayashiKuribayashiEtAl_2020_Attention_Module_is_Not_Only_Weight_Analyzing_Transformers_with_Vector_Norms} recently showed that the input vector norms vary considerably, and the inputs to the self-attention mechanism can have a disproportionate impact relative to their self-attention weight. 
So we consider both the raw attention maps, and, to assess the true impact of the input in the weighted sum, the L2-norm of the transformed input multiplied by the attention weight (for which we annotated 600 more attention maps with the same pattern types as \citet{KovalevaRomanovEtAl_2019_Revealing_Dark_Secrets_of_BERT}). The weighted average of F$_1$ scores
of the classifier on annotated data was 0.81 for the raw attention maps, and 0.74 for the normed attention.

Our results suggest that the super-survivor heads do \textit{not} preferentially encode non-trivial linguistic relations (heterogeneous pattern), in either raw or normed self-attention (\autoref{fig:attention-types-super-survivors}). As compared to all 144 heads (\autoref{fig:attention-types-all}) the ``raw" attention patterns of super-survivors encode considerably more block and vertical attention types. %
Since norming reduces attention to special tokens, the proportion of diagonal patterns (i.e. attention to previous/next tokens) is increased at the cost of vertical+diagonal pattern. Interestingly, for 3 tasks, the super-survivor subnetworks still heavily rely on the vertical pattern even after norming. The vertical pattern indicates a crucial role of the special tokens, and it is unclear why it seems to be less important for MNLI rather than QNLI, MRPC or QQP.

The number of block pattern decreased, and we hypothesize that they are now classified as heterogeneous (as they would be unlikely to look diagonal). But even with the normed attention, the utility of super-survivor heads cannot be attributed only to their linguistic functions (especially given that the fraction of heterogeneous patterns is only a rough upper bound). The Pearson's correlation between heads being super-survivors and their having heterogeneous attention patterns is 0.015 for the raw, and 0.025 for the normed attention. Many ``important" heads have diagonal attention patterns, which seems redundant. %

We conducted the same analysis for the attention patterns in pre-trained vs. fine-tuned BERT for both super-survivors and all heads, and found them to not change considerably after fine-tuning, which is consistent with findings by \citet{KovalevaRomanovEtAl_2019_Revealing_Dark_Secrets_of_BERT}. Full data is available in \autoref{app:attn-patterns-all}.

Note that this result does not exclude the possibility that linguistic information is encoded in certain \textit{combinations} of BERT elements. However, to date most BERT analysis studies focused on the functions of individual components
\cite[see also the overview by \citet{RogersKovalevaEtAl_2020_Primer_in_BERTology_What_we_know_about_how_BERT_works}]{VoitaTalbotEtAl_2019_Analyzing_Multi-Head_Self-Attention_Specialized_Heads_Do_Heavy_Lifting_Rest_Can_Be_Pruneda,htut2019attention,clark2019does,lin2019open,VigBelinkov_2019_Analyzing_Structure_of_Attention_in_Transformer_Language_Model,hewitt2019structural,tenney2019bert}, and this evidence points to the necessity of looking at their interactions. It also adds to the ongoing discussion of interpretability of self-attention \cite{JainWallace_2019_Attention_is_not_Explanation,SerranoSmith_2019_Is_Attention_Interpretable,WiegreffePinter_2019_Attention_is_not_not_Explanation,BrunnerLiuEtAl_2019_On_Identifiability_in_Transformers}. 

Once again, heterogenerous pattern counts are only a crude upper bound estimate on potentially interpretable patterns. More sophisticated alternatives should be explored in future work. For instance, the recent information-theoretic probing by minimum description length \cite{VoitaTitov_2020_Information-Theoretic_Probing_with_Minimum_Description_Length} avoids the problem of false positives with traditional probing classifiers.

\subsection{Information Shared Between Tasks}

\begin{figure}[!t]%
\includegraphics[width=\linewidth,trim=60 0 100 30,clip]{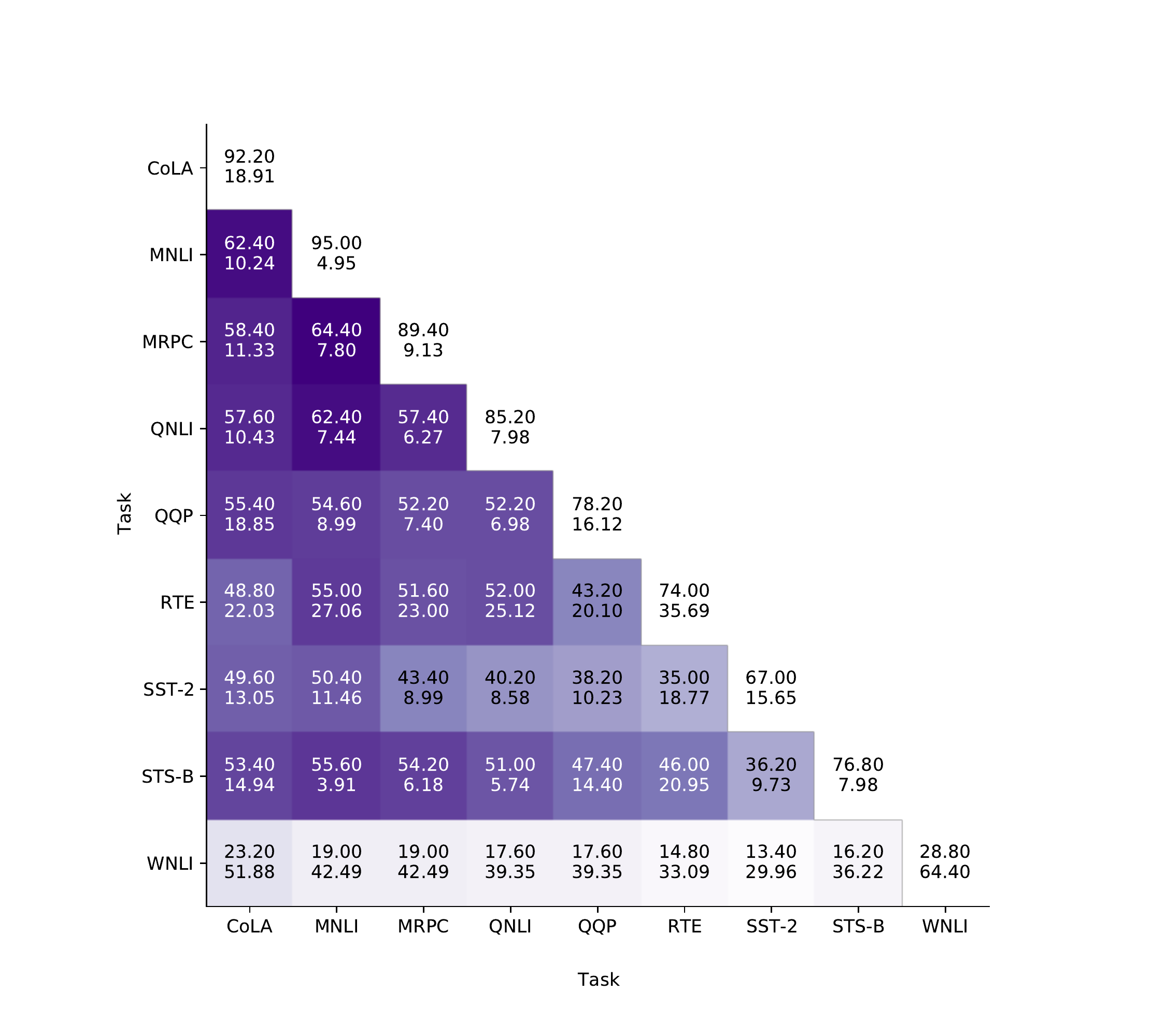}
    \caption{Overlaps in BERT's ``good" subnetworks between GLUE tasks: self-attention heads.}
\label{fig:task-pairs-heads}    
\end{figure}

While %
the ``good" subnetworks are not stable, the overlaps between the ``good" subnetworks may still be used to characterize the tasks themselves. %
We leave detailed exploration to future work, but as a brief illustration,  \autoref{fig:task-pairs-heads} shows pairwise overlaps in the ``good'' subnetworks for the GLUE tasks. 

The overlaps are not particularly large, but still more than what we would expect if the heads were completely independent (e.g. MRPC and QNLI share over a half of their ``good" subnetworks). %
Both heads and MLPs show a similar pattern. Full data for full and super-survivor ``good" subnetworks is available in \autoref{appendix:task-pairs}.

Given our results in \autoref{sec:linguistic}, the overlaps in the ``good" subnetworks are not explainable by two tasks' relying on the same linguistic patterns in individual self-attention heads. They also do not seem to depend on the type of the task. For instance, consider the fact that two tasks targeting paraphrases (MRPC and QQP) have less in common than MRPC and MNLI. %
Alternatively, the overlaps may indicate shared heuristics, or patterns somehow encoded in combinations of BERT elements. This remains to be explored in future work.

\section{Discussion}

This study confirms the main prediction of LTH for pre-trained BERT weights for both m- and s-pruning. An unexpected finding is that with s-pruning, the ``random" subnetworks are still almost as good as the ``good" ones, and even the ``worst" ones perform on par with a strong baseline. This suggests that the weights that do not survive pruning are not just ``inactive" \cite{ZhangBengioEtAl_2019_Are_All_Layers_Created_Equal}.   

An obvious, but very difficult question that arises from this finding is whether the ``bad" subnetworks do well because even they contain some linguistic knowledge, or just because GLUE tasks are overall easy and could be learned even by \textit{random} BERT \cite{KovalevaRomanovEtAl_2019_Revealing_Dark_Secrets_of_BERT}, or even any sufficiently large model. Given that we did not find even the ``good" subnetworks to be stable, or preferentially containing the heads that could have interpretable linguistic functions, the latter seems more likely.

Furthermore, should we perhaps be asking the same question with respect to not only subnetworks, but also full models, such as BERT itself and all the follow-up Transformers? There is a trend to automatically credit any new state-of-the-art model with with better knowledge of language. However, what if that is not the case, and the success of pre-training is rather due to the flatter and wider optima in the optimization surface \cite{HaoDongEtAl_2019_Visualizing_and_Understanding_Effectiveness_of_BERT}? Can similar loss landscapes be obtained from other, non-linguistic pre-training tasks? There are initial results pointing in that direction: \citet{PapadimitriouJurafsky_2020_Learning_Music_Helps_You_Read_Using_Transfer_to_Study_Linguistic_Structure_in_Language_Models} report that even training on MIDI music is helpful for transfer learning for LM task with LSTMs.

\section{Conclusion}

This study systematically tested the lottery ticket hypothesis in BERT fine-tuning with two pruning methods: magnitude-based weight pruning and importance-based pruning of BERT self-attention heads and MLPs.
For both methods, we find that the pruned ``good" subnetworks alone reach the performance comparable with the full model, while the ``bad" ones do not. However, for structured pruning, even the ``bad" subnetworks can be fine-tuned separately to reach fairly strong performance. The ``good" subnetworks are \textit{not} stable across fine-tuning runs, and their success is \textit{not} attributable exclusively to non-trivial linguistic patterns in individual self-attention heads. This suggests that most of pre-trained BERT is potentially useful in fine-tuning, and its success could have more to do with optimization surfaces rather than specific bits of linguistic knowledge.

\begin{spacing}{1}
\footnotesize
\paragraph{Carbon Impact Statement.}  This work contributed 115.644 kg of $\text{CO}_{2eq}$ to the atmosphere and used 249.068 kWh of electricity, having a NLD-specific social cost of carbon of \$-0.14 (\$-0.24, \$-0.04).
The social cost of carbon uses models from \citep{ricke2018country} and this statement and emissions information was generated with \emph{experiment-impact-tracker} \citep{HendersonHuEtAl_2020_Towards_Systematic_Reporting_of_Energy_and_Carbon_Footprints_of_Machine_Learning}.
\end{spacing}
\normalsize

\section*{Acknowledgments}
The authors would like to thank Michael Carbin, Aleksandr Drozd, Jonathan Frankle, Naomi Saphra and Sri Ananda Seelan for their helpful comments, the Zoho corporation for providing access to their clusters for running our experiments, and the anonymous reviewers for their insightful reviews and suggestions. This work is funded in part by the NSF award number IIS-1844740 to Anna Rumshisky.

\bibliography{main}

\begin{thebibliography}{57}
\expandafter\ifx\csname natexlab\endcsname\relax\def\natexlab#1{#1}\fi

\bibitem[{Bentivogli et~al.(2009)Bentivogli, Clark, Dagan, and
  Giampiccolo}]{BentivogliClarkEtAl_2009_Fifth_PASCAL_Recognizing_Textual_Entailment_Challenge}
Luisa Bentivogli, Peter Clark, Ido Dagan, and Danilo Giampiccolo. 2009.
\newblock The {{Fifth PASCAL Recognizing Textual Entailment Challenge}}.
\newblock In \emph{{{TAC}}}.

\bibitem[{Brunner et~al.(2020)Brunner, Liu, Pascual, Richter, Ciaramita, and
  Wattenhofer}]{BrunnerLiuEtAl_2019_On_Identifiability_in_Transformers}
Gino Brunner, Yang Liu, Damian Pascual, Oliver Richter, Massimiliano Ciaramita,
  and Roger Wattenhofer. 2020.
\newblock \href {https://openreview.net/forum?id=BJg1f6EFDB} {On
  {{Identifiability}} in {{Transformers}}}.
\newblock In \emph{International {{Conference}} on {{Learning
  Representations}}}.

\bibitem[{Cer et~al.(2017)Cer, Diab, Agirre, {Lopez-Gazpio}, and
  Specia}]{CerDiabEtAl_2017_SemEval-2017_Task_1_Semantic_Textual_Similarity_Multilingual_and_Crosslingual_Focused_Evaluation}
Daniel Cer, Mona Diab, Eneko Agirre, I{\~n}igo {Lopez-Gazpio}, and Lucia
  Specia. 2017.
\newblock \href {https://doi.org/10.18653/v1/S17-2001} {{{SemEval}}-2017
  {{Task}} 1: {{Semantic Textual Similarity Multilingual}} and {{Crosslingual
  Focused Evaluation}}}.
\newblock In \emph{Proceedings of the 11th {{International Workshop}} on
  {{Semantic Evaluation}} ({{SemEval}}-2017)}, pages 1--14, {Vancouver,
  Canada}. {Association for Computational Linguistics}.

\bibitem[{Chen et~al.(2020)Chen, Frankle, Chang, Liu, Zhang, Wang, and
  Carbin}]{ChenFrankleEtAl_2020_Lottery_Ticket_Hypothesis_for_Pre-trained_BERT_Networks}
Tianlong Chen, Jonathan Frankle, Shiyu Chang, Sijia Liu, Yang Zhang, Zhangyang
  Wang, and Michael Carbin. 2020.
\newblock \href {http://arxiv.org/abs/2007.12223} {The {{Lottery Ticket
  Hypothesis}} for {{Pre}}-trained {{BERT Networks}}}.
\newblock \emph{arXiv:2007.12223 [cs, stat]}.

\bibitem[{Clark et~al.(2019)Clark, Khandelwal, Levy, and
  Manning}]{clark2019does}
Kevin Clark, Urvashi Khandelwal, Omer Levy, and Christopher~D. Manning. 2019.
\newblock \href {https://doi.org/10.18653/v1/W19-4828} {What {{Does BERT Look}}
  at? {{An Analysis}} of {{BERT}}'s {{Attention}}}.
\newblock In \emph{Proceedings of the 2019 {{ACL Workshop BlackboxNLP}}:
  {{Analyzing}} and {{Interpreting Neural Networks}} for {{NLP}}}, pages
  276--286, {Florence, Italy}. {Association for Computational Linguistics}.

\bibitem[{Dagan et~al.(2005)Dagan, Glickman, and
  Magnini}]{DaganGlickmanEtAl_2005_PASCAL_Recognising_Textual_Entailment_Challenge}
Ido Dagan, Oren Glickman, and Bernardo Magnini. 2005.
\newblock The {{PASCAL Recognising Textual Entailment Challenge}}.
\newblock In \emph{Machine {{Learning Challenges Workshop}}}, pages 177--190.
  {Springer}.

\bibitem[{Devlin et~al.(2019)Devlin, Chang, Lee, and
  Toutanova}]{DevlinChangEtAl_2019_BERT_Pre-training_of_Deep_Bidirectional_Transformers_for_Language_Understanding}
Jacob Devlin, Ming-Wei Chang, Kenton Lee, and Kristina Toutanova. 2019.
\newblock \href {https://aclweb.org/anthology/papers/N/N19/N19-1423/}
  {{{BERT}}: {{Pre}}-training of {{Deep Bidirectional Transformers}} for
  {{Language Understanding}}}.
\newblock In \emph{Proceedings of the 2019 {{Conference}} of the {{North
  American Chapter}} of the {{Association}} for {{Computational Linguistics}}:
  {{Human Language Technologies}}, {{Volume}} 1 ({{Long}} and {{Short
  Papers}})}, pages 4171--4186.

\bibitem[{Dodge et~al.(2020)Dodge, Ilharco, Schwartz, Farhadi, Hajishirzi, and
  Smith}]{DodgeIlharcoEtAl_2020_Fine-Tuning_Pretrained_Language_Models_Weight_Initializations_Data_Orders_and_Early_Stopping}
Jesse Dodge, Gabriel Ilharco, Roy Schwartz, Ali Farhadi, Hannaneh Hajishirzi,
  and Noah Smith. 2020.
\newblock \href {http://arxiv.org/abs/2002.06305} {Fine-{{Tuning Pretrained
  Language Models}}: {{Weight Initializations}}, {{Data Orders}}, and {{Early
  Stopping}}}.
\newblock \emph{arXiv:2002.06305 [cs]}.

\bibitem[{Dolan and
  Brockett(2005)}]{DolanBrockett_2005_Automatically_constructing_a_corpus_of_sentential_paraphrases}
W.B. Dolan and C.~Brockett. 2005.
\newblock \href {https://www.aclweb.org/anthology/I/I05/I05-5002.pdf}
  {Automatically constructing a corpus of sentential paraphrases}.
\newblock In \emph{Third {{International Workshop}} on {{Paraphrasing}}
  ({{IWP2005}})}. {Asia Federation of Natural Language Processing}.

\bibitem[{Frankle and
  Carbin(2019)}]{FrankleCarbin_2019_Lottery_Ticket_Hypothesis_Finding_Sparse_Trainable_Neural_Networks}
Jonathan Frankle and Michael Carbin. 2019.
\newblock \href {https://openreview.net/forum?id=rJl-b3RcF7} {The {{Lottery
  Ticket Hypothesis}}: {{Finding Sparse}}, {{Trainable Neural Networks}}}.
\newblock In \emph{International {{Conference}} on {{Learning
  Representations}}}.

\bibitem[{Giampiccolo et~al.(2007)Giampiccolo, Magnini, Dagan, and
  Dolan}]{GiampiccoloMagniniEtAl_2007_Third_PASCAL_Recognizing_Textual_Entailment_Challenge}
Danilo Giampiccolo, Bernardo Magnini, Ido Dagan, and Bill Dolan. 2007.
\newblock The {{Third PASCAL Recognizing Textual Entailment Challenge}}.
\newblock In \emph{Proceedings of the {{ACL}}-{{PASCAL Workshop}} on {{Textual
  Entailment}} and {{Paraphrasing}}}, {{RTE}} '07, pages 1--9, {Prague, Czech
  Republic}. {Association for Computational Linguistics}.

\bibitem[{Gordon et~al.(2020)Gordon, Duh, and
  Andrews}]{Gordon2020CompressingBS}
Mitchell~A. Gordon, Kevin Duh, and Nicholas Andrews. 2020.
\newblock Compressing {{BERT}}: Studying the effects of weight pruning on
  transfer learning.
\newblock \emph{ArXiv}, abs/2002.08307.

\bibitem[{Haim et~al.(2006)Haim, Dagan, Dolan, Ferro, Giampiccolo, Magnini, and
  Szpektor}]{HaimDaganEtAl_2006_Second_Pascal_Recognising_Textual_Entailment_Challenge}
R.~Bar Haim, Ido Dagan, Bill Dolan, Lisa Ferro, Danilo Giampiccolo, Bernardo
  Magnini, and Idan Szpektor. 2006.
\newblock The {{Second Pascal Recognising Textual Entailment Challenge}}.
\newblock In \emph{Proceedings of the {{Second PASCAL Challenges Workshop}} on
  {{Recognising Textual Entailment}}}.

\bibitem[{Hao et~al.(2019)Hao, Dong, Wei, and
  Xu}]{HaoDongEtAl_2019_Visualizing_and_Understanding_Effectiveness_of_BERT}
Yaru Hao, Li~Dong, Furu Wei, and Ke~Xu. 2019.
\newblock \href {https://doi.org/10.18653/v1/D19-1424} {Visualizing and
  {{Understanding}} the {{Effectiveness}} of {{BERT}}}.
\newblock In \emph{Proceedings of the 2019 {{Conference}} on {{Empirical
  Methods}} in {{Natural Language Processing}} and the 9th {{International
  Joint Conference}} on {{Natural Language Processing}}
  ({{EMNLP}}-{{IJCNLP}})}, pages 4143--4152, {Hong Kong, China}. {Association
  for Computational Linguistics}.

\bibitem[{Henderson et~al.(2020)Henderson, Hu, Romoff, Brunskill, Jurafsky, and
  Pineau}]{HendersonHuEtAl_2020_Towards_Systematic_Reporting_of_Energy_and_Carbon_Footprints_of_Machine_Learning}
Peter Henderson, Jieru Hu, Joshua Romoff, Emma Brunskill, Dan Jurafsky, and
  Joelle Pineau. 2020.
\newblock \href {http://arxiv.org/abs/2002.05651} {Towards the {{Systematic
  Reporting}} of the {{Energy}} and {{Carbon Footprints}} of {{Machine
  Learning}}}.
\newblock \emph{arXiv:2002.05651 [cs]}.

\bibitem[{Hewitt and Manning(2019)}]{hewitt2019structural}
John Hewitt and Christopher~D. Manning. 2019.
\newblock \href {https://aclweb.org/anthology/papers/N/N19/N19-1419/} {A
  {{Structural Probe}} for {{Finding Syntax}} in {{Word Representations}}}.
\newblock In \emph{Proceedings of the 2019 {{Conference}} of the {{North
  American Chapter}} of the {{Association}} for {{Computational Linguistics}}:
  {{Human Language Technologies}}, {{Volume}} 1 ({{Long}} and {{Short
  Papers}})}, pages 4129--4138.

\bibitem[{Htut et~al.(2019)Htut, Phang, Bordia, and Bowman}]{htut2019attention}
Phu~Mon Htut, Jason Phang, Shikha Bordia, and Samuel~R Bowman. 2019.
\newblock \href {http://arxiv.org/abs/1911.12246} {Do attention heads in
  {{BERT}} track syntactic dependencies?}
\newblock \emph{arXiv preprint arXiv:1911.12246}.

\bibitem[{Jain and
  Wallace(2019)}]{JainWallace_2019_Attention_is_not_Explanation}
Sarthak Jain and Byron~C. Wallace. 2019.
\newblock \href {https://aclweb.org/anthology/papers/N/N19/N19-1357/}
  {Attention is not {{Explanation}}}.
\newblock In \emph{Proceedings of the 2019 {{Conference}} of the {{North
  American Chapter}} of the {{Association}} for {{Computational Linguistics}}:
  {{Human Language Technologies}}, {{Volume}} 1 ({{Long}} and {{Short
  Papers}})}, pages 3543--3556.

\bibitem[{Jiao et~al.(2019)Jiao, Yin, Shang, Jiang, Chen, Li, Wang, and
  Liu}]{jiao2019tinybert}
Xiaoqi Jiao, Yichun Yin, Lifeng Shang, Xin Jiang, Xiao Chen, Linlin Li, Fang
  Wang, and Qun Liu. 2019.
\newblock \href {https://arxiv.org/abs/1909.10351} {{{TinyBERT: Distilling BERT
  for natural language understanding}}}.
\newblock \emph{arXiv preprint arXiv:1909.10351}.

\bibitem[{Jin et~al.(2020)Jin, Jin, Zhou, and
  Szolovits}]{JinJinEtAl_2020_Is_BERT_Really_Robust_Strong_Baseline_for_Natural_Language_Attack_on_Text_Classification_and_Entailment}
Di~Jin, Zhijing Jin, Joey~Tianyi Zhou, and Peter Szolovits. 2020.
\newblock \href {http://arxiv.org/abs/1907.11932} {Is {{BERT Really Robust}}?
  {{A Strong Baseline}} for {{Natural Language Attack}} on {{Text
  Classification}} and {{Entailment}}}.
\newblock In \emph{{{AAAI}} 2020}.

\bibitem[{Kobayashi et~al.(2020)Kobayashi, Kuribayashi, Yokoi, and
  Inui}]{KobayashiKuribayashiEtAl_2020_Attention_Module_is_Not_Only_Weight_Analyzing_Transformers_with_Vector_Norms}
Goro Kobayashi, Tatsuki Kuribayashi, Sho Yokoi, and Kentaro Inui. 2020.
\newblock \href {http://arxiv.org/abs/2004.10102} {Attention {{Module}} is
  {{Not Only}} a {{Weight}}: {{Analyzing Transformers}} with {{Vector Norms}}}.
\newblock \emph{arXiv:2004.10102 [cs]}.

\bibitem[{Kovaleva et~al.(2019)Kovaleva, Romanov, Rogers, and
  Rumshisky}]{KovalevaRomanovEtAl_2019_Revealing_Dark_Secrets_of_BERT}
Olga Kovaleva, Alexey Romanov, Anna Rogers, and Anna Rumshisky. 2019.
\newblock \href {https://doi.org/10.18653/v1/D19-1445} {Revealing the {{Dark
  Secrets}} of {{BERT}}}.
\newblock In \emph{Proceedings of the 2019 {{Conference}} on {{Empirical
  Methods}} in {{Natural Language Processing}} and the 9th {{International
  Joint Conference}} on {{Natural Language Processing}}
  ({{EMNLP}}-{{IJCNLP}})}, pages 4356--4365, {Hong Kong, China}. {Association
  for Computational Linguistics}.

\bibitem[{Lan et~al.(2020)Lan, Chen, Goodman, Gimpel, Sharma, and
  Soricut}]{model:albert}
Zhenzhong Lan, Mingda Chen, Sebastian Goodman, Kevin Gimpel, Piyush Sharma, and
  Radu Soricut. 2020.
\newblock \href {https://openreview.net/forum?id=H1eA7AEtvS} {{{ALBERT}}: {{A
  Lite BERT}} for {{Self}}-{{Supervised}} {{Learning}} of {{Language
  Representations}}}.
\newblock In \emph{{ICLR}}.

\bibitem[{Lee et~al.(2018)Lee, Ajanthan, and
  Torr}]{LeeAjanthanEtAl_2018_SNIP_Single-shot_network_pruning_based_on_connection_sensitivity}
Namhoon Lee, Thalaiyasingam Ajanthan, and Philip Torr. 2018.
\newblock \href {https://openreview.net/forum?id=B1VZqjAcYX} {{{SNIP}}:
  {{Single}}-shot network pruning based on connection sensitivity}.
\newblock In \emph{International {{Conference}} on {{Learning
  Representations}}}.

\bibitem[{Levesque et~al.(2012)Levesque, Davis, and
  Morgenstern}]{LevesqueDavisEtAl_2012_Winograd_Schema_Challenge}
Hector~J Levesque, Ernest Davis, and Leora Morgenstern. 2012.
\newblock The {{Winograd Schema Challenge}}.
\newblock In \emph{Proceedings of the {{Thirteenth International Conference}}
  on {{Principles}} of {{Knowledge Representation}} and {{Reasoning}}}, pages
  552--561.

\bibitem[{Lin et~al.(2019)Lin, Tan, and Frank}]{lin2019open}
Yongjie Lin, Yi~Chern Tan, and Robert Frank. 2019.
\newblock \href {https://www.aclweb.org/anthology/W19-4825/} {{{Open Sesame:
  Getting inside BERT's Linguistic Knowledge}}}.
\newblock In \emph{Proceedings of the 2019 ACL Workshop BlackboxNLP: Analyzing
  and Interpreting Neural Networks for NLP}, pages 241--253.

\bibitem[{Liu et~al.(2019)Liu, Gardner, Belinkov, Peters, and
  Smith}]{LiuGardnerEtAl_2019_Linguistic_Knowledge_and_Transferability_of_Contextual_Representations}
Nelson~F. Liu, Matt Gardner, Yonatan Belinkov, Matthew~E. Peters, and Noah~A.
  Smith. 2019.
\newblock \href {https://www.aclweb.org/anthology/N19-1112/} {Linguistic
  {{Knowledge}} and {{Transferability}} of {{Contextual Representations}}}.
\newblock In \emph{Proceedings of the 2019 {{Conference}} of the {{North
  American Chapter}} of the {{Association}} for {{Computational Linguistics}}:
  {{Human Language Technologies}}, {{Volume}} 1 ({{Long}} and {{Short
  Papers}})}, pages 1073--1094, {Minneapolis, Minnesota}. {Association for
  Computational Linguistics}.

\bibitem[{McCarley(2019)}]{mccarley2019pruning}
JS~McCarley. 2019.
\newblock \href {https://arxiv.org/abs/1910.06360} {{{Pruning a BERT-based
  Question Answering Model}}}.
\newblock \emph{arXiv preprint arXiv:1910.06360}.

\bibitem[{McCoy et~al.(2019{\natexlab{a}})McCoy, Min, and
  Linzen}]{McCoy2019BERTsOA}
R.~Thomas McCoy, Junghyun Min, and Tal Linzen. 2019{\natexlab{a}}.
\newblock {{BERTs}} of a feather do not generalize together: Large variability
  in generalization across models with similar test set performance.
\newblock \emph{ArXiv}, abs/1911.02969.

\bibitem[{McCoy et~al.(2019{\natexlab{b}})McCoy, Pavlick, and
  Linzen}]{McCoyPavlickEtAl_2019_Right_for_Wrong_Reasons_Diagnosing_Syntactic_Heuristics_in_Natural_Language_Inference}
Tom McCoy, Ellie Pavlick, and Tal Linzen. 2019{\natexlab{b}}.
\newblock \href {https://doi.org/10.18653/v1/P19-1334} {Right for the {{Wrong
  Reasons}}: {{Diagnosing Syntactic Heuristics}} in {{Natural Language
  Inference}}}.
\newblock In \emph{Proceedings of the 57th {{Annual Meeting}} of the
  {{Association}} for {{Computational Linguistics}}}, pages 3428--3448,
  {Florence, Italy}. {Association for Computational Linguistics}.

\bibitem[{Michel et~al.(2019)Michel, Levy, and
  Neubig}]{MichelLevyEtAl_2019_Are_Sixteen_Heads_Really_Better_than_One}
Paul Michel, Omer Levy, and Graham Neubig. 2019.
\newblock \href
  {http://papers.nips.cc/paper/9551-are-sixteen-heads-really-better-than-one}
  {Are {{Sixteen Heads Really Better}} than {{One}}?}
\newblock \emph{Advances in Neural Information Processing Systems 32 (NIPS
  2019)}.

\bibitem[{Molchanov et~al.(2017)Molchanov, Tyree, Karras, Aila, and
  Kautz}]{DBLP:conf/iclr/MolchanovTKAK17}
Pavlo Molchanov, Stephen Tyree, Tero Karras, Timo Aila, and Jan Kautz. 2017.
\newblock \href {https://openreview.net/forum?id=SJGCiw5gl} {Pruning
  convolutional neural networks for resource efficient inference}.
\newblock In \emph{5th International Conference on Learning Representations,
  {ICLR} 2017, Toulon, France, April 24-26, 2017, Conference Track
  Proceedings}. OpenReview.net.

\bibitem[{Morcos et~al.(2019)Morcos, Yu, Paganini, and
  Tian}]{MorcosYuEtAl_2019_One_ticket_to_win_them_all_generalizing_lottery_ticket_initializations_across_datasets_and_optimizers}
Ari Morcos, Haonan Yu, Michela Paganini, and Yuandong Tian. 2019.
\newblock \href
  {http://papers.nips.cc/paper/8739-one-ticket-to-win-them-all-generalizing-lottery-ticket-initializations-across-datasets-and-optimizers.pdf}
  {One ticket to win them all: Generalizing lottery ticket initializations
  across datasets and optimizers}.
\newblock In \emph{Advances in {{Neural Information Processing Systems}} 32},
  pages 4932--4942. {Curran Associates, Inc.}

\bibitem[{Niven and
  Kao(2019)}]{NivenKao_2019_Probing_Neural_Network_Comprehension_of_Natural_Language_Arguments}
Timothy Niven and Hung-Yu Kao. 2019.
\newblock \href {https://www.aclweb.org/anthology/P19-1459} {Probing {{Neural
  Network Comprehension}} of {{Natural Language Arguments}}}.
\newblock In \emph{Proceedings of the 57th {{Annual Meeting}} of the
  {{Association}} for {{Computational Linguistics}}}, pages 4658--4664,
  {Florence, Italy}. {Association for Computational Linguistics}.

\bibitem[{Papadimitriou and
  Jurafsky(2020)}]{PapadimitriouJurafsky_2020_Learning_Music_Helps_You_Read_Using_Transfer_to_Study_Linguistic_Structure_in_Language_Models}
Isabel Papadimitriou and Dan Jurafsky. 2020.
\newblock \href {http://arxiv.org/abs/2004.14601} {Learning {{Music Helps You
  Read}}: {{Using Transfer}} to {{Study Linguistic Structure}} in {{Language
  Models}}}.
\newblock \emph{arXiv:2004.14601 [cs]}.

\bibitem[{Rajpurkar et~al.(2016)Rajpurkar, Zhang, Lopyrev, and
  Liang}]{RajpurkarZhangEtAl_2016_SquAD_100000+_Questions_for_Machine_Comprehension_of_Text}
Pranav Rajpurkar, Jian Zhang, Konstantin Lopyrev, and Percy Liang. 2016.
\newblock \href {https://doi.org/10.18653/v1/D16-1264} {{{SQuAD}}: 100,000+
  {{Questions}} for {{Machine Comprehension}} of {{Text}}}.
\newblock In \emph{Proceedings of the 2016 {{Conference}} on {{Empirical
  Methods}} in {{Natural Language Processing}}}, pages 2383--2392. {Association
  for Computational Linguistics}.

\bibitem[{Ricke et~al.(2018)Ricke, Drouet, Caldeira, and
  Tavoni}]{ricke2018country}
Katharine Ricke, Laurent Drouet, Ken Caldeira, and Massimo Tavoni. 2018.
\newblock Country-level social cost of carbon.
\newblock \emph{Nature Climate Change}, 8(10):895.

\bibitem[{Rogers et~al.(2020{\natexlab{a}})Rogers, Kovaleva, Downey, and
  Rumshisky}]{RogersKovalevaEtAl_2020_Getting_Closer_to_AI_Complete_Question_Answering_Set_of_Prerequisite_Real_Tasks}
Anna Rogers, Olga Kovaleva, Matthew Downey, and Anna Rumshisky.
  2020{\natexlab{a}}.
\newblock \href {https://aaai.org/Papers/AAAI/2020GB/AAAI-RogersA.7778.pdf}
  {Getting {{Closer}} to {{AI Complete Question Answering}}: {{A Set}} of
  {{Prerequisite Real Tasks}}}.
\newblock In \emph{{{AAAI}}}, page~11.

\bibitem[{Rogers et~al.(2020{\natexlab{b}})Rogers, Kovaleva, and
  Rumshisky}]{RogersKovalevaEtAl_2020_Primer_in_BERTology_What_we_know_about_how_BERT_works}
Anna Rogers, Olga Kovaleva, and Anna Rumshisky. 2020{\natexlab{b}}.
\newblock \href {http://arxiv.org/abs/2002.12327} {A {{Primer}} in
  {{BERTology}}: {{What}} we know about how {{BERT}} works}.
\newblock \emph{arXiv:2002.12327 [cs]}.

\bibitem[{Sanh et~al.(2019)Sanh, Debut, Chaumond, and Wolf}]{model:distilBERT}
Victor Sanh, Lysandre Debut, Julien Chaumond, and Thomas Wolf. 2019.
\newblock \href {http://arxiv.org/abs/1910.01108} {{{DistilBERT}}, a distilled
  version of {{BERT}}: Smaller, faster, cheaper and lighter}.
\newblock In \emph{5th {{Workshop}} on {{Energy Efficient Machine Learning}}
  and {{Cognitive Computing}} - {{NeurIPS}} 2019}.

\bibitem[{Serrano and
  Smith(2019)}]{SerranoSmith_2019_Is_Attention_Interpretable}
Sofia Serrano and Noah~A. Smith. 2019.
\newblock \href {http://arxiv.org/abs/1906.03731} {Is {{Attention
  Interpretable}}?}
\newblock \emph{arXiv:1906.03731 [cs]}.

\bibitem[{Socher et~al.(2013)Socher, Perelygin, Wu, Chuang, Manning, Ng, and
  Potts}]{SocherPerelyginEtAl_2013_Recursive_deep_models_for_semantic_compositionality_over_sentiment_treebank}
Richard Socher, Alex Perelygin, Jean Wu, Jason Chuang, Christopher~D. Manning,
  Andrew Ng, and Christopher Potts. 2013.
\newblock \href {http://www.aclweb.org/anthology/D13-1170} {Recursive deep
  models for semantic compositionality over a sentiment treebank}.
\newblock In \emph{Proceedings of the 2013 {{Conference}} on {{Empirical
  Methods}} in {{Natural Language Processing}}}, pages 1631--1642.

\bibitem[{Srivastava et~al.(2014)Srivastava, Hinton, Krizhevsky, Sutskever, and
  Salakhutdinov}]{Srivastava2014DropoutAS}
Nitish Srivastava, Geoffrey~E. Hinton, Alex Krizhevsky, Ilya Sutskever, and
  Ruslan Salakhutdinov. 2014.
\newblock Dropout: a simple way to prevent neural networks from overfitting.
\newblock \emph{J. Mach. Learn. Res.}, 15:1929--1958.

\bibitem[{Tenney et~al.(2019)Tenney, Das, and Pavlick}]{tenney2019bert}
Ian Tenney, Dipanjan Das, and Ellie Pavlick. 2019.
\newblock \href {https://www.aclweb.org/anthology/P19-1452.pdf} {{{BERT
  Rediscovers the Classical NLP Pipeline}}}.
\newblock In \emph{Proceedings of the 57th Annual Meeting of the Association
  for Computational Linguistics}, pages 4593--4601.

\bibitem[{Vaswani et~al.(2017)Vaswani, Shazeer, Parmar, Uszkoreit, Jones,
  Gomez, Kaiser, and
  Polosukhin}]{VaswaniShazeerEtAl_2017_Attention_is_all_you_need}
Ashish Vaswani, Noam Shazeer, Niki Parmar, Jakob Uszkoreit, Llion Jones,
  Aidan~N. Gomez, \textbackslash{}Lukasz Kaiser, and Illia Polosukhin. 2017.
\newblock Attention is all you need.
\newblock In \emph{{{NIPS}}}, pages 5998--6008, {Long Beach, CA, USA}.

\bibitem[{Vig and
  Belinkov(2019)}]{VigBelinkov_2019_Analyzing_Structure_of_Attention_in_Transformer_Language_Model}
Jesse Vig and Yonatan Belinkov. 2019.
\newblock \href {https://doi.org/10.18653/v1/W19-4808} {Analyzing the
  {{Structure}} of {{Attention}} in a {{Transformer Language Model}}}.
\newblock In \emph{Proceedings of the 2019 {{ACL Workshop BlackboxNLP}}:
  {{Analyzing}} and {{Interpreting Neural Networks}} for {{NLP}}}, pages
  63--76, {Florence, Italy}. {Association for Computational Linguistics}.

\bibitem[{Voita et~al.(2019)Voita, Talbot, Moiseev, Sennrich, and
  Titov}]{VoitaTalbotEtAl_2019_Analyzing_Multi-Head_Self-Attention_Specialized_Heads_Do_Heavy_Lifting_Rest_Can_Be_Pruneda}
Elena Voita, David Talbot, Fedor Moiseev, Rico Sennrich, and Ivan Titov. 2019.
\newblock \href {https://doi.org/10.18653/v1/P19-1580} {Analyzing
  {{Multi}}-{{Head Self}}-{{Attention}}: {{Specialized Heads Do}} the {{Heavy
  Lifting}}, the {{Rest Can Be Pruned}}}.
\newblock In \emph{Proceedings of the 57th {{Annual Meeting}} of the
  {{Association}} for {{Computational Linguistics}}}, pages 5797--5808,
  {Florence, Italy}. {Association for Computational Linguistics}.

\bibitem[{Voita and
  Titov(2020)}]{VoitaTitov_2020_Information-Theoretic_Probing_with_Minimum_Description_Length}
Elena Voita and Ivan Titov. 2020.
\newblock \href {http://arxiv.org/abs/2003.12298} {Information-{{Theoretic
  Probing}} with {{Minimum Description Length}}}.
\newblock \emph{arXiv:2003.12298 [cs]}.

\bibitem[{Wang et~al.(2018)Wang, Singh, Michael, Hill, Levy, and
  Bowman}]{WangSinghEtAl_2018_GLUE_A_Multi-Task_Benchmark_and_Analysis_Platform_for_Natural_Language_Understanding}
Alex Wang, Amapreet Singh, Julian Michael, Felix Hill, Omer Levy, and Samuel~R.
  Bowman. 2018.
\newblock \href {http://aclweb.org/anthology/W18-5446} {{{GLUE}}: {{A
  Multi}}-{{Task Benchmark}} and {{Analysis Platform}} for {{Natural Language
  Understanding}}}.
\newblock In \emph{Proceedings of the 2018 {{EMNLP Workshop BlackboxNLP}}:
  {{Analyzing}} and {{Interpreting Neural Networks}} for {{NLP}}}, pages
  353--355, {Brussels, Belgium}. {Association for Computational Linguistics}.

\bibitem[{Warstadt et~al.(2019)Warstadt, Singh, and
  Bowman}]{WarstadtSinghEtAl_2019_Neural_Network_Acceptability_Judgments}
Alex Warstadt, Amanpreet Singh, and Samuel~R. Bowman. 2019.
\newblock \href {https://doi.org/10.1162/tacl_a_00290} {Neural {{Network
  Acceptability Judgments}}}.
\newblock \emph{Transactions of the Association for Computational Linguistics},
  7:625--641.

\bibitem[{Wiegreffe and
  Pinter(2019)}]{WiegreffePinter_2019_Attention_is_not_not_Explanation}
Sarah Wiegreffe and Yuval Pinter. 2019.
\newblock \href {https://doi.org/10.18653/v1/D19-1002} {Attention is not not
  {{Explanation}}}.
\newblock In \emph{Proceedings of the 2019 {{Conference}} on {{Empirical
  Methods}} in {{Natural Language Processing}} and the 9th {{International
  Joint Conference}} on {{Natural Language Processing}}
  ({{EMNLP}}-{{IJCNLP}})}, pages 11--20, {Hong Kong, China}. {Association for
  Computational Linguistics}.

\bibitem[{Williams et~al.(2017)Williams, Nangia, and
  Bowman}]{WilliamsNangiaEtAl_2017_A_Broad-Coverage_Challenge_Corpus_for_Sentence_Understanding_through_Inference}
Adina Williams, Nikita Nangia, and Samuel~R. Bowman. 2017.
\newblock \href {https://doi.org/10.18653/v1/N18-1101} {A {{Broad}}-{{Coverage
  Challenge Corpus}} for {{Sentence Understanding}} through {{Inference}}}.
\newblock In \emph{Proceedings of the 2018 {{Conference}} of the {{North
  American Chapter}} of the {{Association}} for {{Computational Linguistics}}:
  {{Human Language Technologies}}, {{Volume}} 1 ({{Long Papers}})}, pages
  1112--1122, {New Orleans, Louisiana}. {Association for Computational
  Linguistics}.

\bibitem[{Wolf et~al.(2020)Wolf, Debut, Sanh, Chaumond, Delangue, Moi, Cistac,
  Rault, Louf, Funtowicz, and
  Brew}]{WolfDebutEtAl_2020_HuggingFaces_Transformers_State-of-the-art_Natural_Language_Processing}
Thomas Wolf, Lysandre Debut, Victor Sanh, Julien Chaumond, Clement Delangue,
  Anthony Moi, Pierric Cistac, Tim Rault, R{\'e}mi Louf, Morgan Funtowicz, and
  Jamie Brew. 2020.
\newblock \href {http://arxiv.org/abs/1910.03771} {{{HuggingFace}}'s
  {{Transformers}}: {{State}}-of-the-art {{Natural Language Processing}}}.
\newblock \emph{arXiv:1910.03771 [cs]}.

\bibitem[{Yu et~al.(2020)Yu, Edunov, Tian, and
  Morcos}]{YuEdunovEtAl_2020_Playing_lottery_with_rewards_and_multiple_languages_lottery_tickets_in_RL_and_NLP}
Haonan Yu, Sergey Edunov, Yuandong Tian, and Ari~S. Morcos. 2020.
\newblock \href {http://arxiv.org/abs/1906.02768} {Playing the lottery with
  rewards and multiple languages: Lottery tickets in {{RL}} and {{NLP}}}.
\newblock In \emph{{{ICLR}}}.

\bibitem[{Zellers et~al.(2019)Zellers, Holtzman, Bisk, Farhadi, and
  Choi}]{ZellersHoltzmanEtAl_2019_HellaSwag_Can_Machine_Really_Finish_Your_Sentence}
Rowan Zellers, Ari Holtzman, Yonatan Bisk, Ali Farhadi, and Yejin Choi. 2019.
\newblock \href {http://arxiv.org/abs/1905.07830} {{{HellaSwag}}: {{Can}} a
  {{Machine Really Finish Your Sentence}}?}
\newblock In \emph{{{ACL}} 2019}.

\bibitem[{Zhang et~al.(2019)Zhang, Bengio, and
  Singer}]{ZhangBengioEtAl_2019_Are_All_Layers_Created_Equal}
Chiyuan Zhang, Samy Bengio, and Yoram Singer. 2019.
\newblock \href {https://openreview.net/forum?id=ryg1P4Sh2E} {Are {{All Layers
  Created Equal}}?}
\newblock In \emph{{{ICML}} 2019 {{Workshop Deep Phenomena}}}.

\bibitem[{Zhou et~al.(2019)Zhou, Lan, Liu, and
  Yosinski}]{ZhouLanEtAl_2019_Deconstructing_Lottery_Tickets_Zeros_Signs_and_Supermask}
Hattie Zhou, Janice Lan, Rosanne Liu, and Jason Yosinski. 2019.
\newblock \href
  {http://papers.nips.cc/paper/8618-deconstructing-lottery-tickets-zeros-signs-and-the-supermask.pdf}
  {Deconstructing {{Lottery Tickets}}: {{Zeros}}, {{Signs}}, and the
  {{Supermask}}}.
\newblock In \emph{Advances in {{Neural Information Processing Systems}} 32},
  pages 3597--3607. {Curran Associates, Inc.}

\end{thebibliography}
\bibliographystyle{acl_natbib}

\clearpage

\appendix

\begin{figure*}
\section{``Good" Subnetworks in BERT Fine-tuned on GLUE Tasks}
\label{appendix:good-subnets-per-task}
\begin{multicols}{2}
        Each figure in this section shows the ``good" subnetwork of heads and layers that survived the pruning process described in \autoref{sec:methodology}. Each task was run with 5 different random seeds. The top number in each cell indicates how likely a given head or MLP was to survive pruning, with 1.0 indicating that it survived on every run. The bottom number indicates the standard deviation across runs. 
        
        The figures in this appendix show that each task has a varying number of heads and layers that survive pruning on all fine-tuning runs, while some heads and layers were only ``picked up" by some random seeds. Note also that in addition to the architecture elements that survive across many runs, there are also some that are useful for over half of the tasks, as shown in \autoref{fig:head-heatmap}, and some \textit{always} survive the pruning. %
        
        Visualizing the ``good" subnetwork illustrates the core problem with WNLI, the most difficult task of GLUE. \autoref{fig:wnli} shows that each run is completely different, indicating that BERT fails to find any consistent pattern between the task and the information in the available pre-trained weights. %
WNLI is described as ``somewhat adversarial" by \citet{WangSinghEtAl_2018_GLUE_A_Multi-Task_Benchmark_and_Analysis_Platform_for_Natural_Language_Understanding} because it has similar sentences in train and dev sets with opposite labels. %
        
\end{multicols}
        
        \begin{subfigure}[b]{0.5\textwidth}
            \includegraphics[trim=-20 -10 -30 30,clip,width=\linewidth]{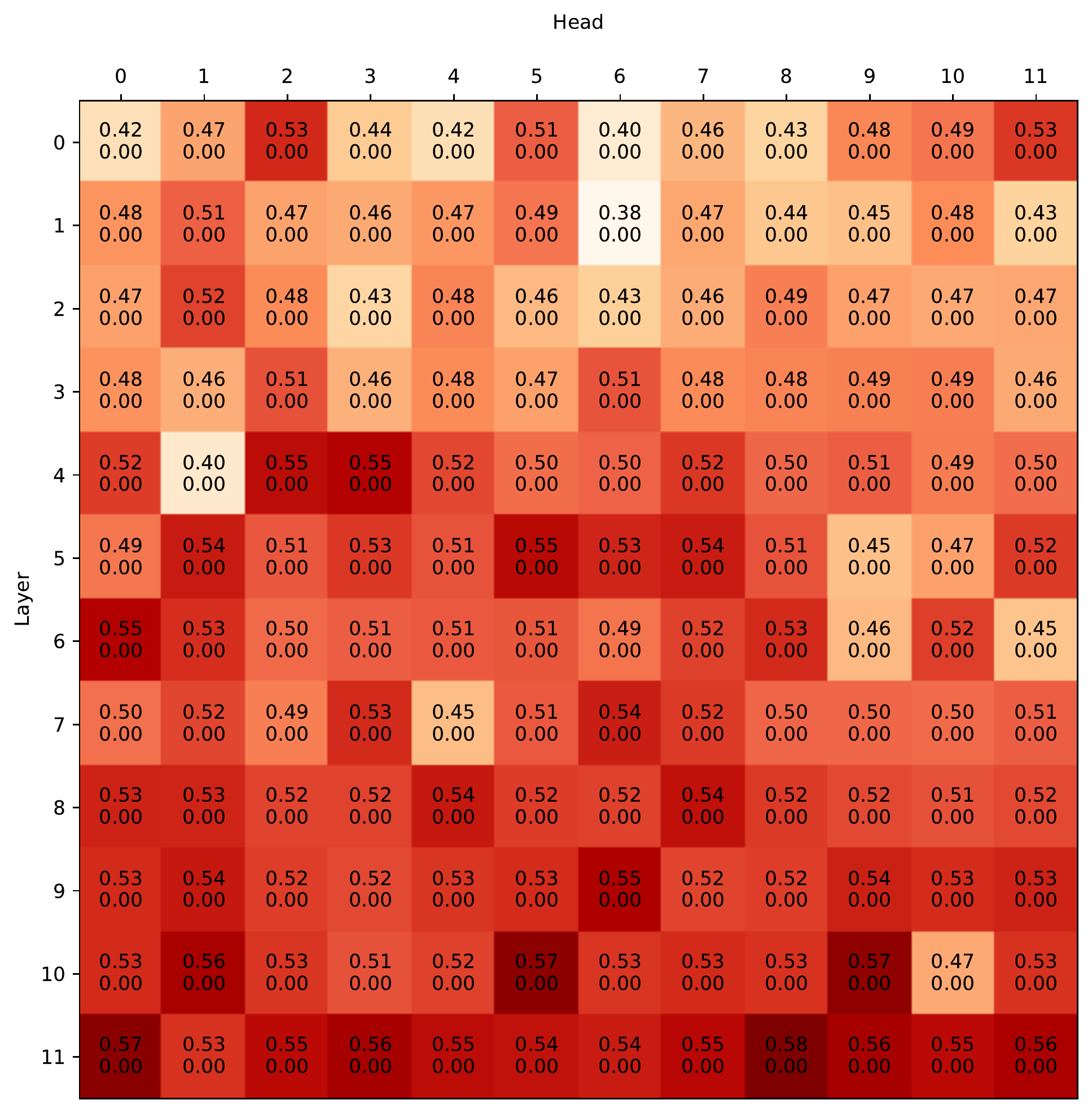}
            \includegraphics[trim=-20 -30 -20 30,clip,width=\linewidth]{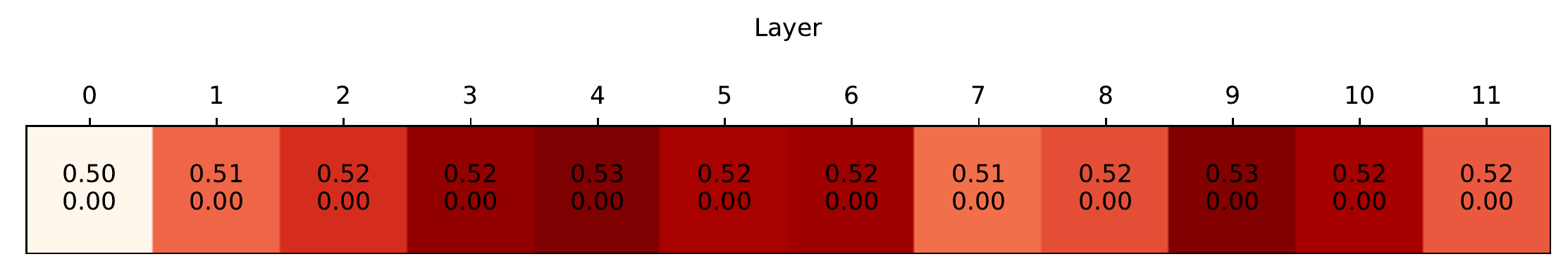}
            \caption{M-pruning}
        \end{subfigure}
\hfill
        \begin{subfigure}[b]{0.5\textwidth}
            \includegraphics[trim=-20 -10 -30 30,clip,width=\linewidth]{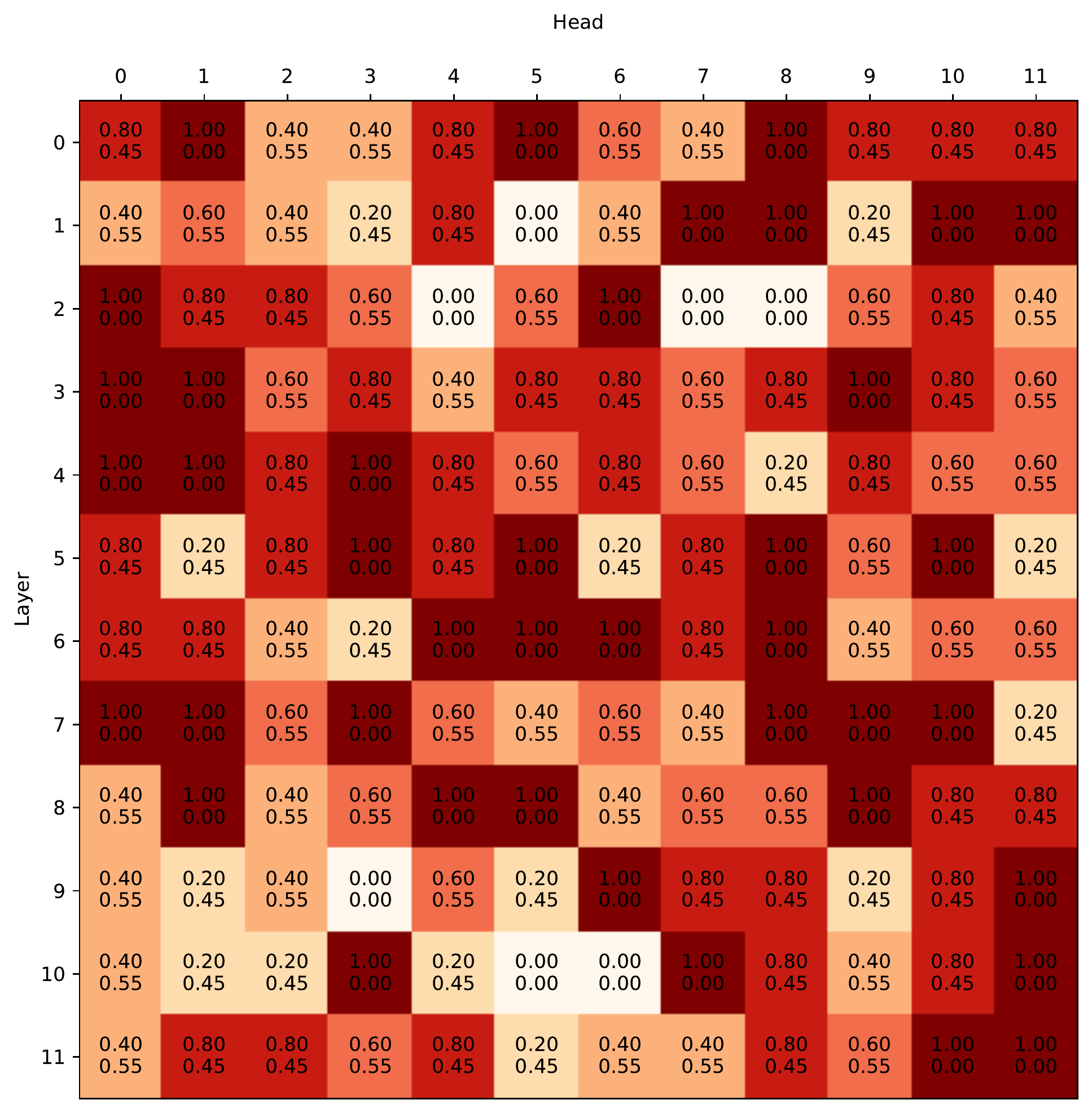}
            \includegraphics[trim=-20 -30 -20 30,clip,width=\linewidth]{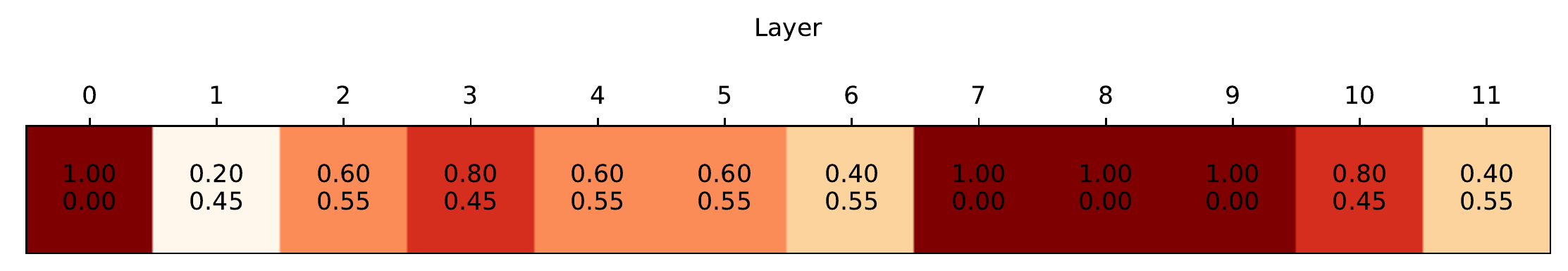}
            \caption{S-pruning}
        \end{subfigure}
        \caption{MNLI}
\end{figure*}

\begin{figure*}
        \begin{subfigure}[b]{0.5\textwidth}
            \includegraphics[trim=-20 -10 -30 30,clip,width=\linewidth]{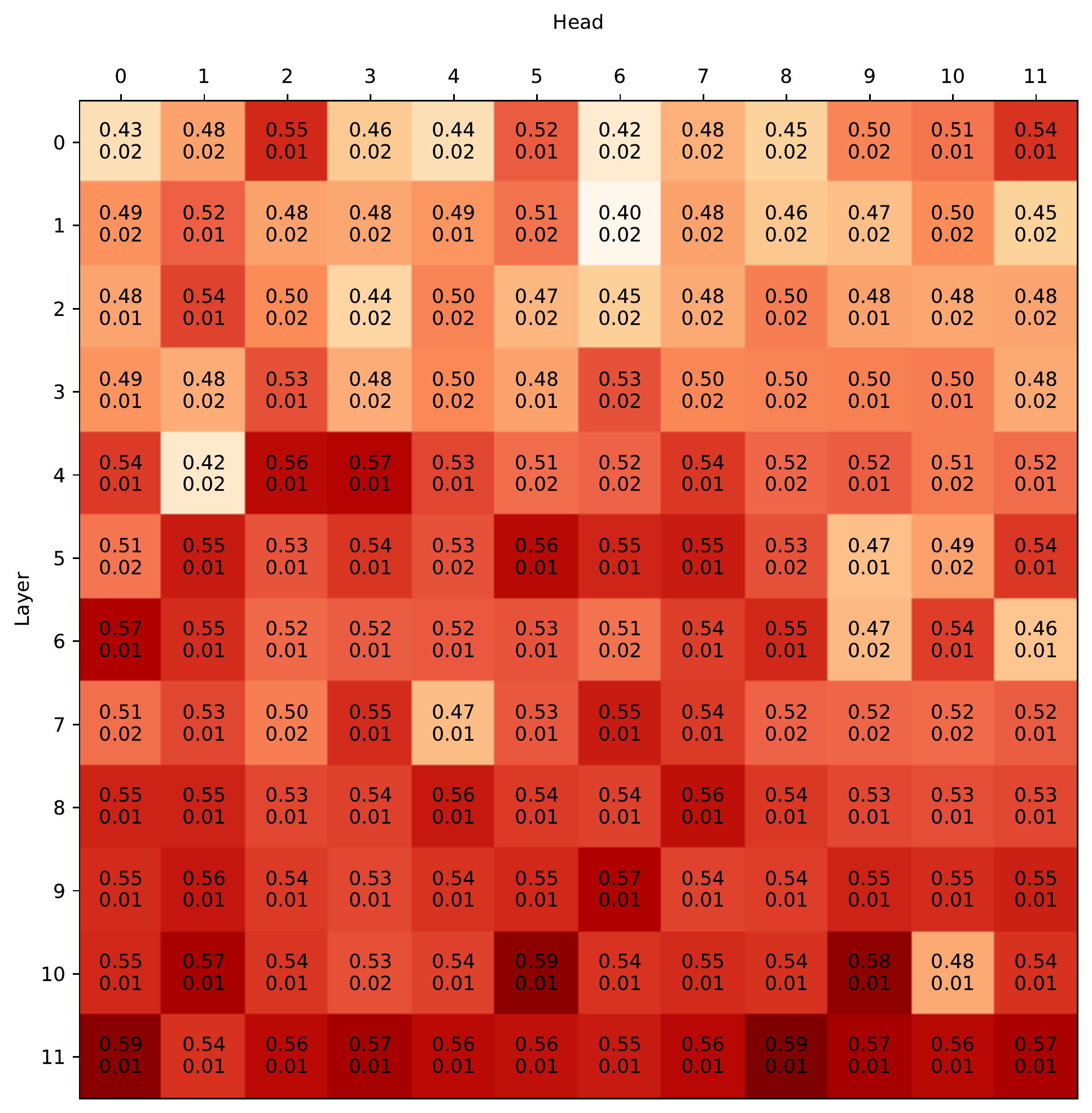}
            \includegraphics[trim=-20 -30 -20 30,clip,width=\linewidth]{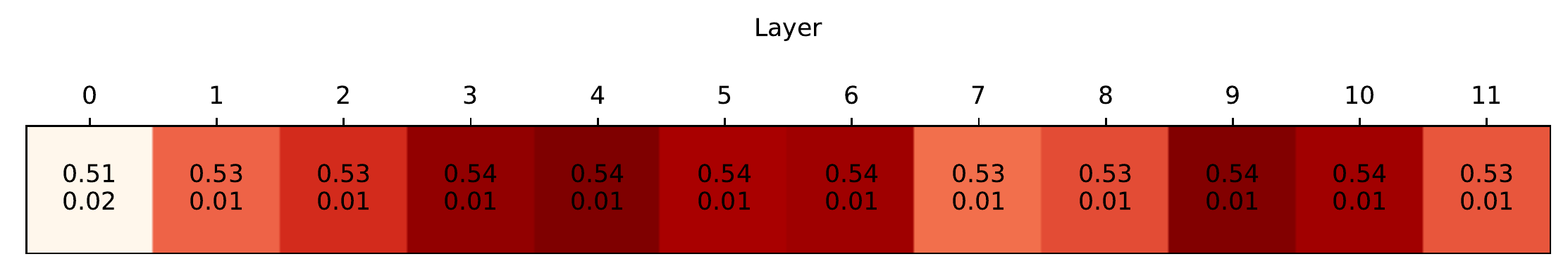}
            \caption{M-pruning}
        \end{subfigure}
\hfill
        \begin{subfigure}[b]{0.5\textwidth}
            \includegraphics[trim=-20 -10 -30 30,clip,width=\linewidth]{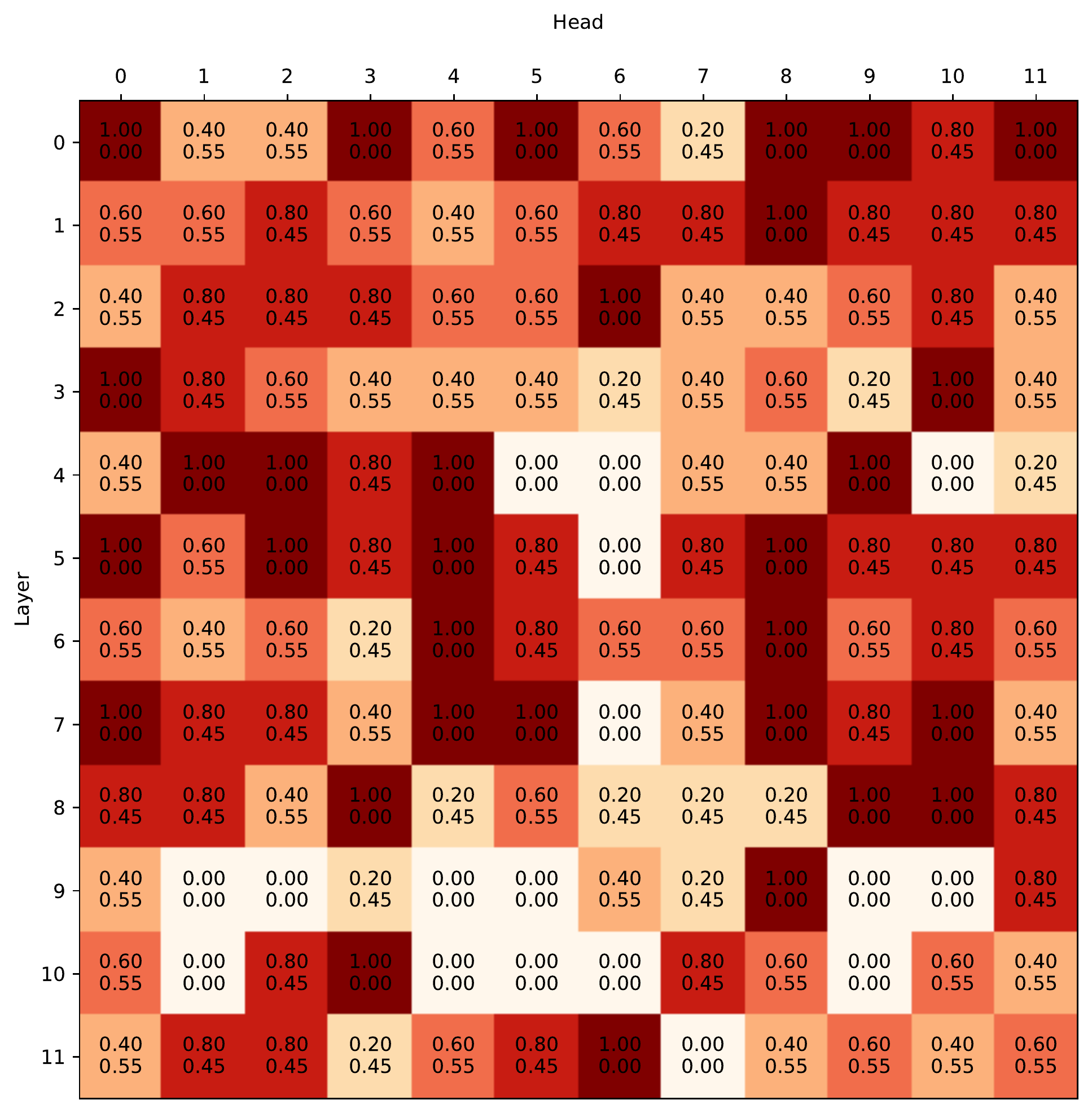}
            \includegraphics[trim=-20 -30 -20 30,clip,width=\linewidth]{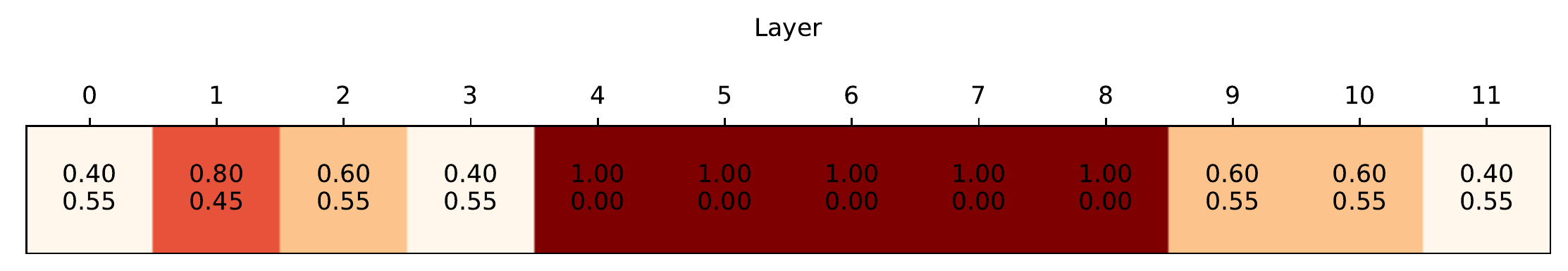}
            \caption{S-pruning}
        \end{subfigure}
        \caption{QNLI}
\end{figure*}

\begin{figure*}
        \begin{subfigure}[b]{0.5\textwidth}
            \includegraphics[trim=-20 -10 -30 30,clip,width=\linewidth]{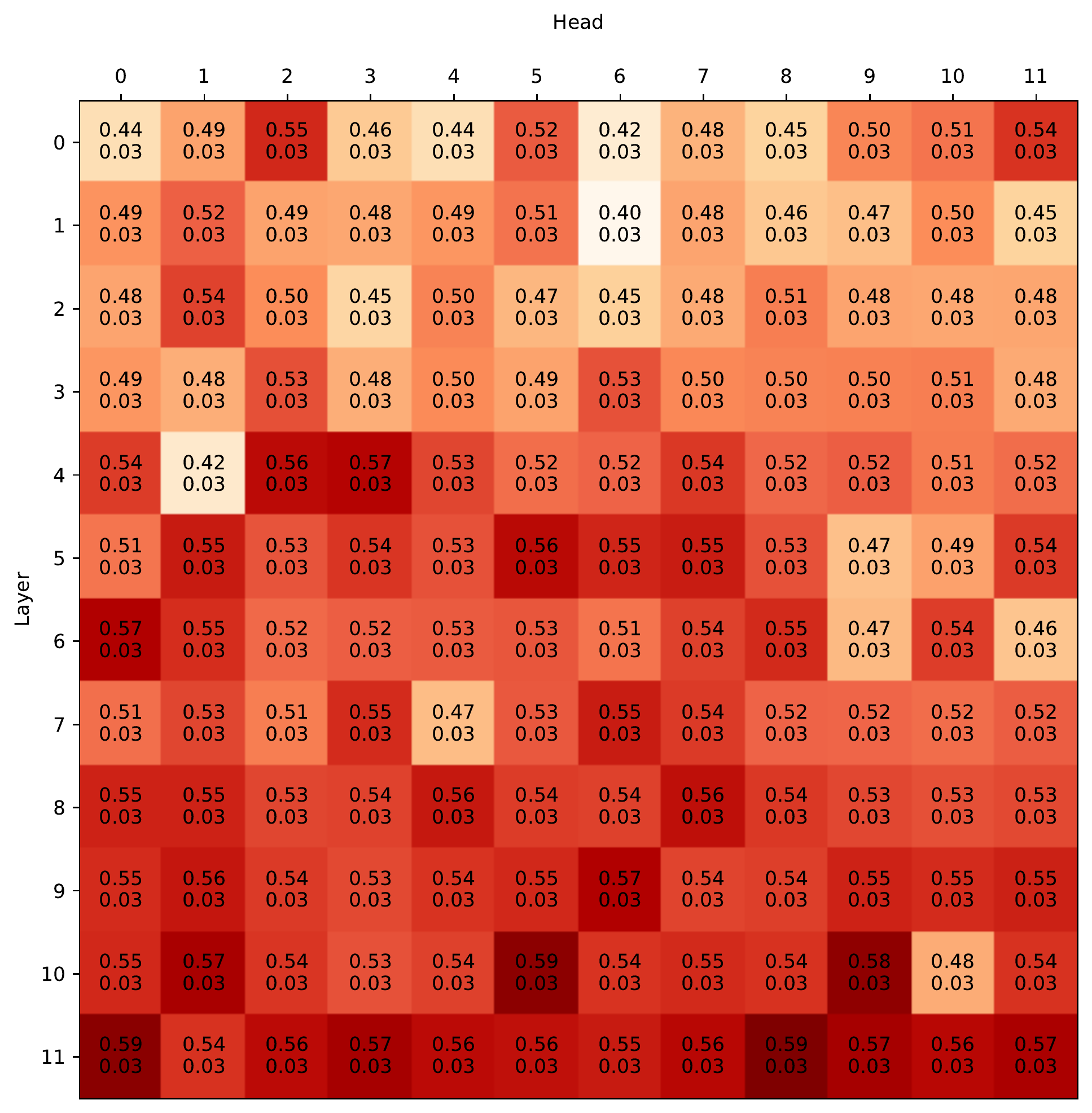}
            \includegraphics[trim=-20 -30 -20 30,clip,width=\linewidth]{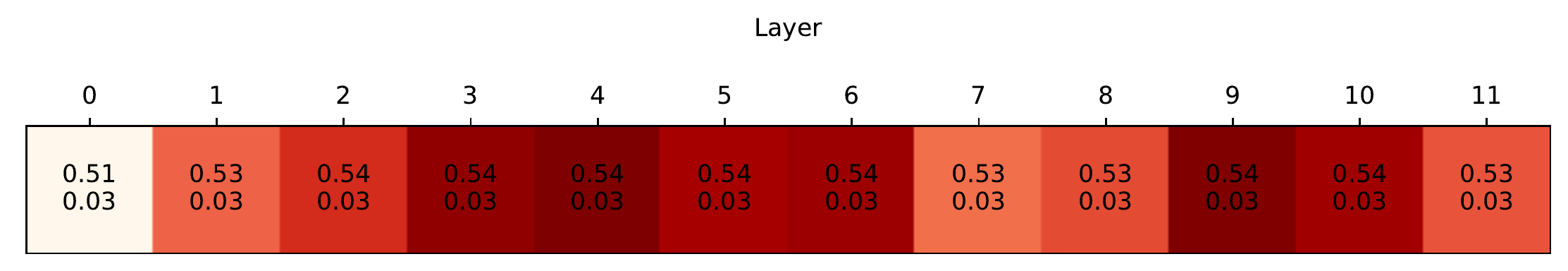}
            \caption{M-pruning}
        \end{subfigure}
\hfill
        \begin{subfigure}[b]{0.5\textwidth}
            \includegraphics[trim=-20 -10 -30 30,clip,width=\linewidth]{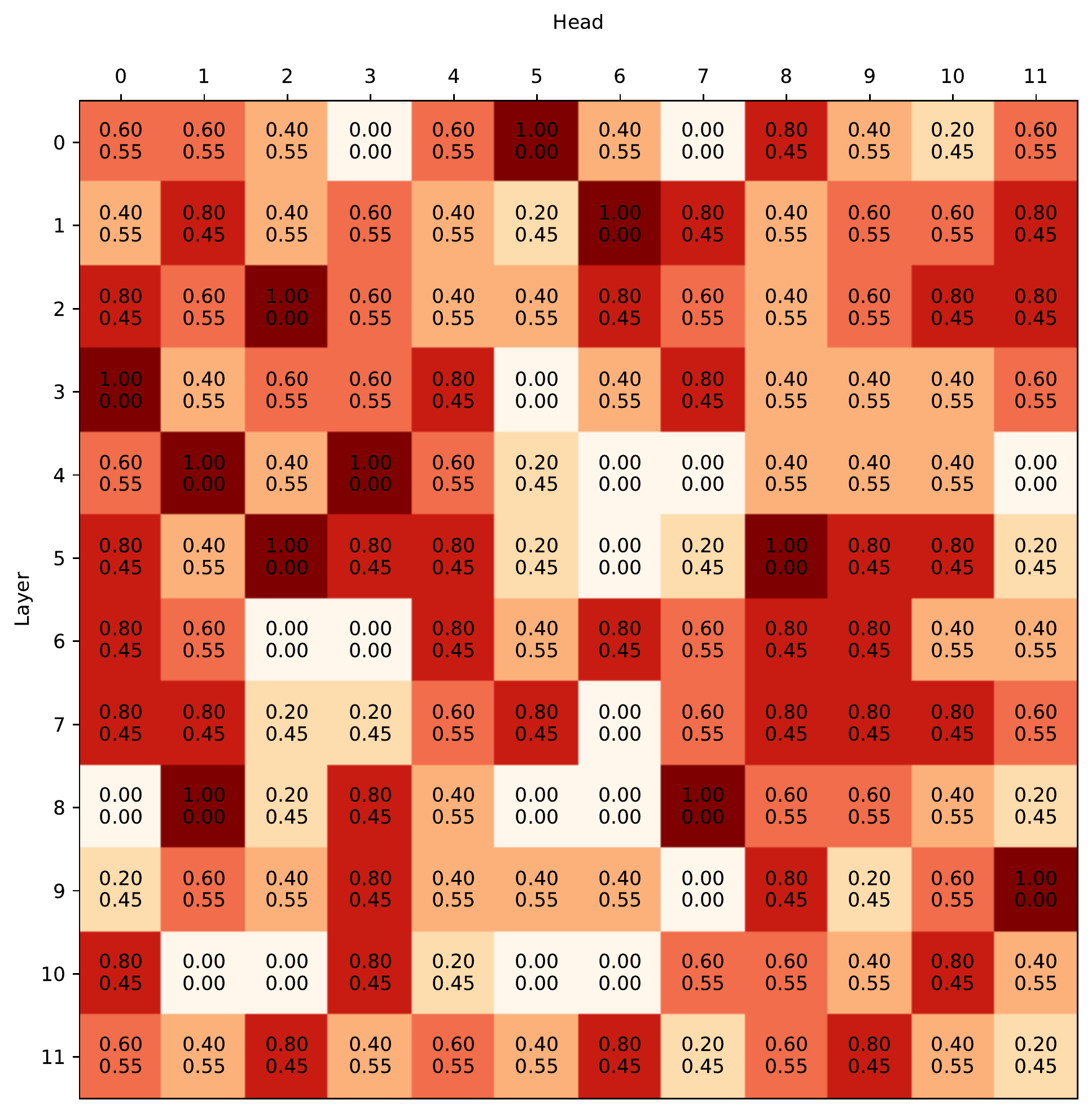}
            \includegraphics[trim=-20 -30 -20 30,clip,width=\linewidth]{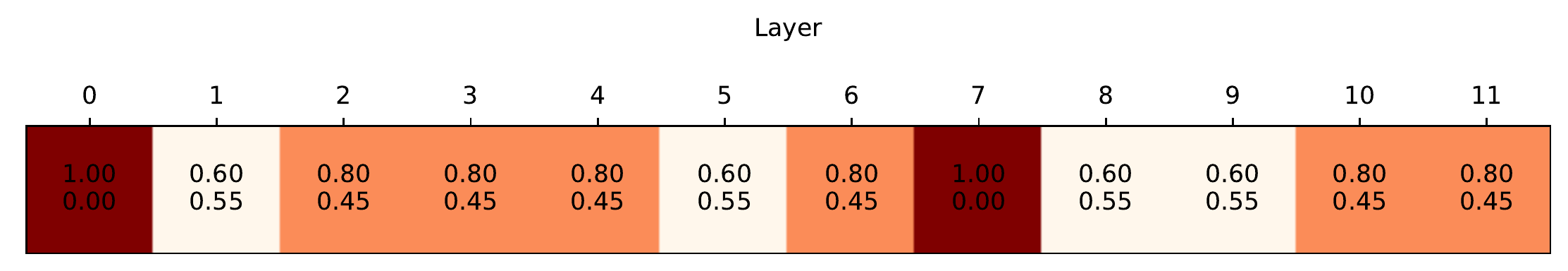}
            \caption{S-pruning}
        \end{subfigure}
        \caption{RTE}
\end{figure*}

\begin{figure*}
        \begin{subfigure}[b]{0.5\textwidth}
            \includegraphics[trim=-20 -10 -30 30,clip,width=\linewidth]{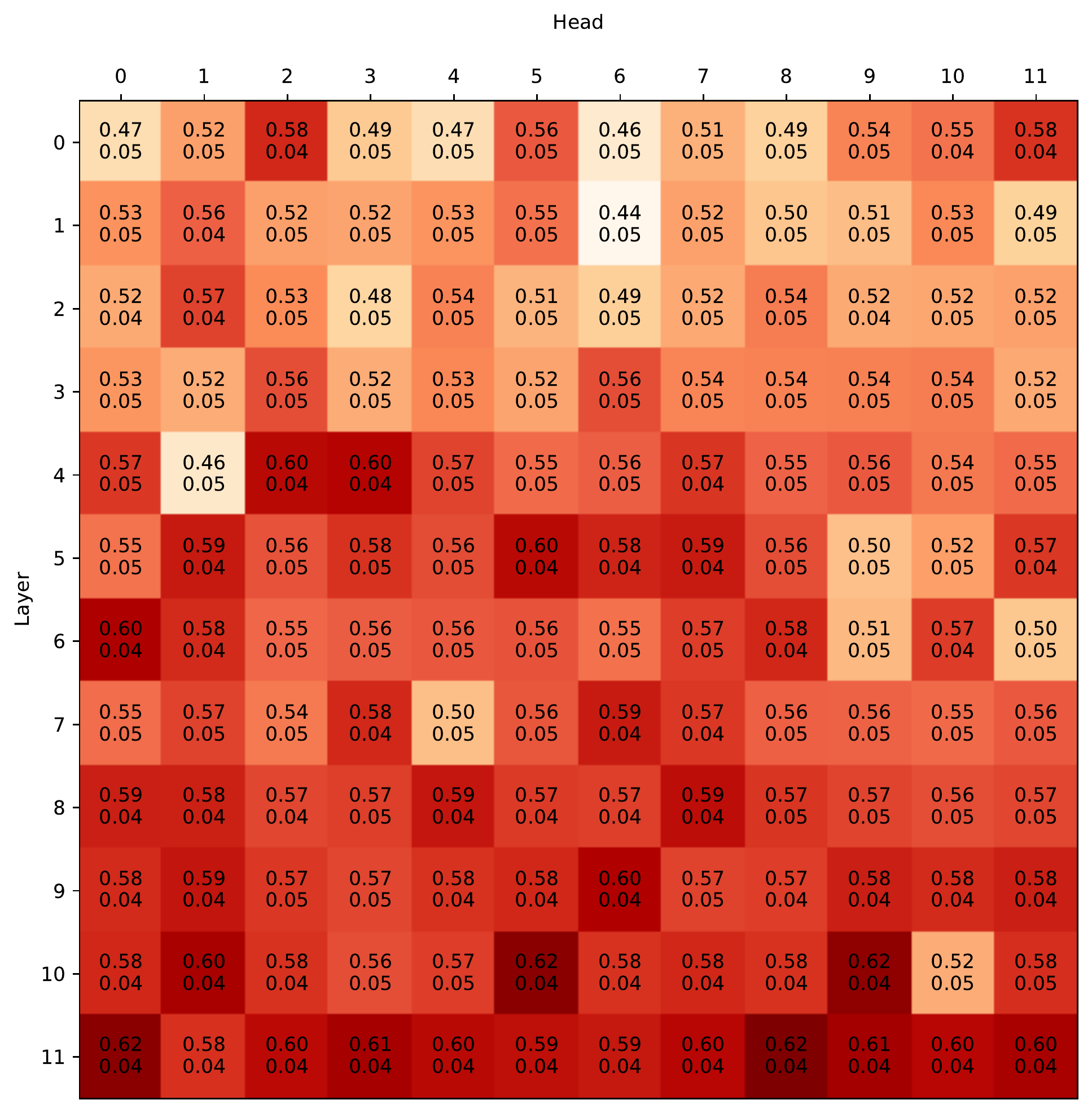}
            \includegraphics[trim=-20 -30 -20 30,clip,width=\linewidth]{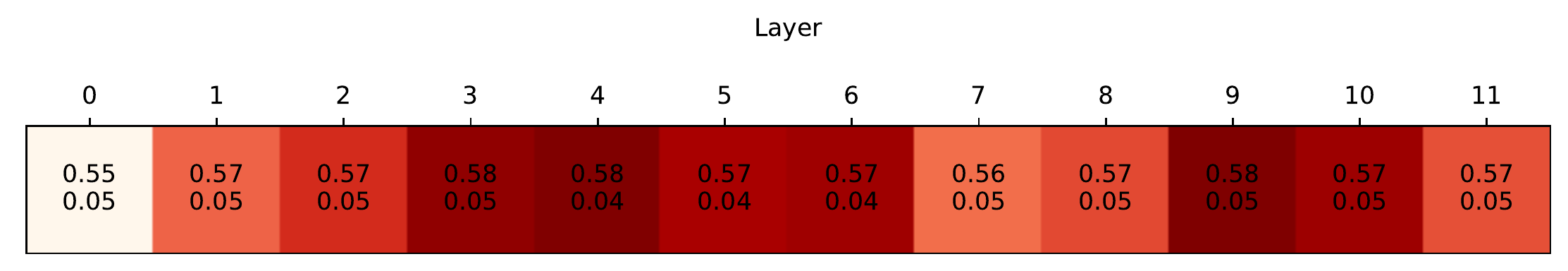}
            \caption{M-pruning}
        \end{subfigure}
\hfill
        \begin{subfigure}[b]{0.5\textwidth}
            \includegraphics[trim=-20 -10 -30 30,clip,width=\linewidth]{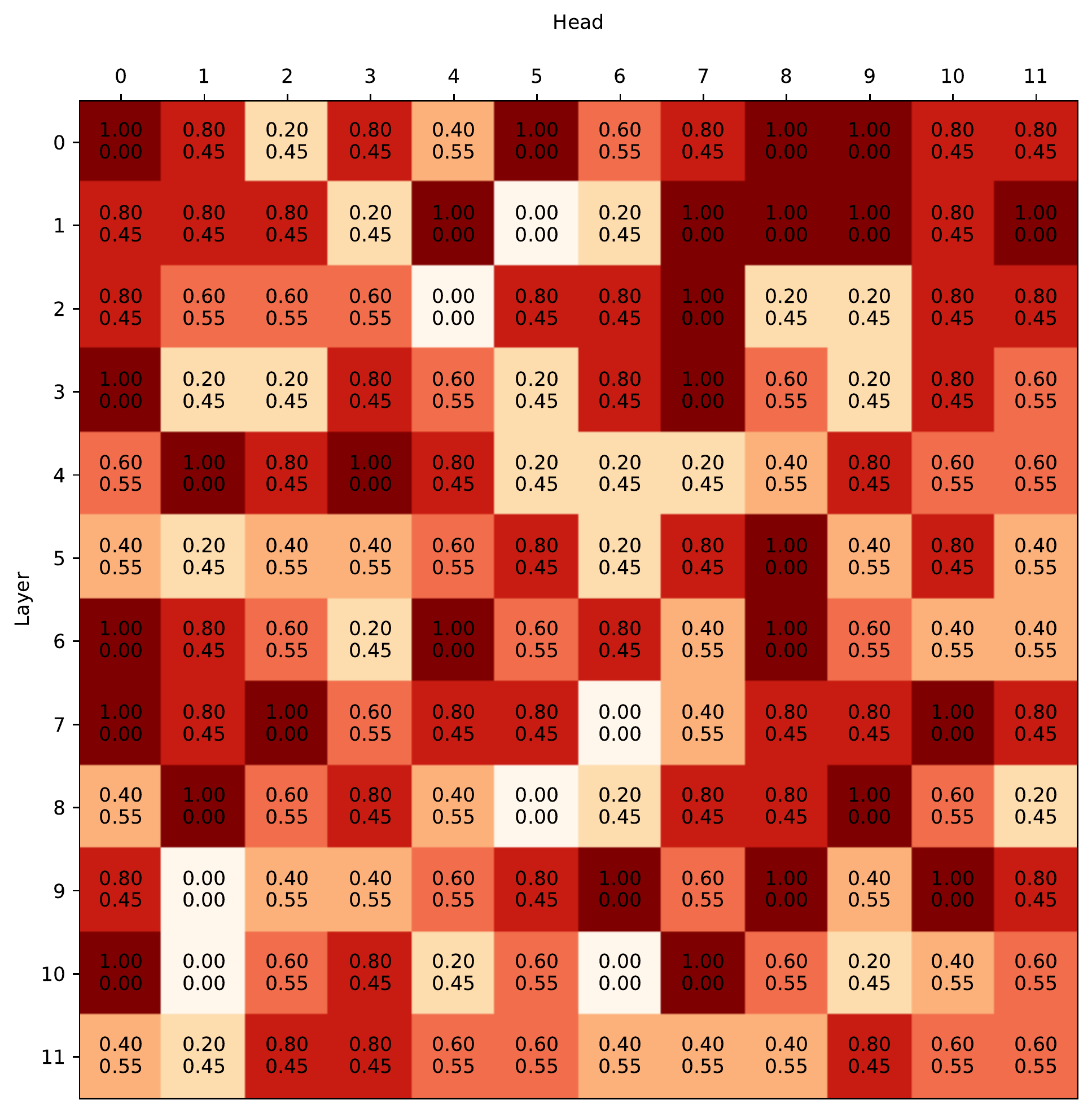}
            \includegraphics[trim=-20 -30 -20 30,clip,width=\linewidth]{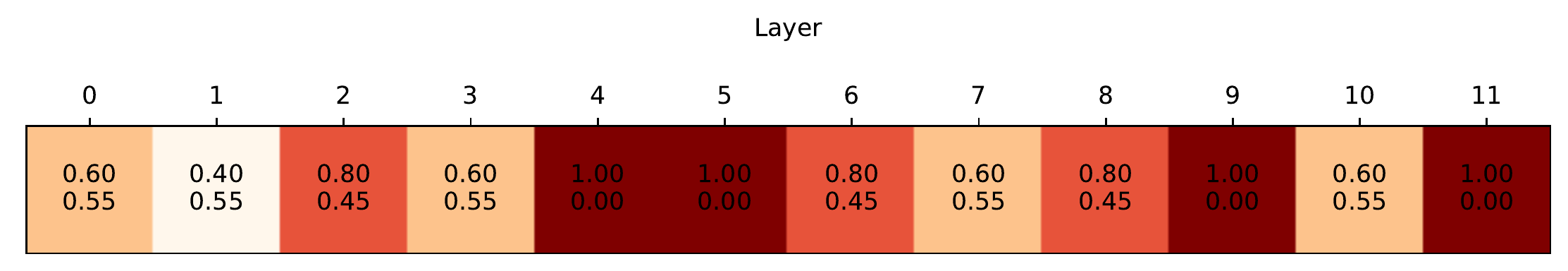}
            \caption{S-pruning}
        \end{subfigure}
        \caption{MRPC}
\end{figure*}

\begin{figure*}
        \begin{subfigure}[b]{0.5\textwidth}
            \includegraphics[trim=-20 -10 -30 30,clip,width=\linewidth]{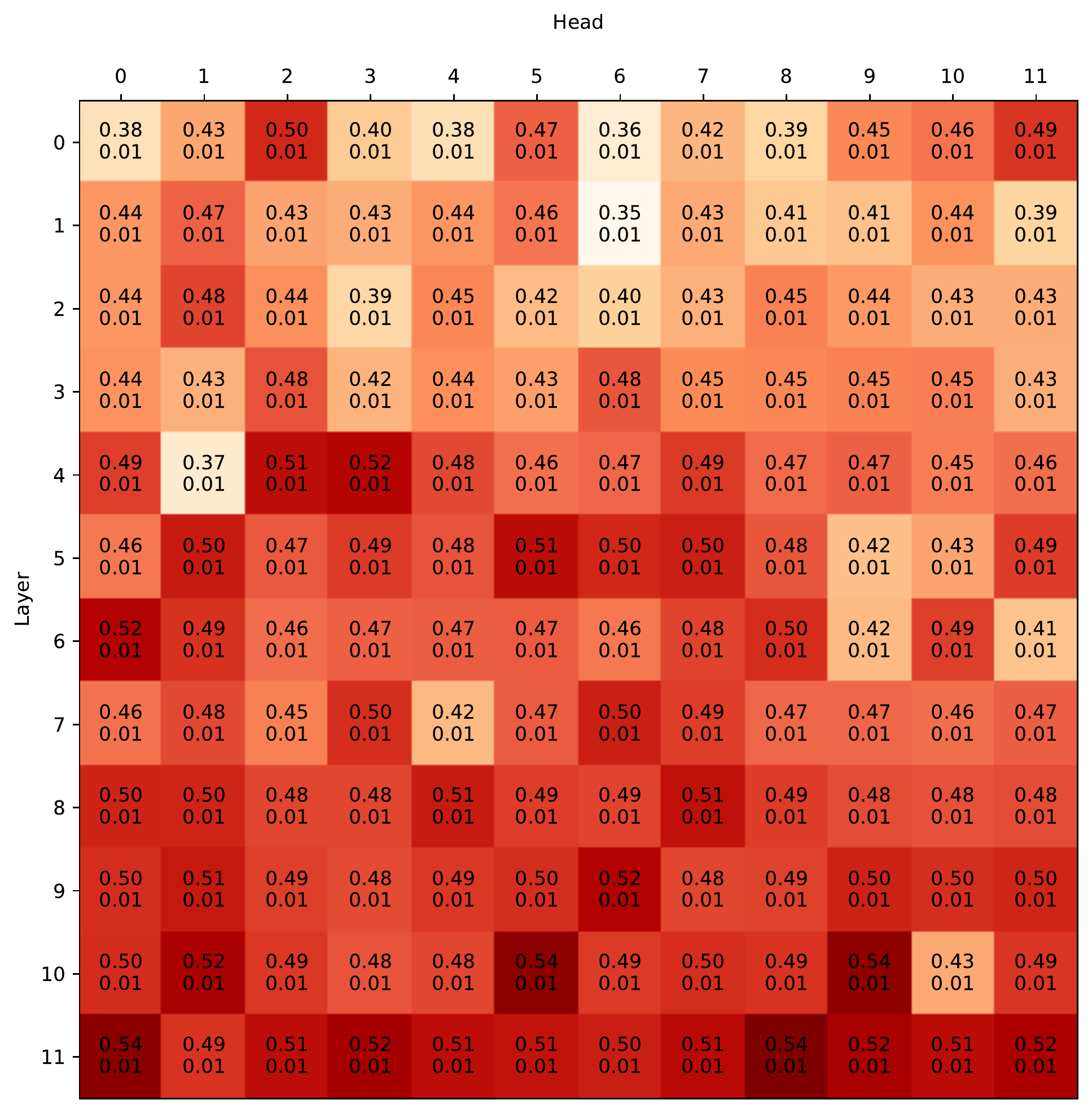}
            \includegraphics[trim=-20 -30 -20 30,clip,width=\linewidth]{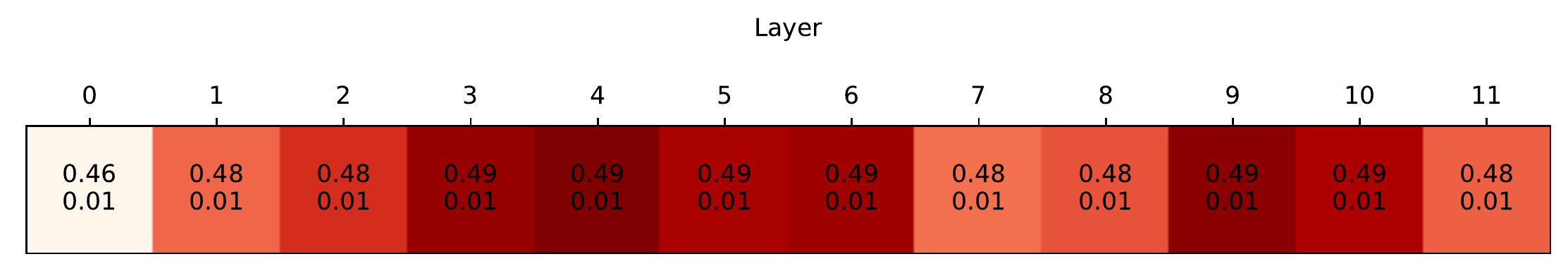}
            \caption{M-pruning}
        \end{subfigure}
\hfill
        \begin{subfigure}[b]{0.5\textwidth}
            \includegraphics[trim=-20 -10 -30 30,clip,width=\linewidth]{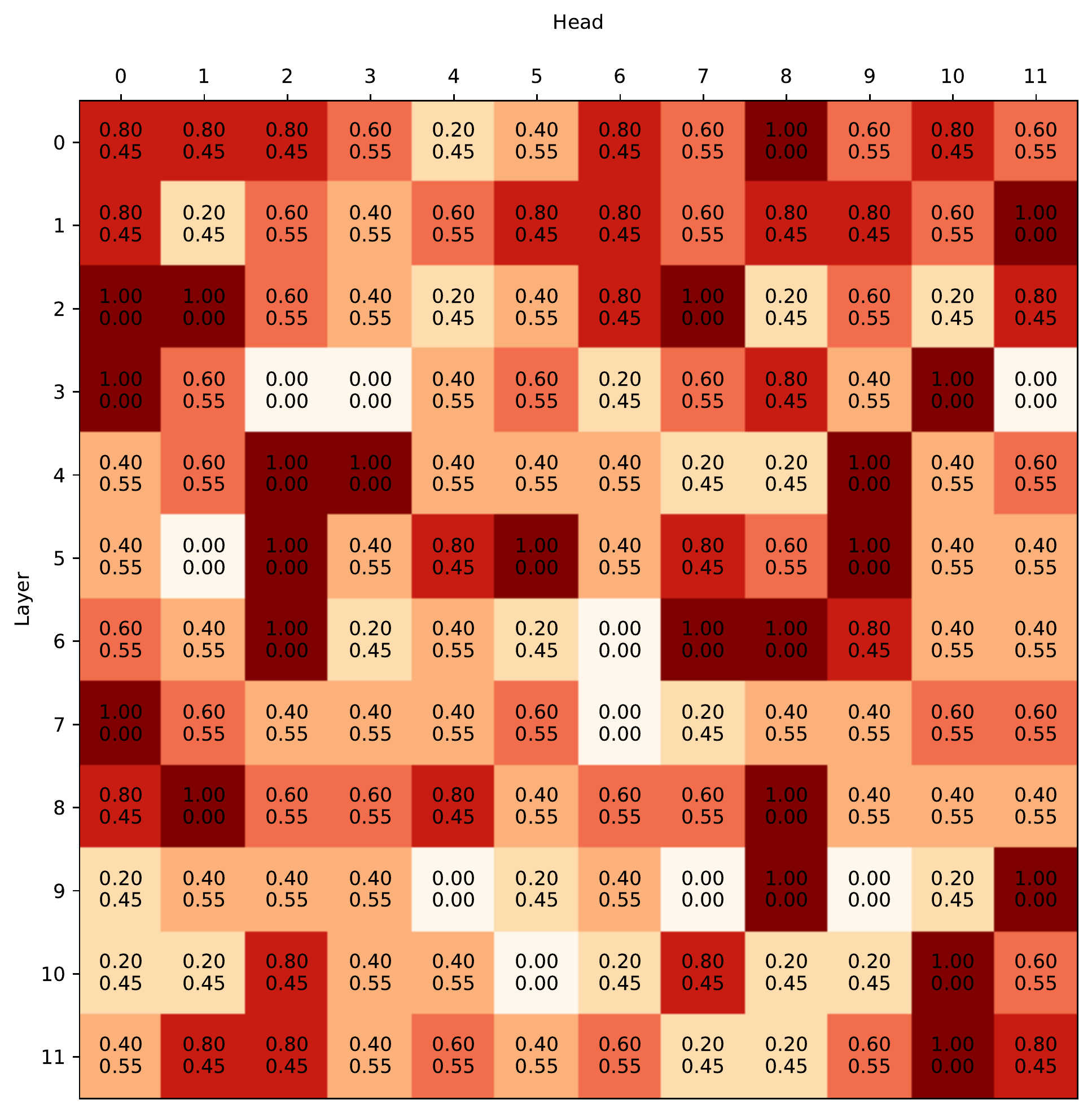}
            \includegraphics[trim=-20 -30 -20 30,clip,width=\linewidth]{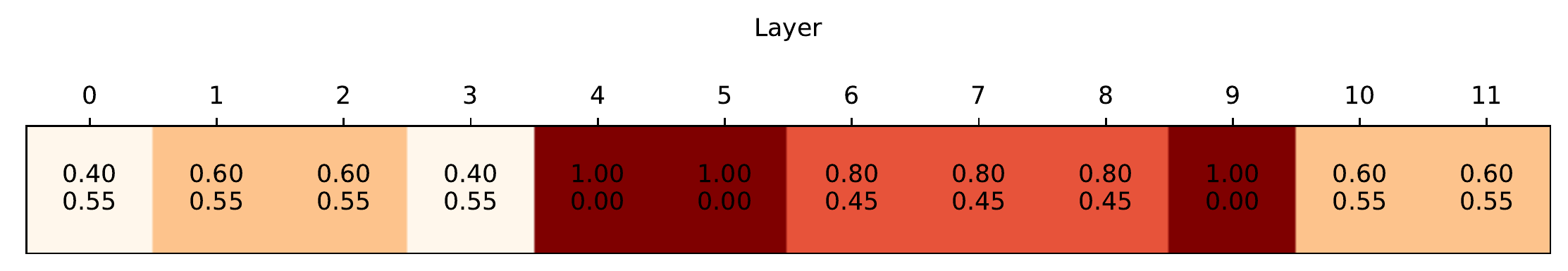}
            \caption{S-pruning}
        \end{subfigure}
        \caption{QQP}
\end{figure*}

\begin{figure*}
        \begin{subfigure}[b]{0.5\textwidth}
            \includegraphics[trim=-20 -10 -30 30,clip,width=\linewidth]{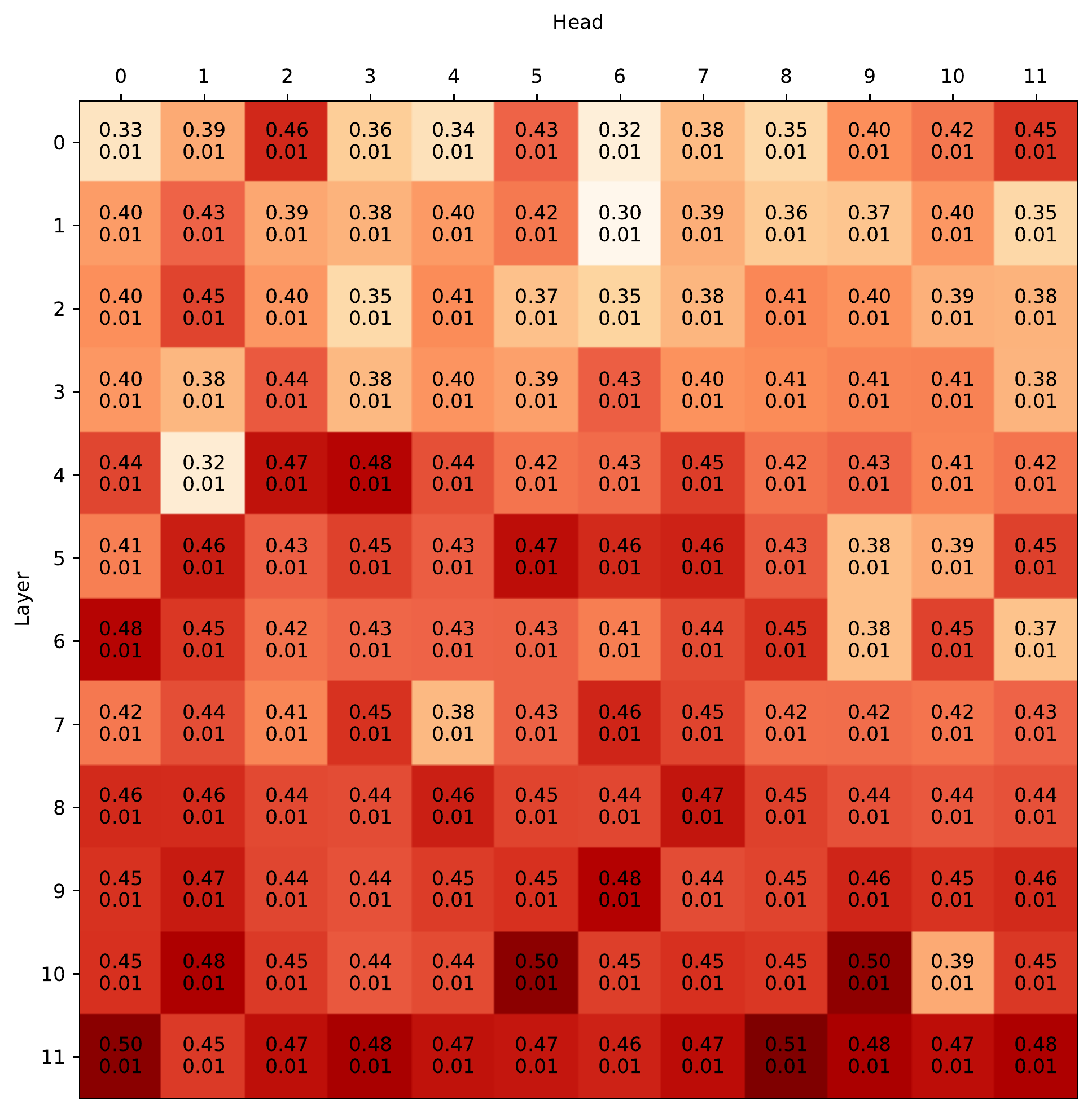}
            \includegraphics[trim=-20 -30 -20 30,clip,width=\linewidth]{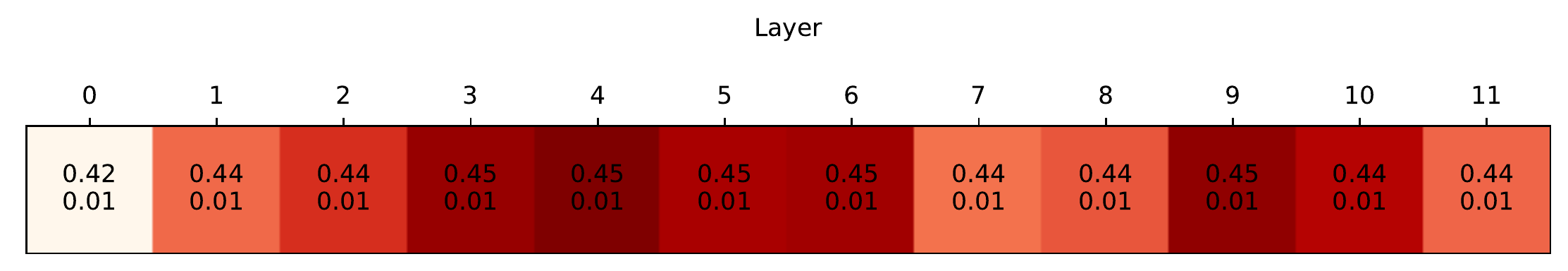}
            \caption{M-pruning}
        \end{subfigure}
\hfill
        \begin{subfigure}[b]{0.5\textwidth}
            \includegraphics[trim=-20 -10 -30 30,clip,width=\linewidth]{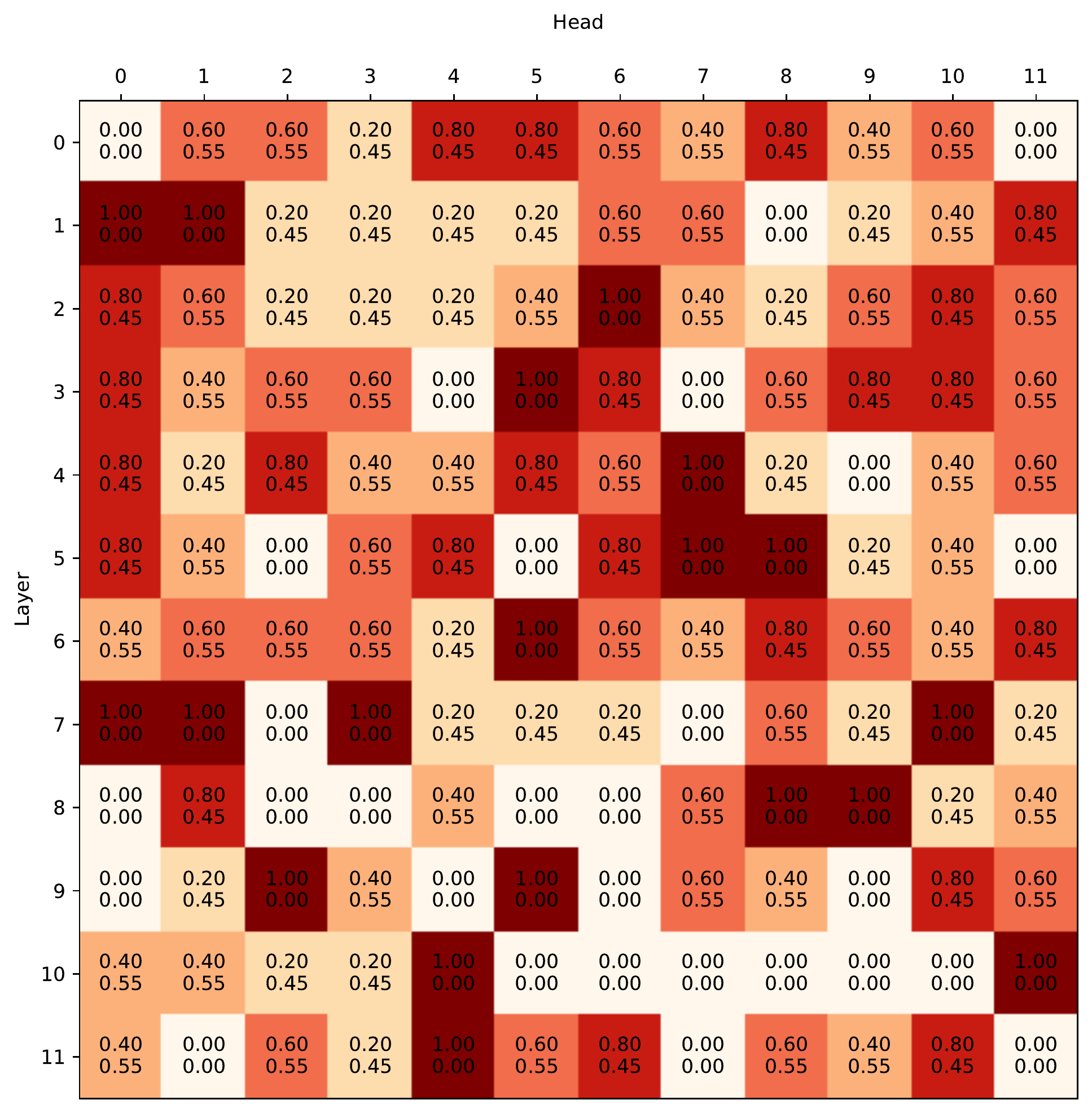}
            \includegraphics[trim=-20 -30 -20 30,clip,width=\linewidth]{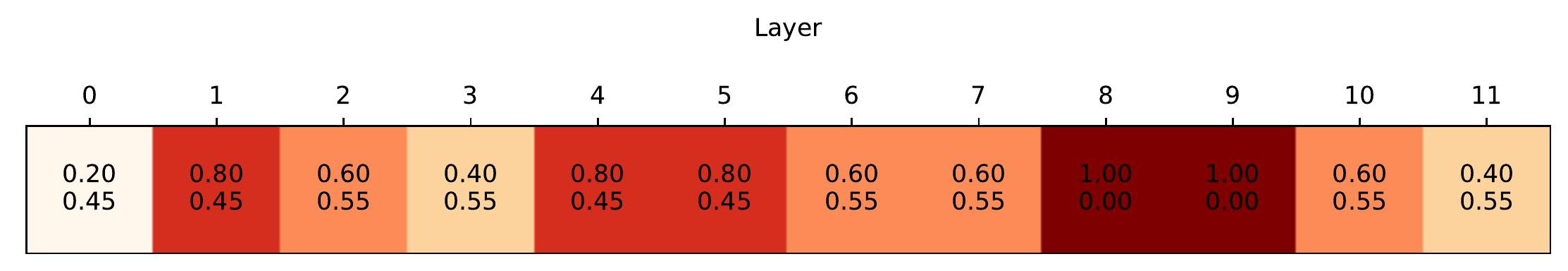}
            \caption{S-pruning}
        \end{subfigure}
        \caption{SST-2}
\end{figure*}

\begin{figure*}
        \begin{subfigure}[b]{0.5\textwidth}
            \includegraphics[trim=-20 -10 -30 30,clip,width=\linewidth]{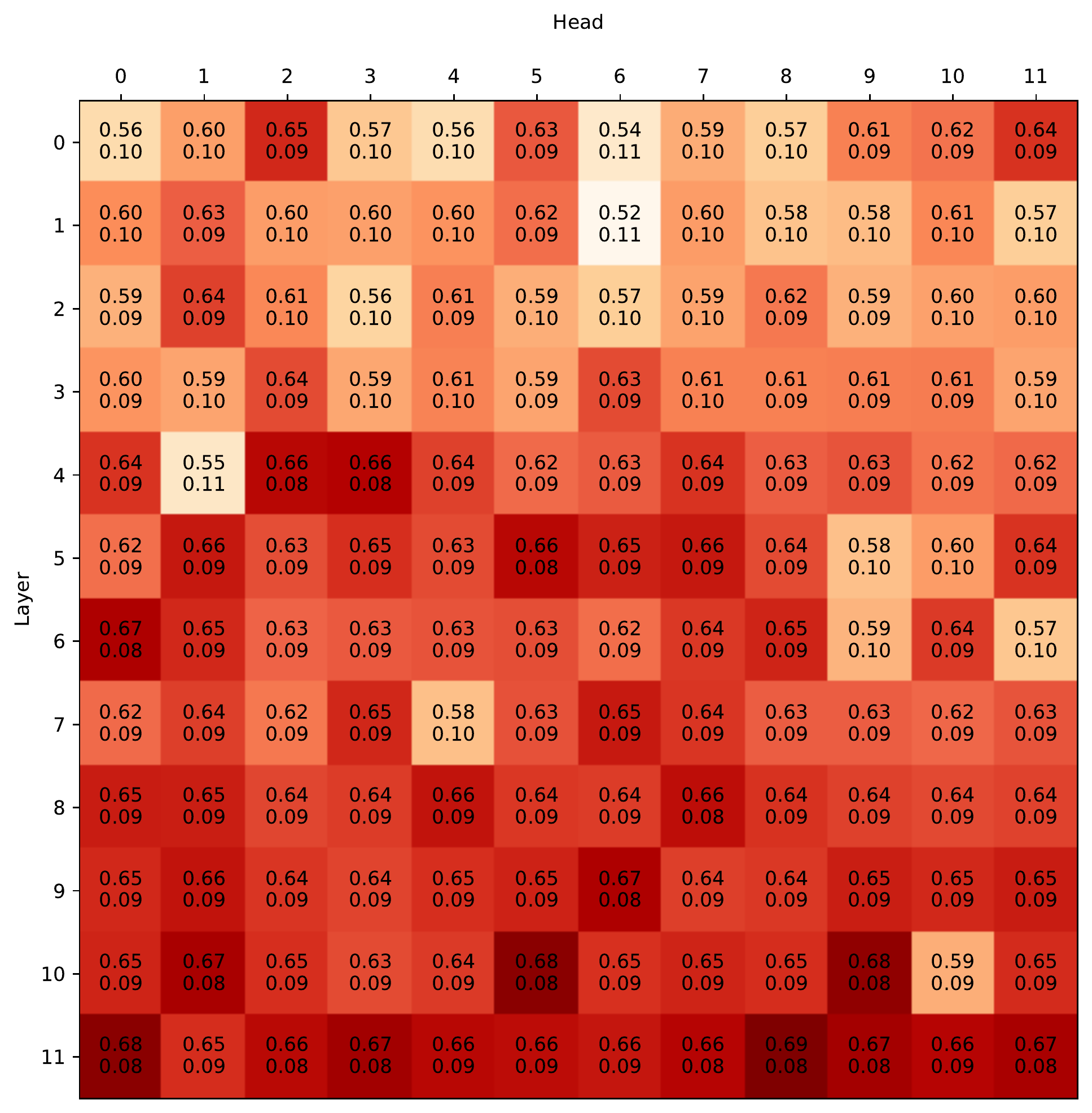}
            \includegraphics[trim=-20 -30 -20 30,clip,width=\linewidth]{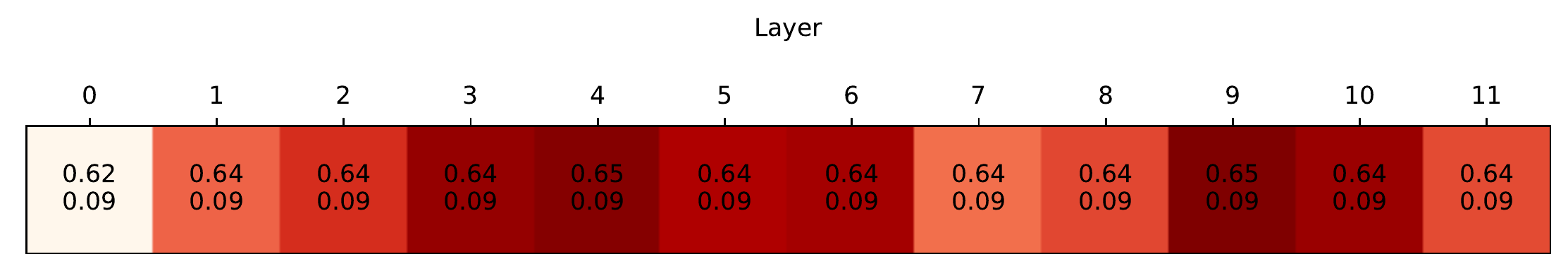}
            \caption{M-pruning}
        \end{subfigure}
\hfill
        \begin{subfigure}[b]{0.5\textwidth}
            \includegraphics[trim=-20 -10 -30 30,clip,width=\linewidth]{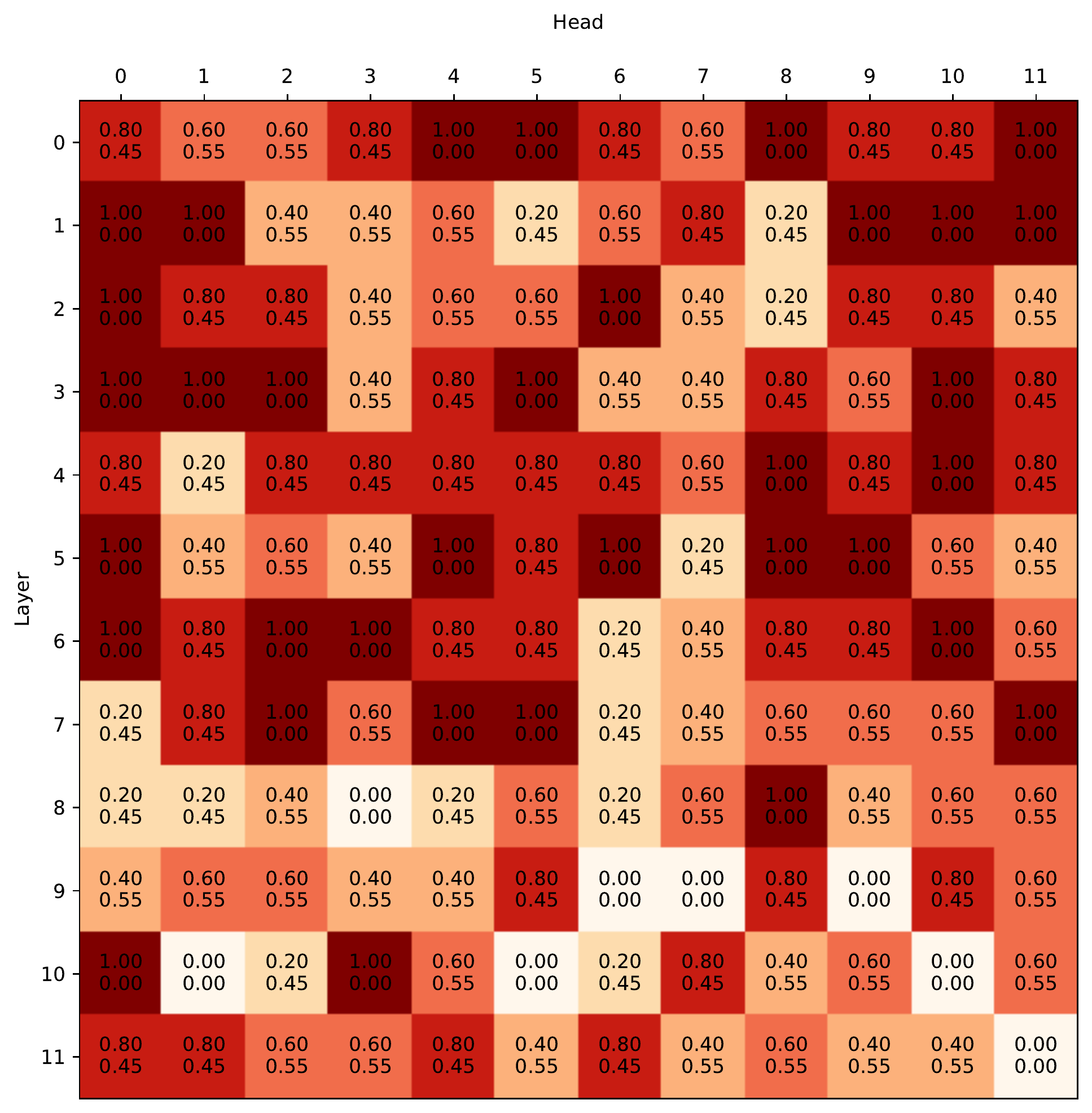}
            \includegraphics[trim=-20 -30 -20 30,clip,width=\linewidth]{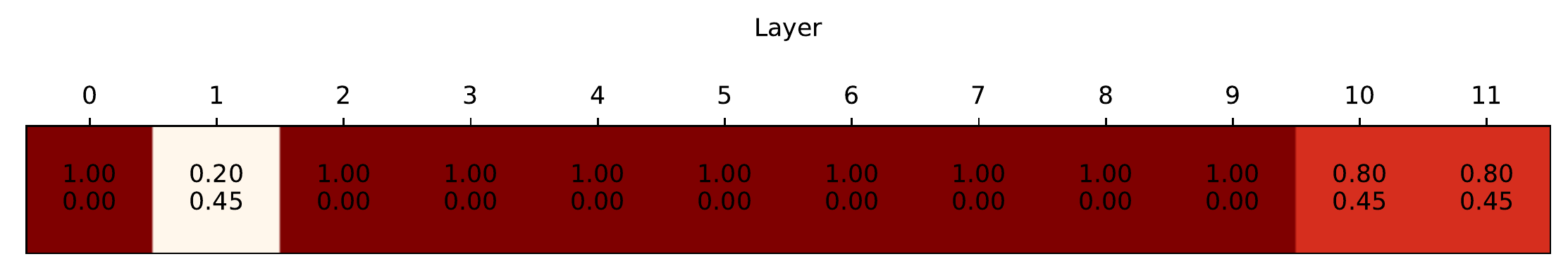}
            \caption{S-pruning}
        \end{subfigure}
        \caption{CoLA}
\end{figure*}

\begin{figure*}
        \begin{subfigure}[b]{0.5\textwidth}
            \includegraphics[trim=-20 -10 -30 30,clip,width=\linewidth]{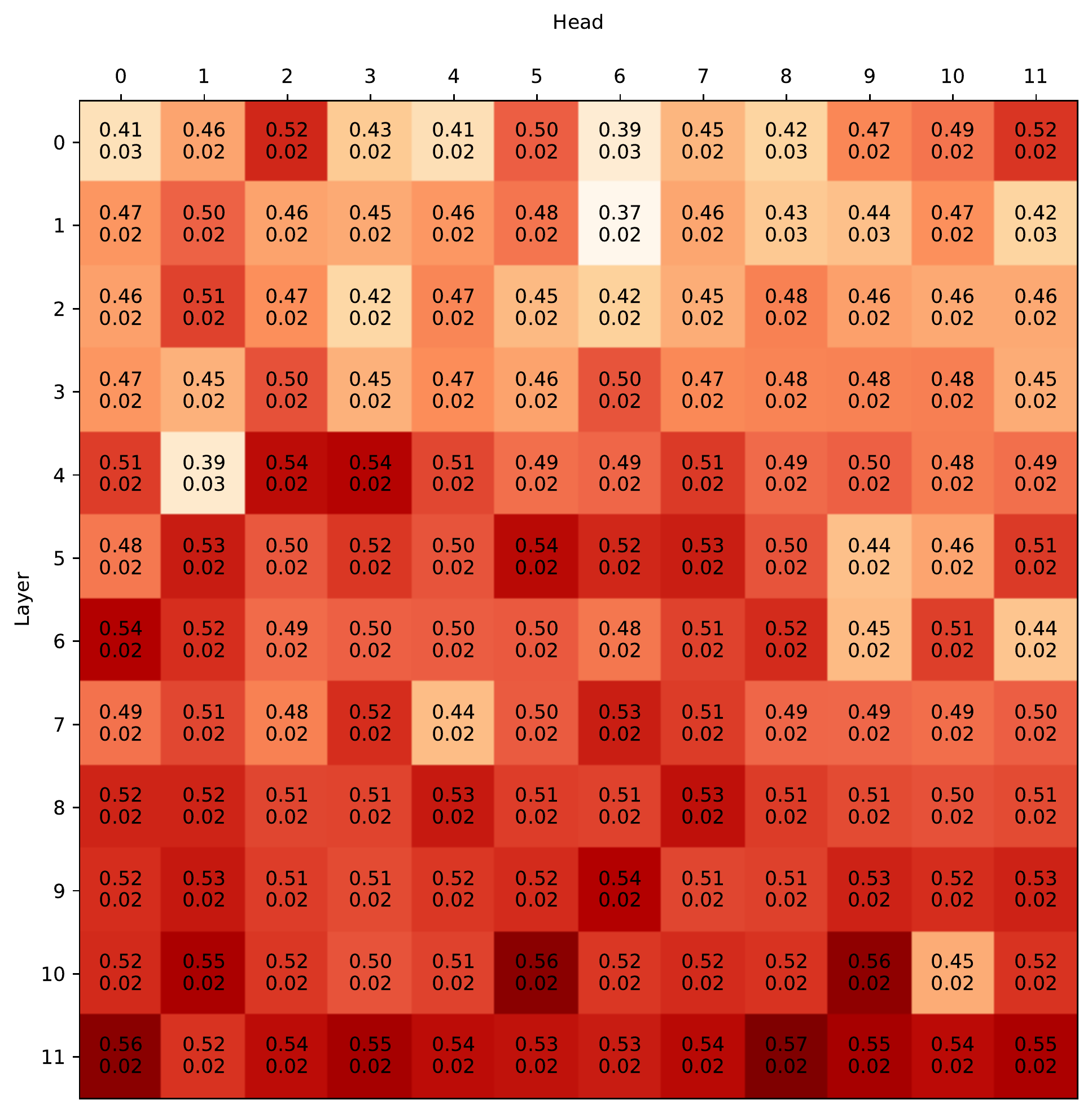}
            \includegraphics[trim=-20 -30 -20 30,clip,width=\linewidth]{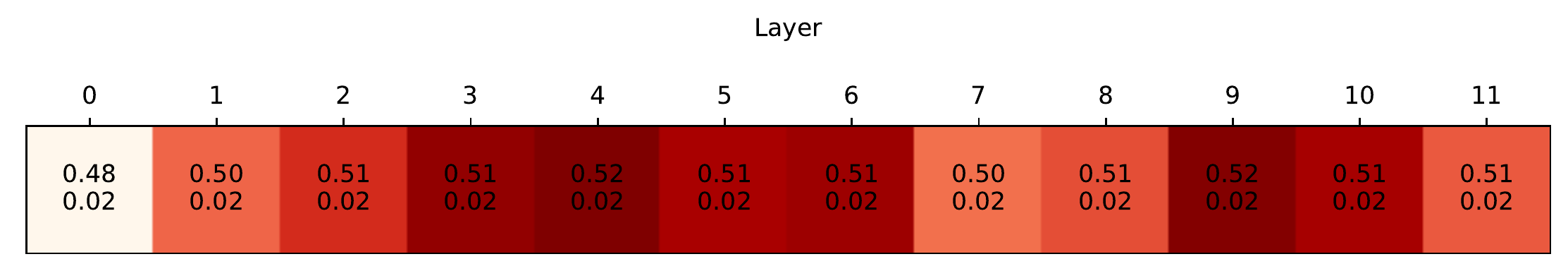}
            \caption{M-pruning}
        \end{subfigure}
\hfill
        \begin{subfigure}[b]{0.5\textwidth}
            \includegraphics[trim=-20 -10 -30 30,clip,width=\linewidth]{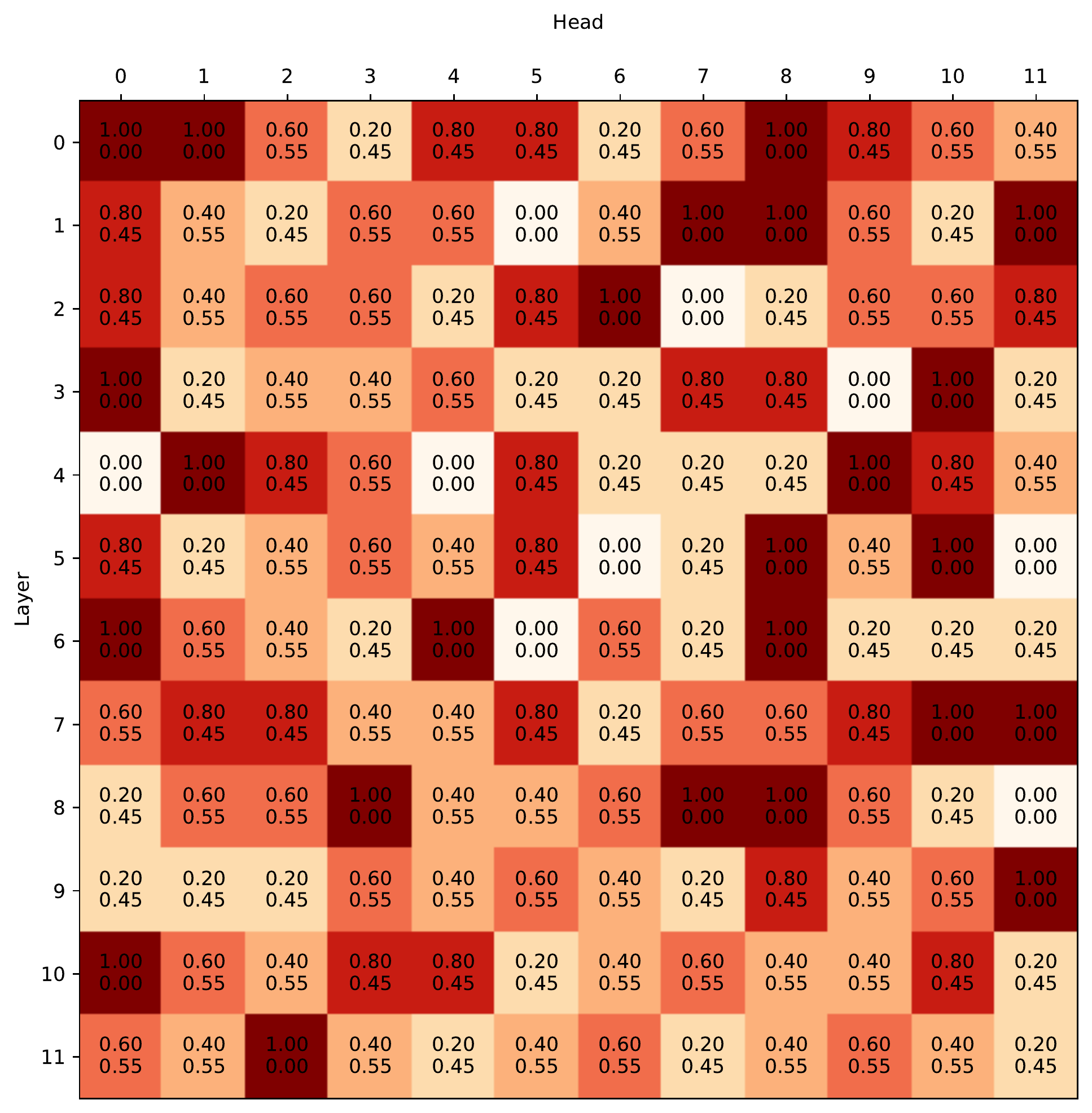}
            \includegraphics[trim=-20 -30 -20 30,clip,width=\linewidth]{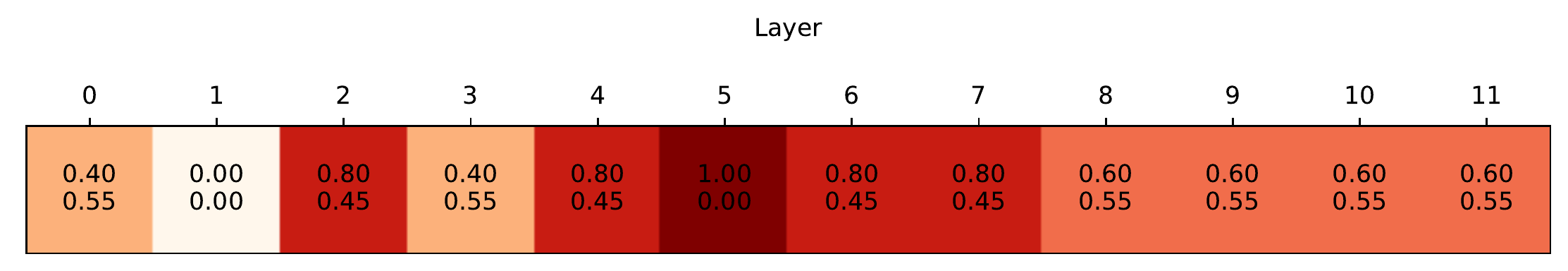}
            \caption{S-pruning}
        \end{subfigure}
        \caption{STS-B}
\end{figure*}

\begin{figure*}
        \begin{subfigure}[b]{0.5\textwidth}
            \includegraphics[trim=-20 -10 -30 30,clip,width=\linewidth]{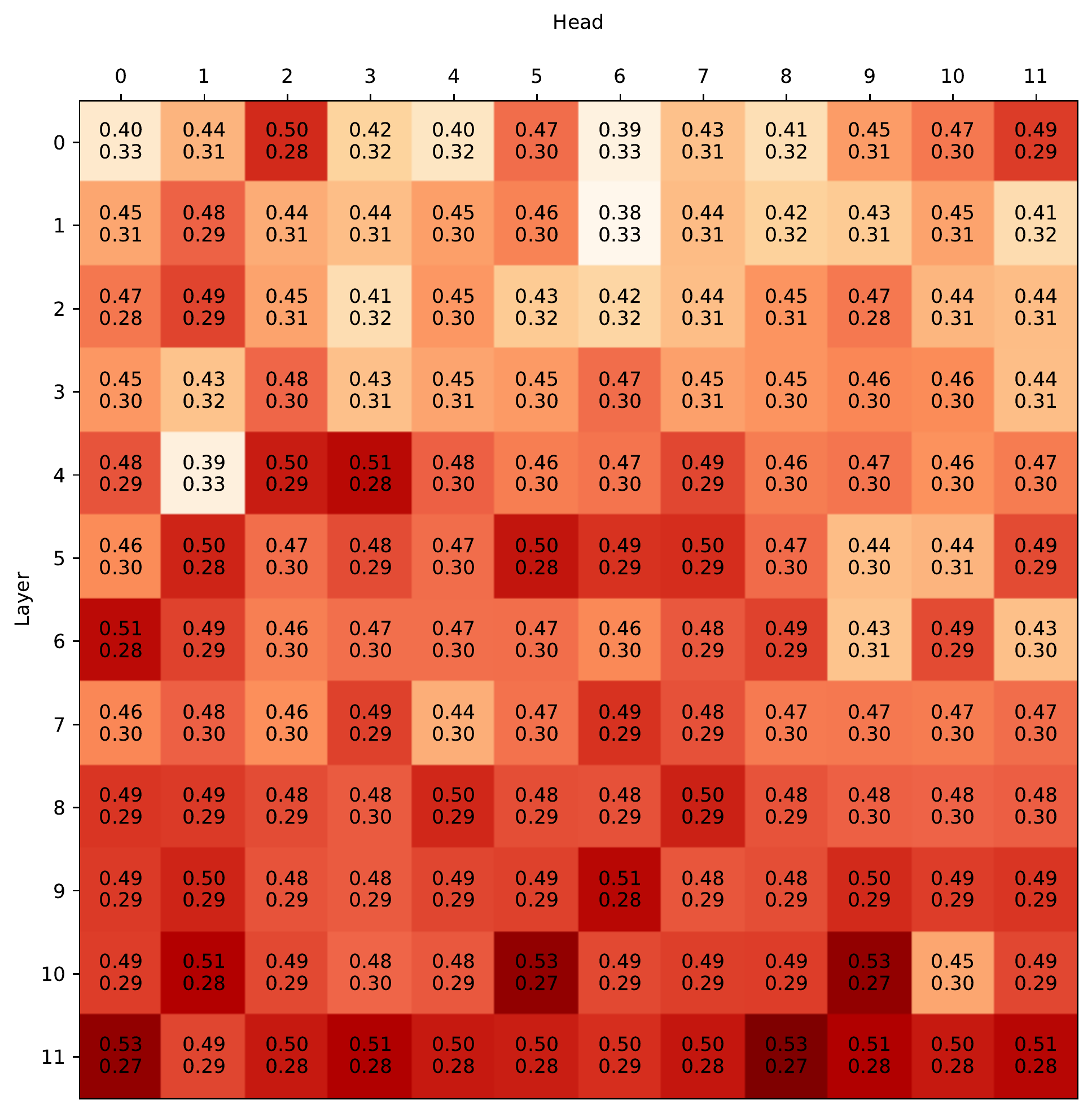}
            \includegraphics[trim=-20 -30 -20 30,clip,width=\linewidth]{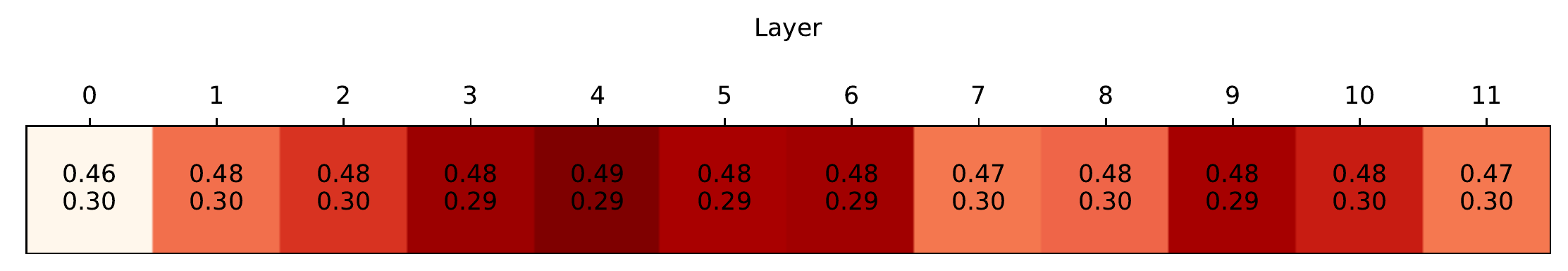}
            \caption{M-pruning}
        \end{subfigure}
\hfill
        \begin{subfigure}[b]{0.5\textwidth}
            \includegraphics[trim=-20 -10 -30 30,clip,width=\linewidth]{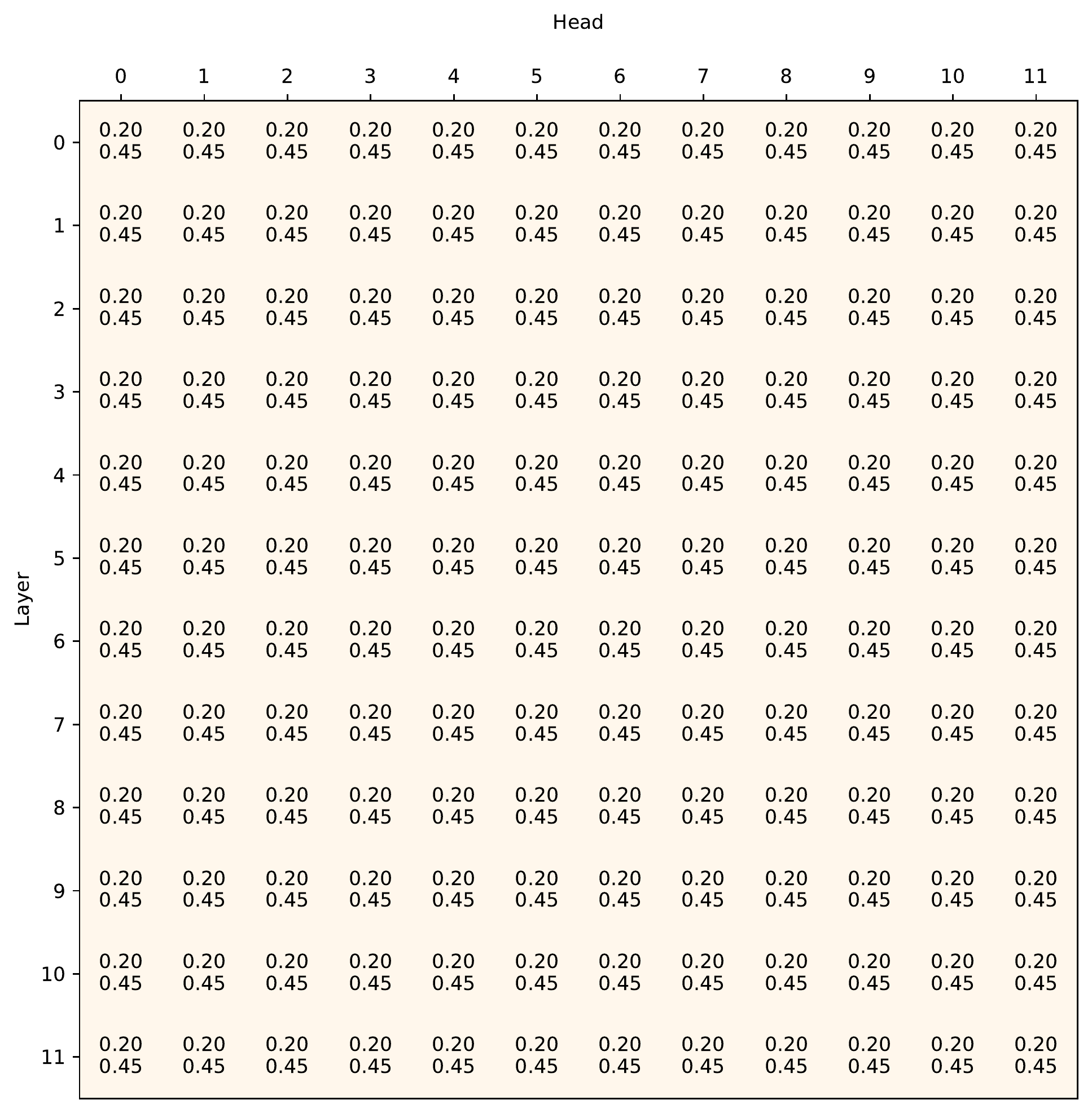}
            \includegraphics[trim=-20 -30 -20 30,clip,width=\linewidth]{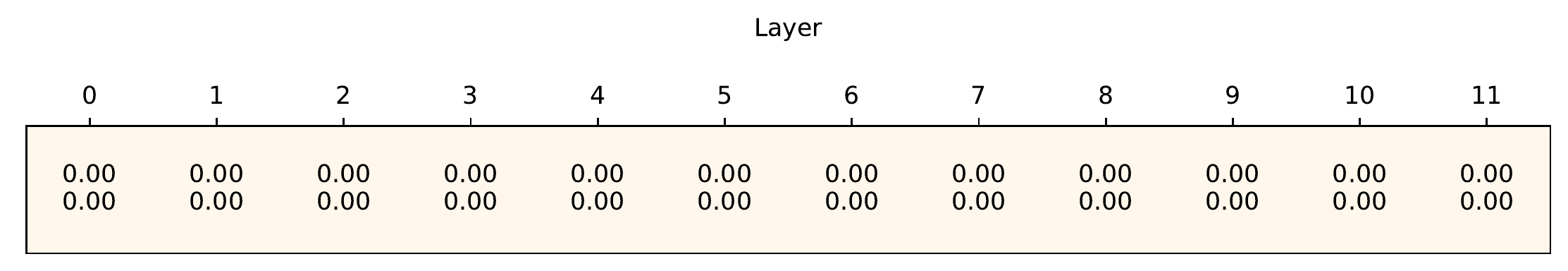}
            \caption{S-pruning}
        \end{subfigure}
        \caption{WNLI}
        \label{fig:wnli}        
\end{figure*}

\clearpage

\begin{figure*}[t]
\section{Iterative Pruning Modes}
\label{appendix:pruning-modes}

\begin{multicols}{2}
We conducted additional experiments with the following settings for iterative pruning based on importance scores:

\begin{itemize}
    \item \textit{Heads only}: in each iteration, we mask as many of the unmasked heads with the lowest importance scores as we can (144 heads in the full BERT-base model).%
    \item \textit{MLPs only}: we iteratively mask one of the remaining MLPs that has the smallest importance score (\autoref{eq:head_importance}). 
    \item \textit{Heads and MLPs}: we compute head (\autoref{eq:head_importance}) and MLP (\autoref{eq:head_importance}) importance scores in a single backward pass, pruning 10\% heads and one MLP with the smallest scores until the performance on the \textit{dev} set is within 90\%. Then we continue pruning heads alone, and then MLPs alone. This strategy results in a larger number of total components pruned within our performance threshold.
\end{itemize}
\end{multicols}

        \begin{subfigure}[b]{0.49\textwidth}
                \includegraphics[trim=-20 -10 -30 30,clip,width=\linewidth]{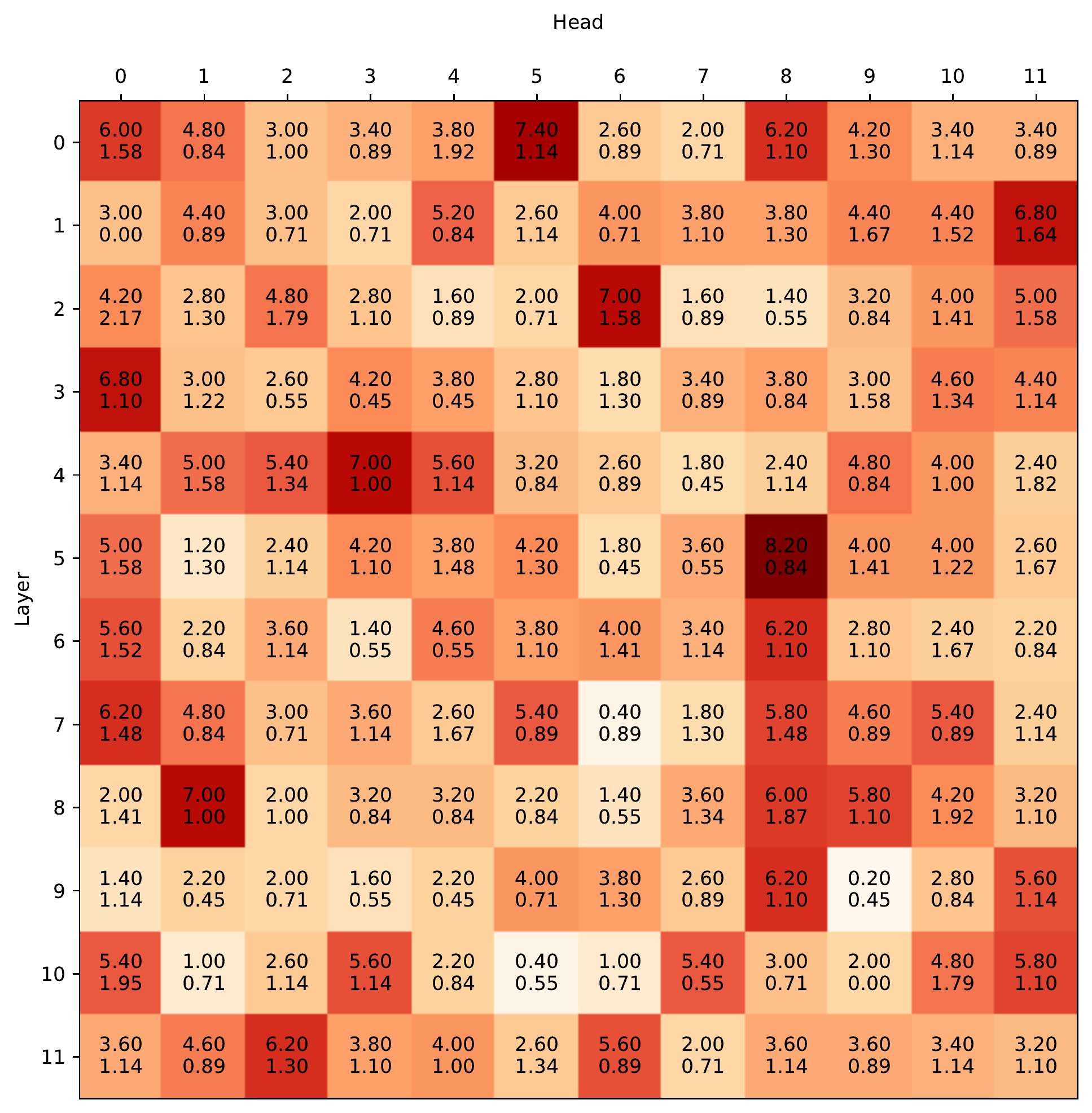}
                \caption{Surviving heads (masking heads only)}
                \label{fig:head-heatmap-separate}
        \end{subfigure}%
        \begin{subfigure}[b]{0.49\textwidth}
                \includegraphics[trim=-20 -10 -30 30,clip,width=\linewidth]{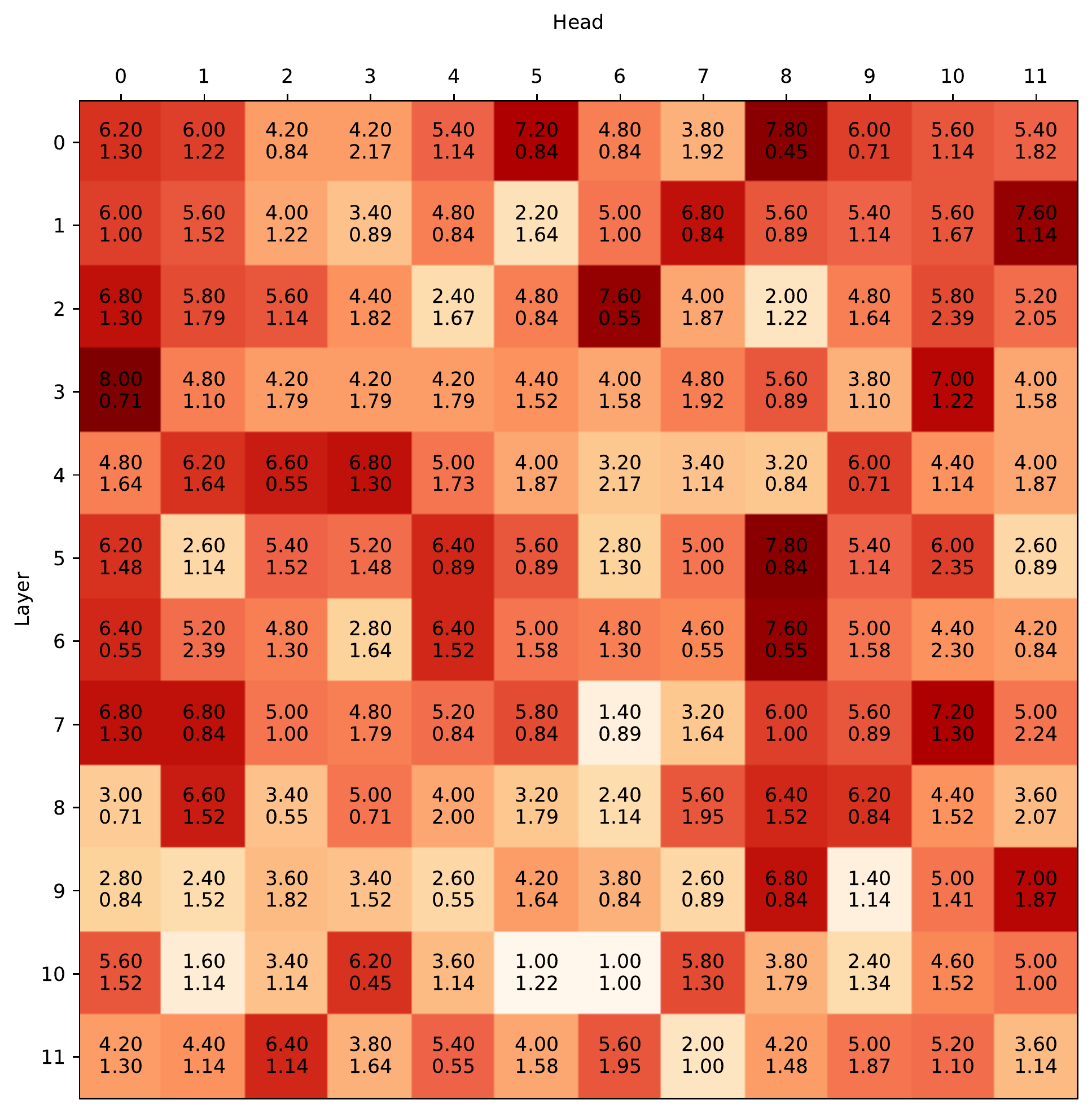}
                \caption{Surviving heads (masking heads and MLPs)}
                \label{fig:head-heatmap-together}
        \end{subfigure}
\centering        
        \begin{subfigure}[b]{0.75\textwidth}
                \includegraphics[trim=-20 -10 -30 30,clip,width=\linewidth]{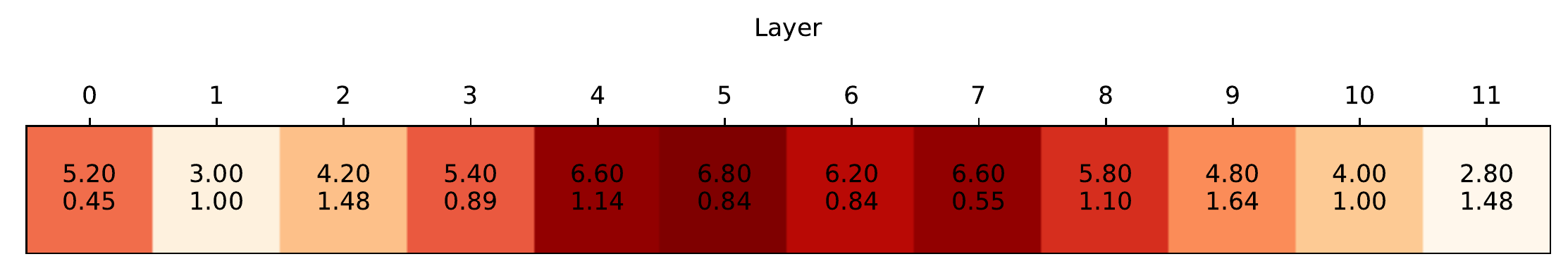}
                \caption{Surviving MLPs (masking MLPs only)}
                \label{fig:mlp-heatmap-separate}
        \end{subfigure}%
        \hfill
        \begin{subfigure}[b]{0.75\textwidth}
                \includegraphics[trim=-20 -10 -30 30,clip,width=\linewidth]{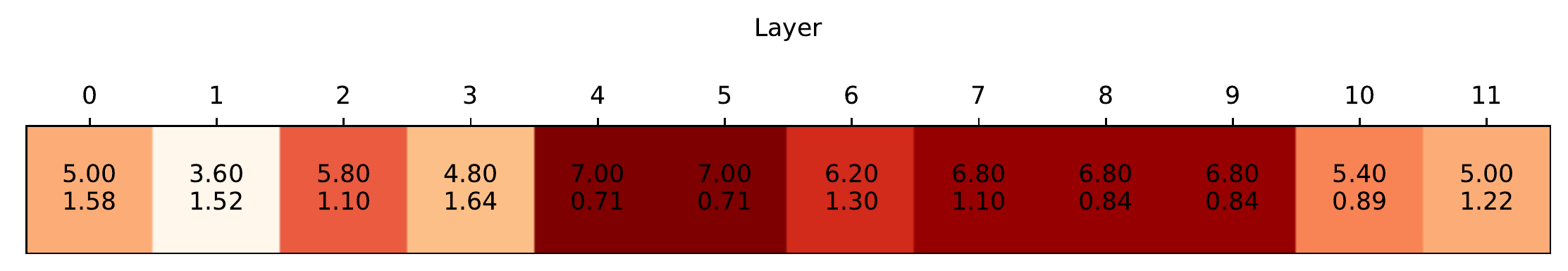}
                \caption{Surviving MLPs (masking heads and MLPs)}
                \label{fig:mlp-heatmap-together}
        \end{subfigure}        
        
        \caption{The ``good" subnetworks: self-attention heads and MLPs that survive pruning. Each cell gives the average  number of GLUE tasks in which a given head/MLP survived, and the standard deviation across 5 fine-tuning initializations.}
        \label{fig:head-heatmap}
\end{figure*}

\clearpage
\begin{sidewaystable*}[t!]
    \section{Evaluation on GLUE Tasks}
    \label{appendix:evaluation-metrics-glue}
    \centering
    \begin{tabular}{|l|l|l|l|l|l|l|l|l|l|}
    \hline
        Experiment & CoLA & MNLI & MRPC & QNLI & QQP & RTE & SST-2 & STS-B & WNLI \\ \hline
        majority baseline & 0.00 $\pm$ 0.00 & 0.35 $\pm$ 0.00 & 0.68 $\pm$ 0.00 & 0.51 $\pm$ 0.00 & 0.63 $\pm$ 0.00 & 0.53 $\pm$ 0.00 & 0.51 $\pm$ 0.00 & 0.02 $\pm$ 0.00 & 0.56 $\pm$ 0.00 \\ \hline
        full model & 0.56 $\pm$ 0.01 & 0.84 $\pm$ 0.00 & 0.84 $\pm$ 0.01 & 0.92 $\pm$ 0.00 & 0.91 $\pm$ 0.00 & 0.63 $\pm$ 0.03 & 0.93 $\pm$ 0.00 & 0.89 $\pm$ 0.00 & 0.34 $\pm$ 0.06 \\ \hline
        \multicolumn{10}{|c|}{S-pruning Subnetworks} \\ \hline
        `good' (pruned) & 0.51 $\pm$ 0.01 & 0.76 $\pm$ 0.00 & 0.77 $\pm$ 0.01 & 0.83 $\pm$ 0.00 & 0.83 $\pm$ 0.01 & 0.57 $\pm$ 0.03 & 0.84 $\pm$ 0.01 & 0.80 $\pm$ 0.01 & 0.54 $\pm$ 0.06 \\ \hline
        `good' (retrained) & 0.50 $\pm$ 0.04 & 0.80 $\pm$ 0.04 & 0.77 $\pm$ 0.06 & 0.89 $\pm$ 0.02 & 0.89 $\pm$ 0.01 & 0.58 $\pm$ 0.06 & 0.87 $\pm$ 0.07 & 0.75 $\pm$ 0.23 & 0.54 $\pm$ 0.06 \\ \hline
        random (pruned) & 0.43 $\pm$ 0.11 & 0.64 $\pm$ 0.19 & 0.48 $\pm$ 0.23 & 0.69 $\pm$ 0.12 & 0.74 $\pm$ 0.07 & 0.51 $\pm$ 0.04 & 0.69 $\pm$ 0.13 & 0.53 $\pm$ 0.25 & 0.54 $\pm$ 0.06 \\ \hline
        random (retrained) & 0.42 $\pm$ 0.23 & 0.79 $\pm$ 0.08 & 0.68 $\pm$ 0.21 & 0.84 $\pm$ 0.05 & 0.88 $\pm$ 0.03 & 0.56 $\pm$ 0.05 & 0.89 $\pm$ 0.01 & 0.79 $\pm$ 0.08 & 0.54 $\pm$ 0.06 \\ \hline
        `bad' (pruned) & 0.34 $\pm$ 0.09 & 0.60 $\pm$ 0.13 & 0.43 $\pm$ 0.15 & 0.63 $\pm$ 0.06 & 0.68 $\pm$ 0.06 & 0.49 $\pm$ 0.04 & 0.62 $\pm$ 0.15 & 0.32 $\pm$ 0.38 & 0.54 $\pm$ 0.06 \\ \hline
        `bad' (retrained) & 0.41 $\pm$ 0.11 & 0.80 $\pm$ 0.05 & 0.68 $\pm$ 0.14 & 0.77 $\pm$ 0.12 & 0.82 $\pm$ 0.11 & 0.59 $\pm$ 0.04 & 0.86 $\pm$ 0.06 & 0.61 $\pm$ 0.21 & 0.54 $\pm$ 0.06 \\ \hline
        \multicolumn{10}{|c|}{Importance Pruning - Super Subnetworks} \\ \hline
        `good' (pruned) & 0.13 $\pm$ 0.06 & 0.34 $\pm$ 0.01 & 0.39 $\pm$ 0.16 & 0.56 $\pm$ 0.03 & 0.63 $\pm$ 0.00 & 0.52 $\pm$ 0.02 & 0.54 $\pm$ 0.03 & 0.38 $\pm$ 0.07 & 0.54 $\pm$ 0.06 \\ \hline
        `good' (retrained) & 0.49 $\pm$ 0.02 & 0.76 $\pm$ 0.00 & 0.71 $\pm$ 0.00 & 0.83 $\pm$ 0.00 & 0.87 $\pm$ 0.00 & 0.50 $\pm$ 0.03 & 0.84 $\pm$ 0.00 & 0.80 $\pm$ 0.00 & 0.56 $\pm$ 0.00 \\ \hline
        random (pruned) & 0.02 $\pm$ 0.05 & 0.32 $\pm$ 0.01 & 0.39 $\pm$ 0.16 & 0.50 $\pm$ 0.01 & 0.63 $\pm$ 0.00 & 0.47 $\pm$ 0.00 & 0.50 $\pm$ 0.01 & 0.06 $\pm$ 0.06 & 0.54 $\pm$ 0.06 \\ \hline
        random (retrained) & 0.15 $\pm$ 0.15 & 0.75 $\pm$ 0.01 & 0.69 $\pm$ 0.01 & 0.76 $\pm$ 0.08 & 0.83 $\pm$ 0.04 & 0.50 $\pm$ 0.03 & 0.85 $\pm$ 0.00 & 0.12 $\pm$ 0.03 & 0.56 $\pm$ 0.00 \\ \hline
        `bad' (pruned) & 0.01 $\pm$ 0.01 & 0.32 $\pm$ 0.00 & 0.39 $\pm$ 0.16 & 0.49 $\pm$ 0.02 & 0.60 $\pm$ 0.07 & 0.52 $\pm$ 0.02 & 0.51 $\pm$ 0.02 & 0.06 $\pm$ 0.02 & 0.54 $\pm$ 0.06 \\ \hline
        `bad' (retrained) & 0.13 $\pm$ 0.03 & 0.77 $\pm$ 0.00 & 0.69 $\pm$ 0.01 & 0.57 $\pm$ 0.03 & 0.87 $\pm$ 0.00 & 0.49 $\pm$ 0.03 & 0.83 $\pm$ 0.01 & 0.13 $\pm$ 0.01 & 0.56 $\pm$ 0.00 \\ \hline
        \multicolumn{10}{|c|}{M-pruning Subnetworks} \\ \hline
        `good' (pruned) & 0.52 $\pm$ 0.01 & 0.78 $\pm$ 0.00 & 0.78 $\pm$ 0.02 & 0.84 $\pm$ 0.01 & 0.83 $\pm$ 0.01 & 0.61 $\pm$ 0.01 & 0.85 $\pm$ 0.01 & 0.82 $\pm$ 0.02 & 0.48 $\pm$ 0.10 \\ \hline
        `good' (retrained) & 0.54 $\pm$ 0.02 & 0.84 $\pm$ 0.00 & 0.84 $\pm$ 0.01 & 0.91 $\pm$ 0.00 & 0.91 $\pm$ 0.00 & 0.61 $\pm$ 0.02 & 0.92 $\pm$ 0.00 & 0.88 $\pm$ 0.00 & 0.41 $\pm$ 0.11 \\ \hline
        random (pruned) & 0.02 $\pm$ 0.03 & 0.33 $\pm$ 0.01 & 0.39 $\pm$ 0.16 & 0.51 $\pm$ 0.02 & 0.63 $\pm$ 0.00 & 0.47 $\pm$ 0.00 & 0.55 $\pm$ 0.09 & 0.08 $\pm$ 0.04 & 0.51 $\pm$ 0.08 \\ \hline
        random (retrained) & 0.16 $\pm$ 0.06 & 0.76 $\pm$ 0.00 & 0.70 $\pm$ 0.01 & 0.81 $\pm$ 0.01 & 0.86 $\pm$ 0.00 & 0.56 $\pm$ 0.02 & 0.83 $\pm$ 0.01 & 0.24 $\pm$ 0.03 & 0.47 $\pm$ 0.12 \\ \hline
        `bad' (pruned) & 0.00 $\pm$ 0.00 & 0.32 $\pm$ 0.00 & 0.39 $\pm$ 0.16 & 0.51 $\pm$ 0.02 & 0.63 $\pm$ 0.00 & 0.49 $\pm$ 0.03 & 0.49 $\pm$ 0.01 & 0.05 $\pm$ 0.07 & 0.53 $\pm$ 0.06 \\ \hline
        `bad' (retrained) & 0.02 $\pm$ 0.03 & 0.62 $\pm$ 0.00 & 0.68 $\pm$ 0.01 & 0.61 $\pm$ 0.00 & 0.78 $\pm$ 0.00 & 0.49 $\pm$ 0.03 & 0.82 $\pm$ 0.00 & 0.08 $\pm$ 0.00 & 0.49 $\pm$ 0.11 \\ \hline
    \end{tabular}
    \caption{Mean and standard deviation of GLUE tasks metrics evaluated on five seeds.}
\end{sidewaystable*}
\clearpage

\begin{figure*}[!ht]
\section{Longer Fine-tuning of ``Bad" s-pruned Subnetworks}
\label{appendix:train-badsubnets-longer}

{
\centering
\footnotesize
\begin{tabular}{lrrrrrrrrrr}
\toprule
Epoch                   & CoLA & SST-2 & MRPC & QQP  & STS-B & MNLI & QNLI & RTE  & WNLI & Avg\\ \midrule
3  & 0.422  & \textbf{0.873}  & \textbf{0.71}  & \textbf{0.832}    & 0.651  & \textbf{0.805}  & \textbf{0.764}  & 0.579    & 0.498  & \textbf{0.6815}   \\
4  & 0.423  & 0.859  & 0.663  & 0.828    & 0.652  & 0.804  & 0.762  & 0.587    & \textbf{0.554}  & 0.6813    \\
5  & \textbf{0.432}  & 0.862  & 0.665  & 0.831    & \textbf{0.668}  & 0.801  & 0.752  & 0.590  & 0.523  & 0.6804    \\
6  & 0.425  & 0.867  & 0.655  & 0.830  & 0.667  & 0.800  & 0.753  & \textbf{0.594}    & 0.521  & 0.6791   \\
\bottomrule
\end{tabular}

\caption{The mean of GLUE tasks metrics evaluated on five seeds at different epochs (the best one is bolded). \small{*Slight divergence in metrics from the previously reported ones due to this being an new fine-tuning run.}}
}
\end{figure*}
\begin{figure*}[!ht]

\section{Performance of the ``Super Survivor" Subnetworks}
\label{appendix:super-survivors}

\begin{multicols}{2}

In this experiment, we explore three settings:

\begin{itemize*}
    \item \textit{``good" subnetworks}: the subnetworks consisting only of ``super-survivors": the self-attention heads and MLPs that survived in all random seeds, shown in \autoref{appendix:good-subnets-per-task}. These subnetworks are much smaller than the pruned subnetworks discussed in \autoref{sec:lottery} (10-30\% vs 50-70\% of the full model);
    \item \textit{``bad" subnetworks}: the subnetworks the same size as the super-survivor subnetworks, but selected from heads and MLPs the \textit{least} likely to survive importance pruning;
    \item \textit{random subnetworks}: same size as super-survivor subnetworks, but selected from elements that were neither super-survivors, nor the ones in the ``bad" subnetworks.
\end{itemize*}

The striking conclusion is that on 6 out of 9 tasks the bad and random subnetworks behaved nearly as well as the ``good" ones, suggesting that the ``super-survivor" self-attention heads and MLPs did \textit{not} survive importance pruning because of their encoding some unique linguistic information necessary for solving the GLUE tasks.

\end{multicols}

\includegraphics[width=\linewidth]{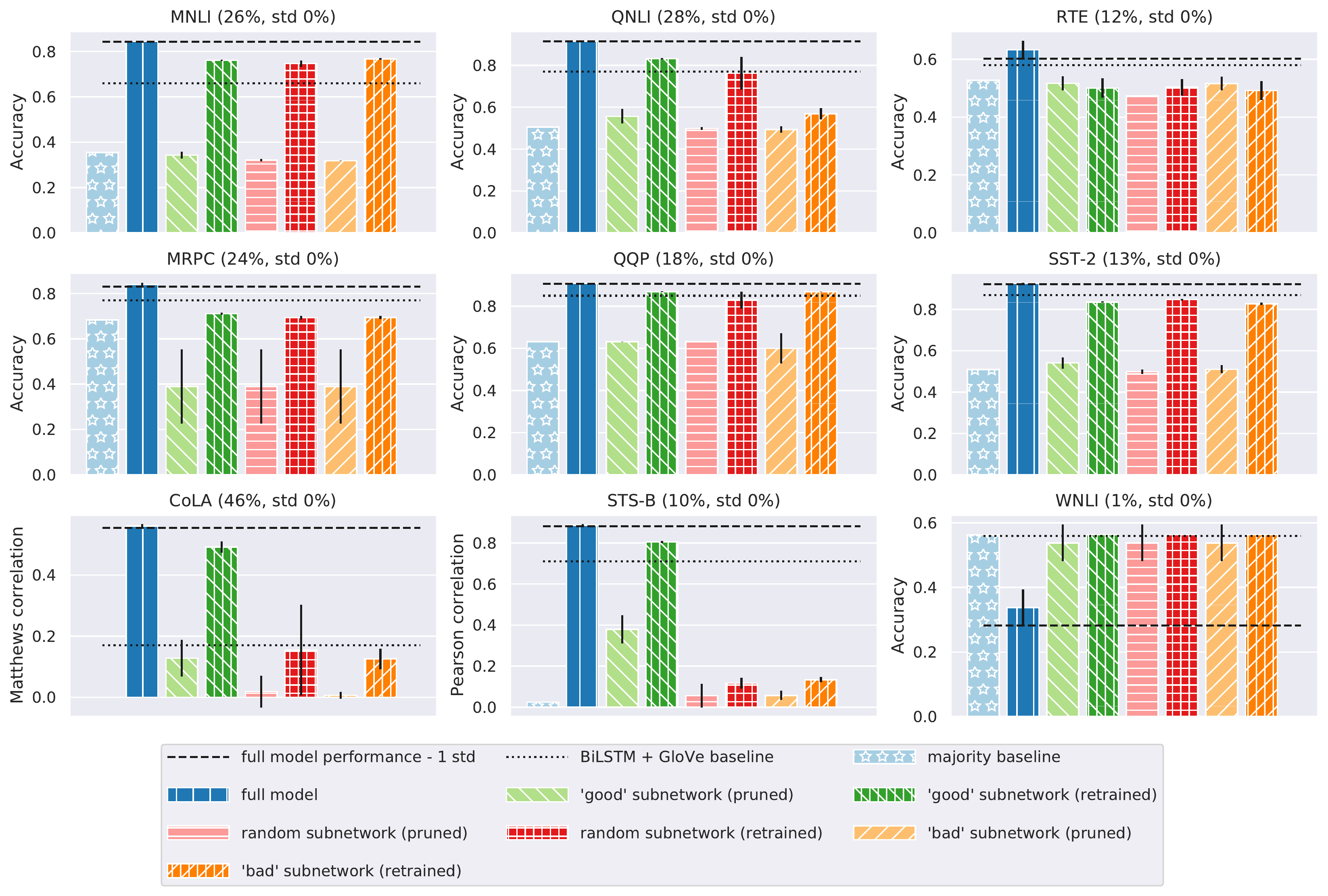}
            \caption{The performance of ``super survivor" subnetworks in BERT fine-tuning: performance on GLUE tasks (error bars indicate standard deviation across 5 fine-tuning runs). The size of the super-survivor subnetwork as \% of full model weights is shown next to the task names.}
            \label{fig:super-survivors-barplot}
\end{figure*}
\clearpage

\begin{figure*}[!ht]
\section{Attention Pattern Type Distribution}
\label{app:attn-patterns-all}

\begin{multicols}{2}
We use two separately trained CNN classifiers to analyze the BERT's self-attention maps, both ``raw" head outputs and weight-normed attention, following \citet{KobayashiKuribayashiEtAl_2020_Attention_Module_is_Not_Only_Weight_Analyzing_Transformers_with_Vector_Norms}. For the former, we use 400 annotated maps by \citet{KovalevaRomanovEtAl_2019_Revealing_Dark_Secrets_of_BERT}, and for the latter we additionally annotate 600 more maps.

We run the classifiers on pre-trained and fine-tuned BERT, both the full model and the model pruned by the ``super-survivor" mask (only the heads and MLPs that survived across GLUE tasks). For each experiment, we report the fraction of attention patterns %
estimated from a hundred dev-set samples for each task across five random seeds.

See \autoref{fig:attention-types} for attention types illustration.

\end{multicols}

\vspace{0.5cm}

        \begin{subfigure}[b]{0.49\textwidth}
            \includegraphics[trim=10 10 10 10,clip,width=\linewidth]{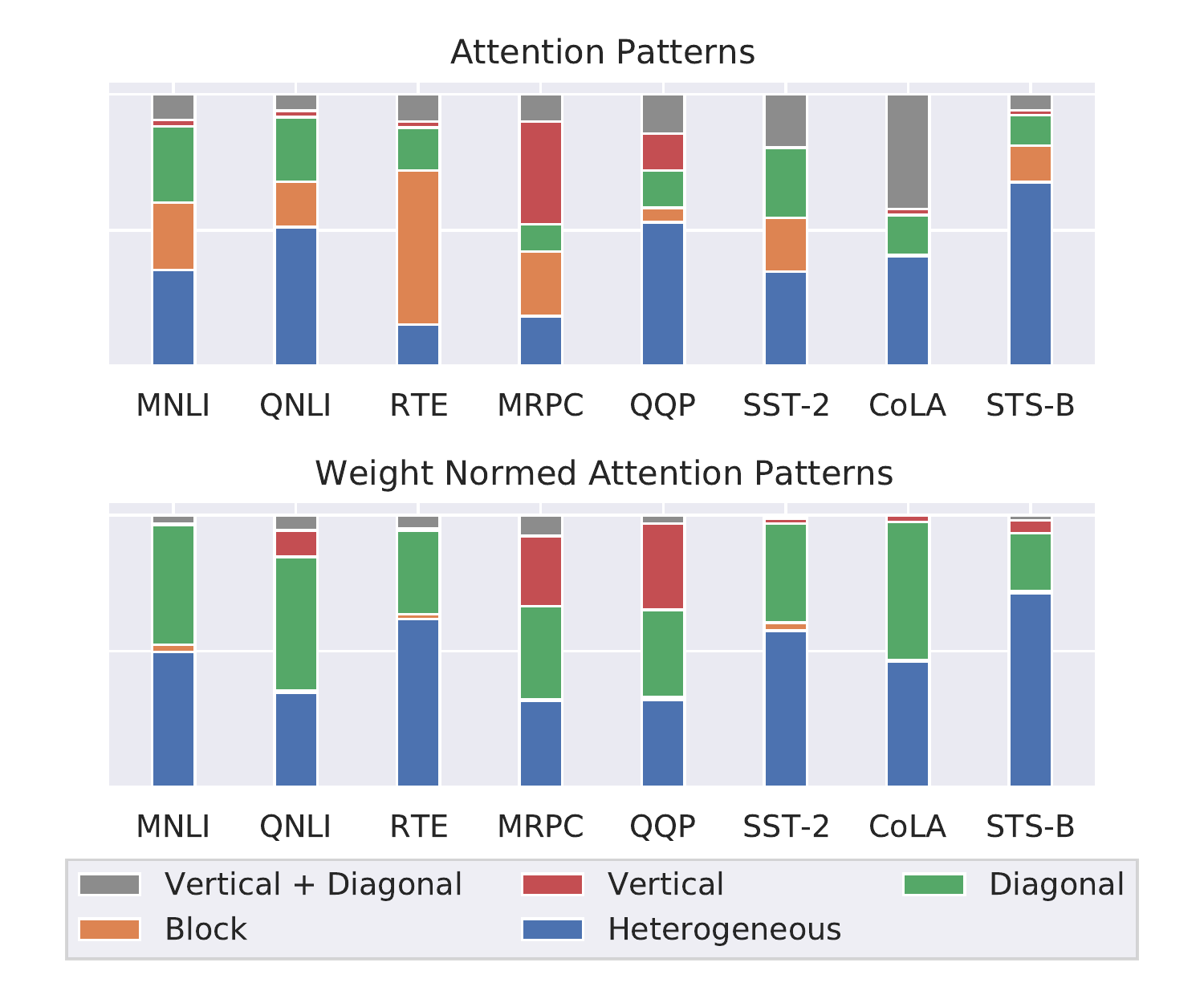}
            \subcaption{Super-survivor heads, fine-tuned}
        \end{subfigure}
        \hfill
        \begin{subfigure}[b]{0.49\textwidth}
            \includegraphics[trim=10 10 10 10,clip,width=\linewidth]{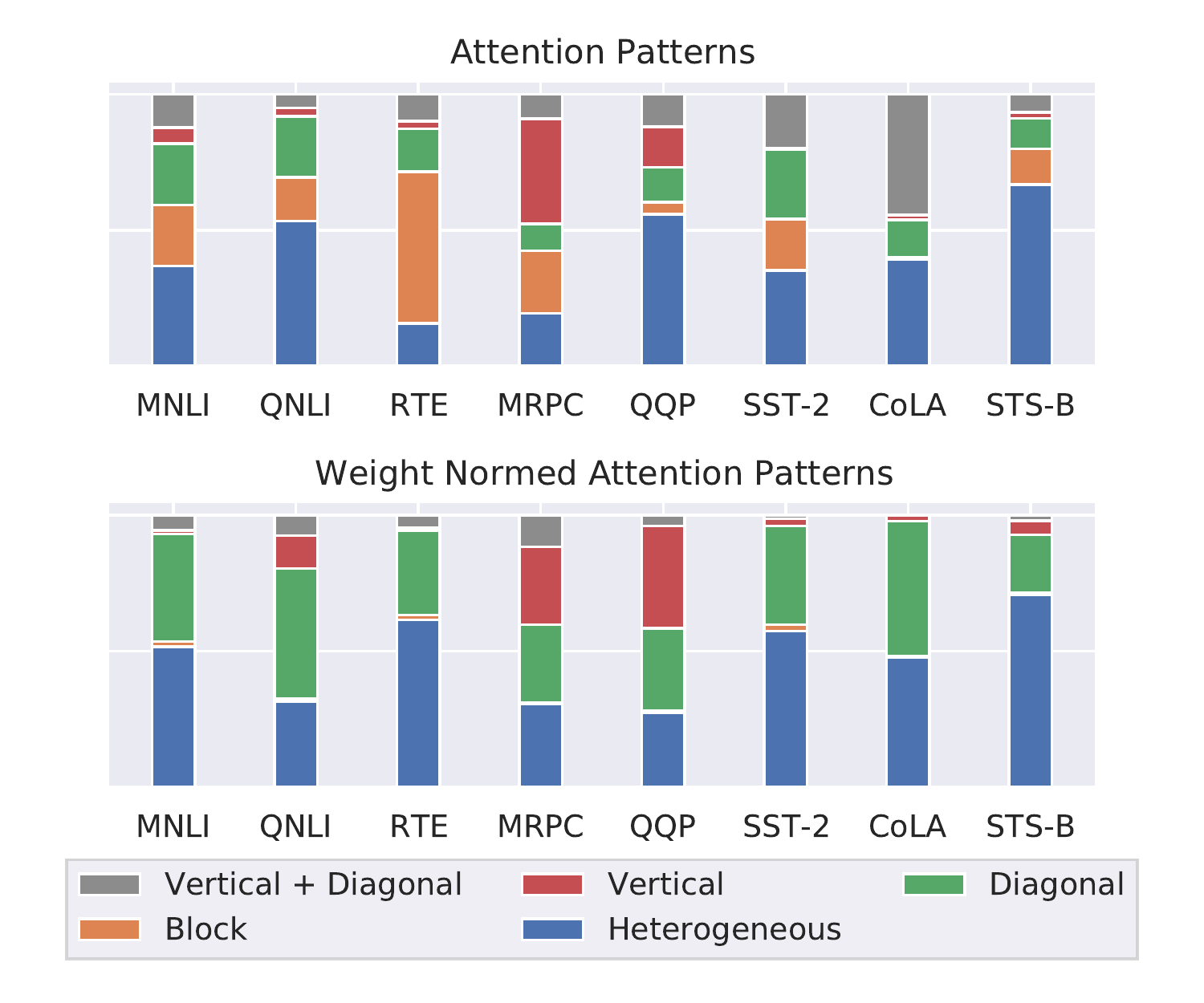}
            \subcaption{Super-survivor heads, pre-trained}
        \end{subfigure}
        \\[3ex]
        \begin{subfigure}[b]{0.49\textwidth}
            \includegraphics[trim=10 10 10 10,clip,width=\linewidth]{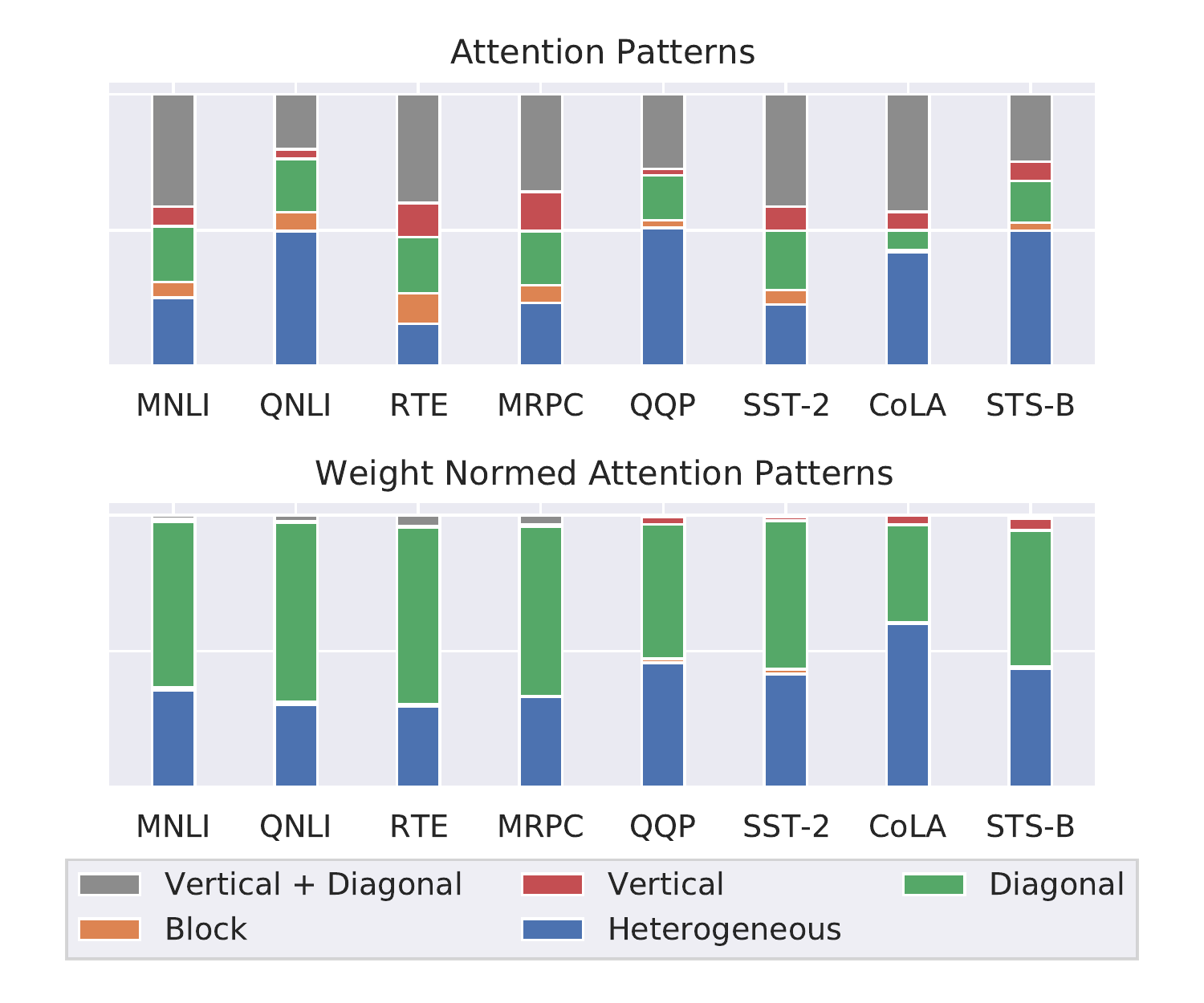}
            \subcaption{All heads, fine-tuned}
        \end{subfigure}
        \hfill
        \begin{subfigure}[b]{0.49\textwidth}
            \includegraphics[trim=10 10 10 10,clip,width=\linewidth]{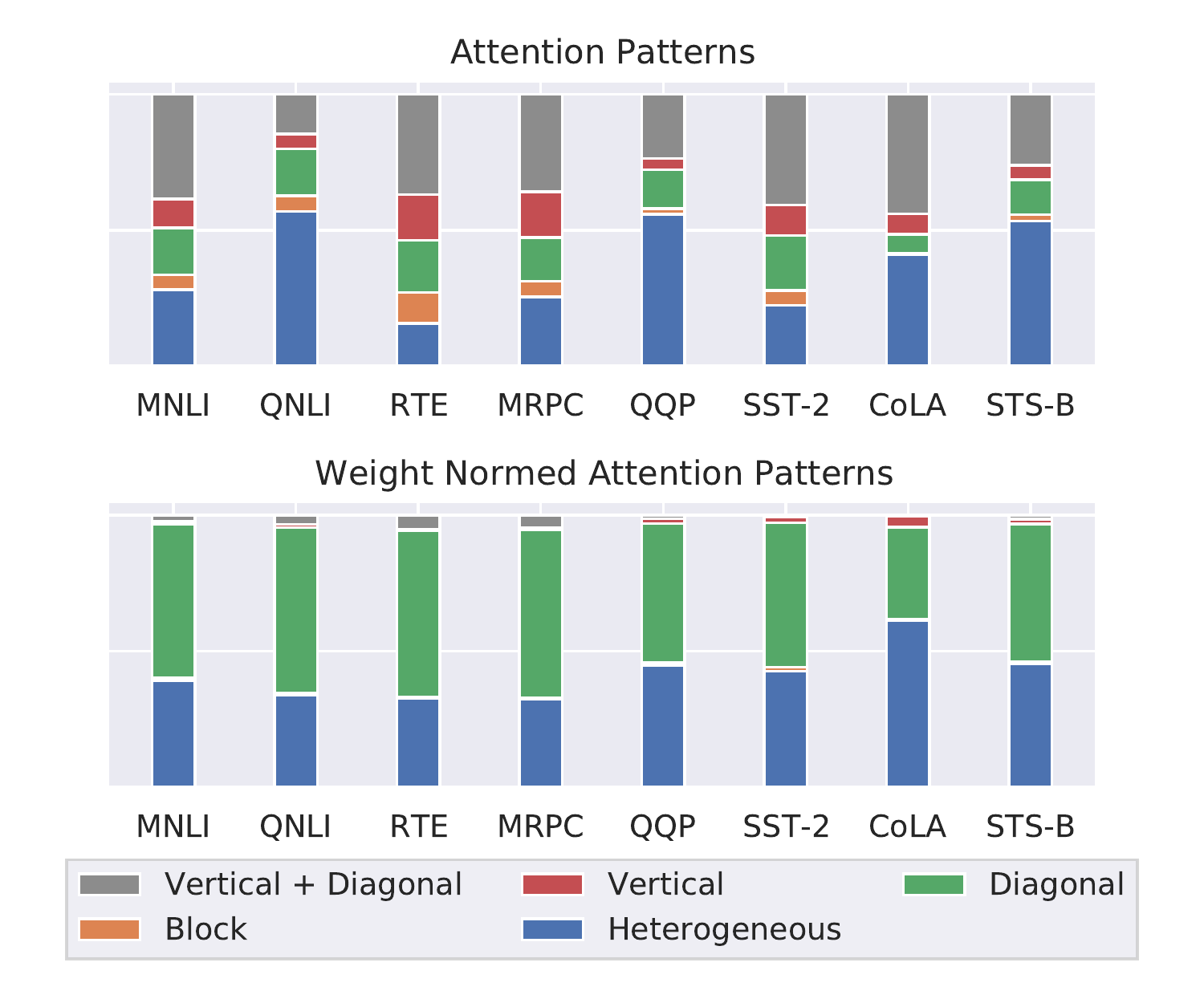}
            \subcaption{All heads, pre-trained}
        \end{subfigure}
\caption{Attention pattern distribution in all BERT self-attention heads and the ``super-survivor" heads.}
\label{fig:attn-patterns-all}
\end{figure*}

\begin{figure*}

\section{How Task-independent are the ``Good" Subnetworks?}
\label{appendix:task-pairs}

\begin{multicols}{2}
The parts of the ``good" subnetworks that are only relevant for some specific tasks, but consistently survive across fine-tuning runs for that task, should encode the information useful for that task, even if it is not deeply linguistic. Therefore, the degree to which the ``good" subnetworks overlap across tasks may be a useful way to characterize the tasks themselves. 

\autoref{fig:task-pairs-all} shows pairwise comparisons between all GLUE tasks with respect to the number of shared heads and MLPs in two conditions: the ``good" subnetworks found by structured importance pruning that were described in \autoref{sec:good-subnets}, and super-survivors (the heads/MLPs that survived in all random seeds).

\end{multicols}

        \begin{subfigure}[b]{0.49\textwidth}
                \includegraphics[trim=20 20 20 35,clip,width=\linewidth]{figs/task_v_task_heads_together.pdf}
                \caption{Heads shared between tasks}
                \label{fig:task-pairs-heads-2}
        \end{subfigure}%
        \begin{subfigure}[b]{0.49\textwidth}
                \includegraphics[trim=20 20 20 35,clip,width=\linewidth]{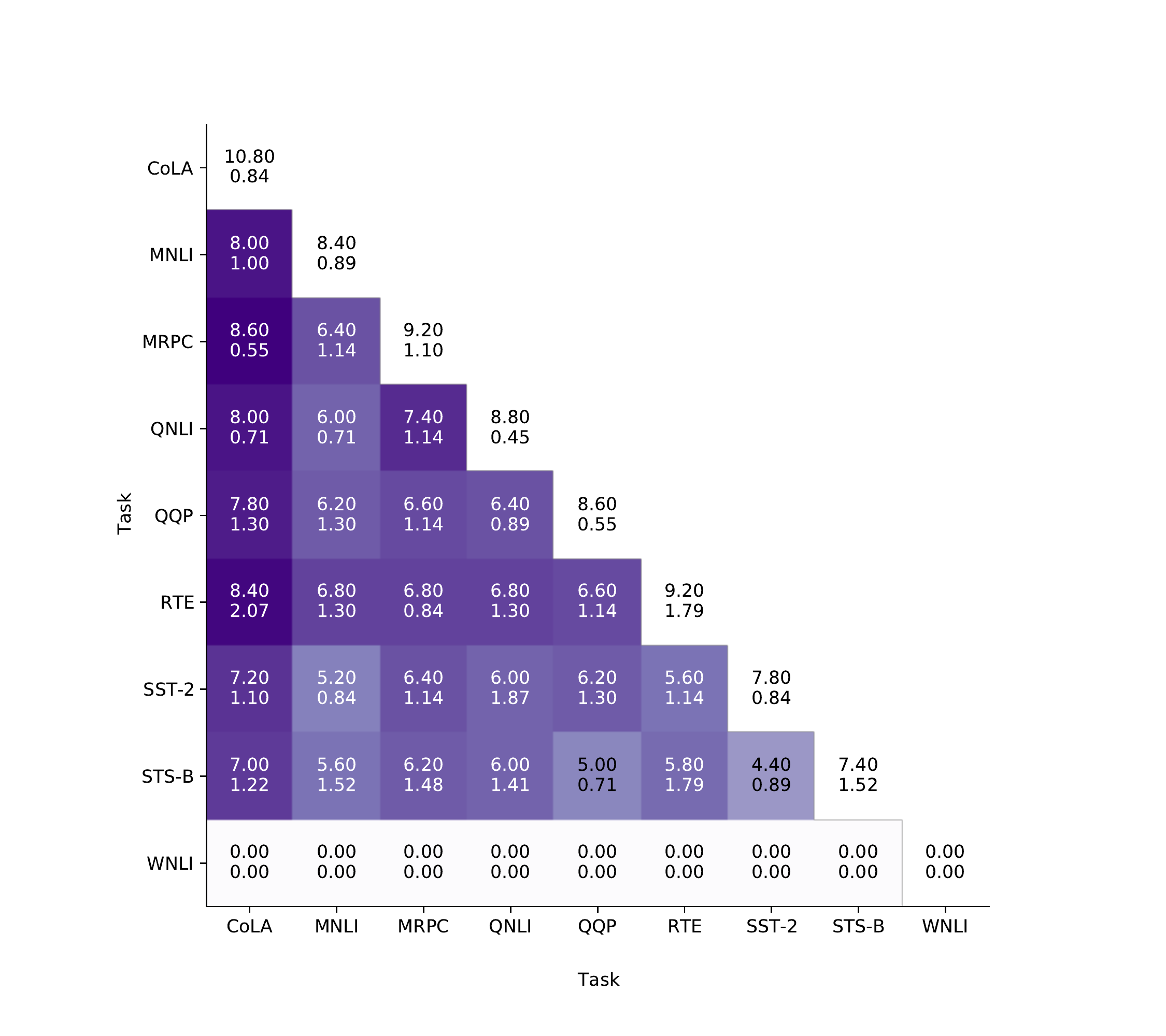}
                \caption{MLPs shared between tasks}
                \label{fig:task-pairs-mlps-2}
        \end{subfigure}
        
         \begin{subfigure}[b]{0.49\textwidth}
                \includegraphics[trim=20 20 20 35,clip,width=\linewidth]{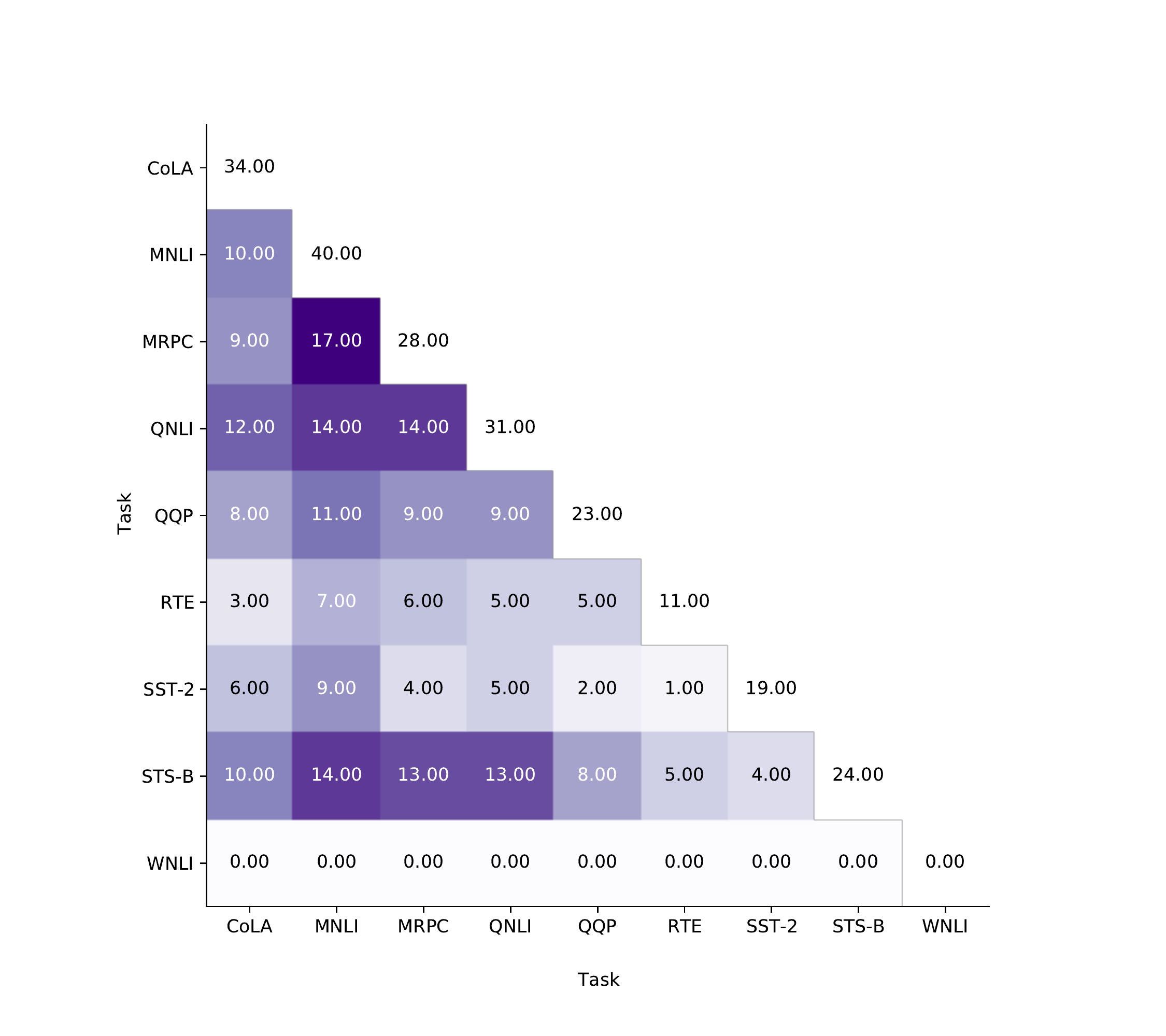}
                \caption{Super-survivor Heads shared between tasks}
                \label{fig:task-pairs-heads-super}
        \end{subfigure}%
        \begin{subfigure}[b]{0.49\textwidth}
                \includegraphics[trim=20 20 20 35,clip,width=\linewidth]{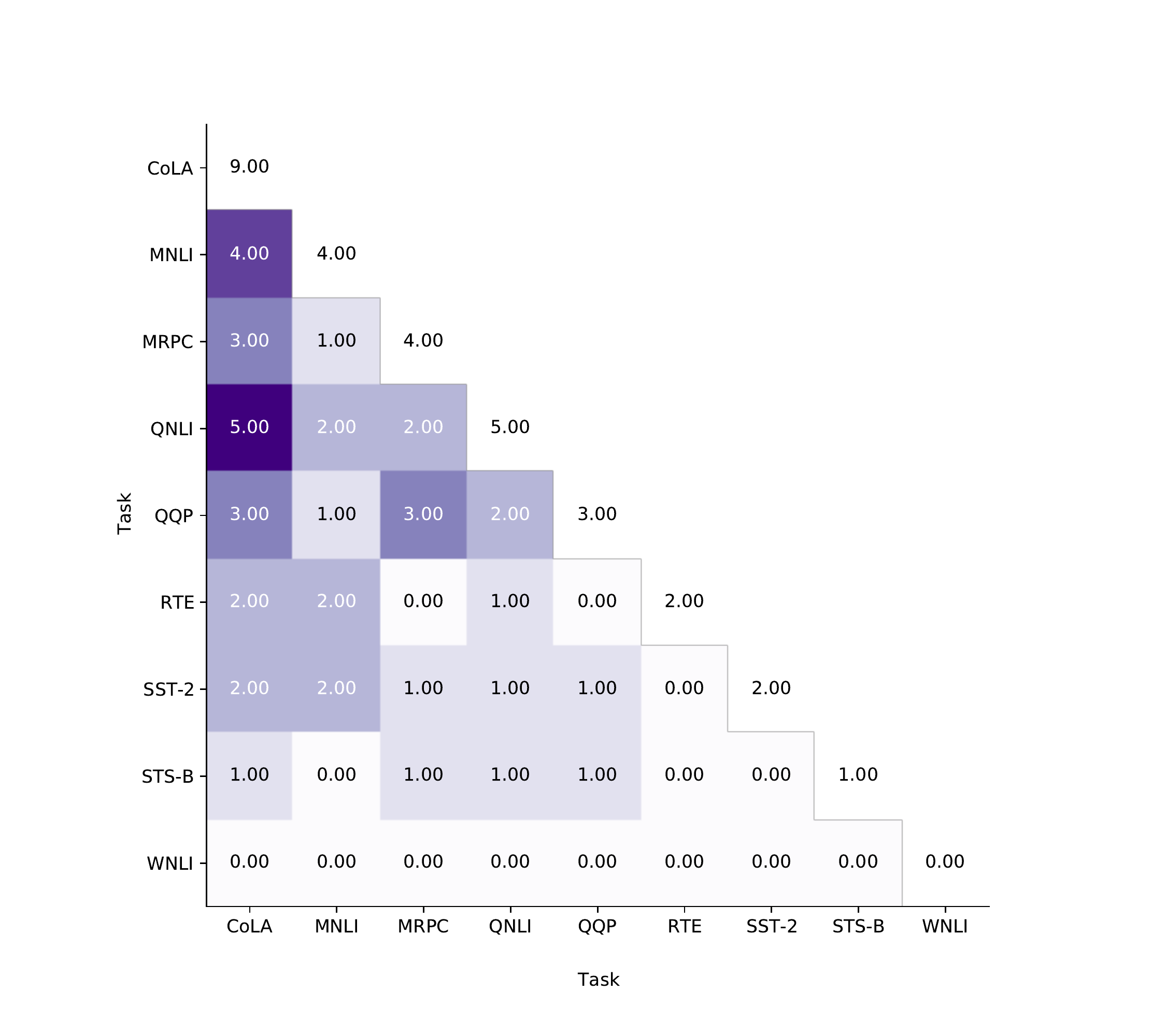}
                \caption{Super-survivor MLPs shared between tasks}
                \label{fig:task-pairs-mlps-super}
        \end{subfigure}
        
        \caption{The ``good" subnetwork: The diagonal represents the BERT architecture components that survive pruning for a given task and remaining elements represent the common surviving components across GLUE tasks. Each cell gives the average number of heads (out of 144) or layers (out of 12), together with standard deviation across 5 random initializations (for (a) and (b)).}
        \label{fig:task-pairs-all}
\end{figure*}

\end{document}